\documentclass[manuscript]{acmart}

\AtBeginDocument{%
  \providecommand\BibTeX{{%
    \normalfont B\kern-0.5em{\scshape i\kern-0.25em b}\kern-0.8em\TeX}}}

\copyrightyear{2024}
\acmYear{2024}
\setcopyright{rightsretained}
\acmConference[FAccT '24]{The 2024 ACM Conference on Fairness, Accountability, and Transparency}{June 3--6, 2024}{Rio de Janeiro, Brazil}
\acmBooktitle{The 2024 ACM Conference on Fairness, Accountability, and Transparency (FAccT '24), June 3--6, 2024, Rio de Janeiro, Brazil}\acmDOI{10.1145/3630106.3659039}
\acmISBN{979-8-4007-0450-5/24/06}

\usepackage{booktabs}
\usepackage{adjustbox}
\usepackage{subcaption}
\usepackage{wrapfig}
\usepackage{multirow}

\begin{document}

\title{Embracing Diversity: Interpretable Zero-shot classification beyond one vector per class}

\author{Mazda Moayeri}
\authornote{Work done while interning with FAIR at Meta.}
\email{mmoayeri@umd.edu}
\affiliation{%
  \institution{University of Maryland}
  \city{College Park}
  \country{USA}
}

\author{Michael Rabbat}
\affiliation{%
  \institution{FAIR at Meta}
  \city{Montreal}
  \country{Canada}
  }
\author{Mark Ibrahim}
\authornote{Equal contribution, co-principal investigators.}
\affiliation{%
  \institution{FAIR at Meta}
  \city{New York City}
  \country{USA}
  }
\author{Diane Bouchacourt}
\authornotemark[2]
\affiliation{%
  \institution{FAIR at Meta}
  \city{Montreal}
  \country{Canada}
  }

\renewcommand{\shortauthors}{Moayeri, Rabbat, Ibrahim*, and Bouchacourt*.}

\begin{abstract}
Vision-language models enable open-world classification of objects without the need for any retraining. While this zero-shot paradigm marks a significant advance, even today’s best models exhibit skewed performance when objects are dissimilar from their typical depiction. Real world objects such as pears appear in a variety of forms --- from diced to whole, on a table or in a bowl ---
yet standard VLM classifiers map all instances of a class
to a \textit{single vector based on the class label}. 
We argue that to represent this rich diversity within a class, zero-shot classification should move beyond a single vector. 
We propose a method to encode and account for diversity within a class using inferred attributes, still in the zero-shot setting without retraining.
We find our method consistently outperforms standard zero-shot classification over a large suite of datasets encompassing hierarchies, diverse object states, and 
real-world geographic diversity, as well finer-grained datasets where intra-class diversity may be less prevalent. Importantly, our method is inherently interpretable, offering faithful explanations for each inference to facilitate model debugging and enhance transparency.
We also find our method scales efficiently to a large number of attributes to account for diversity---leading to more accurate predictions for atypical instances.
Finally, we characterize a principled trade-off between overall and worst class accuracy, which can be tuned via a hyperparameter of our method.
We hope this work spurs further research into the promise of zero-shot classification beyond a single class vector for capturing diversity in the world, and building transparent AI systems without compromising performance.
\end{abstract}



\begin{teaserfigure}
    \centering
    \begin{minipage}{0.69\textwidth}
    \includegraphics[width=\textwidth]{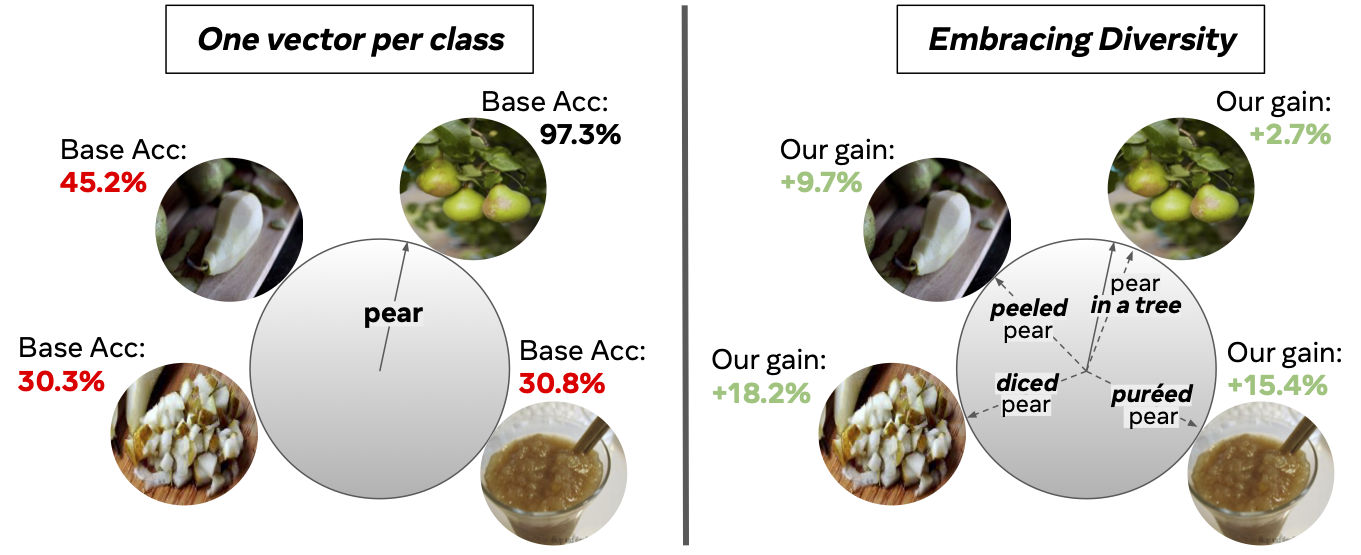}
    \end{minipage}
    \begin{minipage}{0.3\textwidth}
    \centering
    \includegraphics[width=\textwidth]{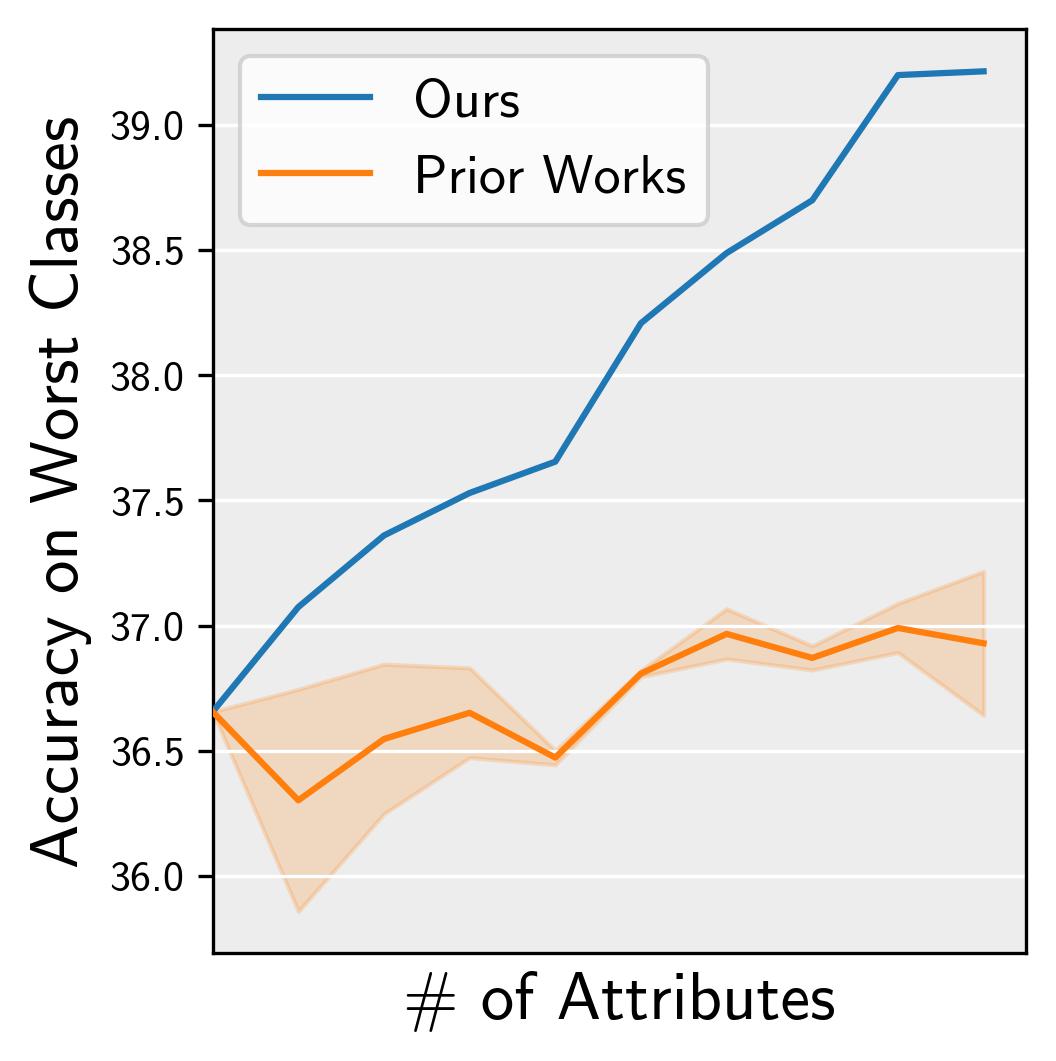}
    \end{minipage}
    \caption{(\textbf{Left}) Instances of a class can appear in many diverse ways, like the pears above. Using one vector (the classname embedding) to represent the whole class results in disparate performance, particularly for atypical instances. (\textbf{Middle}) To address this issue, we infer attributes and embed multiple vectors, reducing disparities and enhancing interpretability. (\textbf{Right}) Our method scales better than prior works as we include more attributes (Section \ref{sec:adding_in_attrs}), enabling us to account for the many ways in which diversity can arise.}
    \label{fig:fig1}
\end{teaserfigure}

\maketitle

\section{Introduction}

A pivotal advance in machine learning is the advent of \textit{foundation models}.
A single foundation model trained on large-scale data can supplant multiple task-specific models.
Vision-Language models (VLMs) are popular foundation models capable of encoding text and images in the same representation space.
Compared to standard classifiers which can only classify objects from a predefined list of classes with examples, VLMs are capable of open-world, zero-shot classification---meaning, VLMs can classify any object using text descriptions without any additional training.
This zero-shot paradigm has spurred the development of many VLMs \cite{radford2021learning, li2023blip2, Yu2022CoCaCC} with impressive classification performance.

\begin{figure}
    \centering
    \begin{subfigure}{0.322\linewidth}
    \centering
    \includegraphics[width=\textwidth]{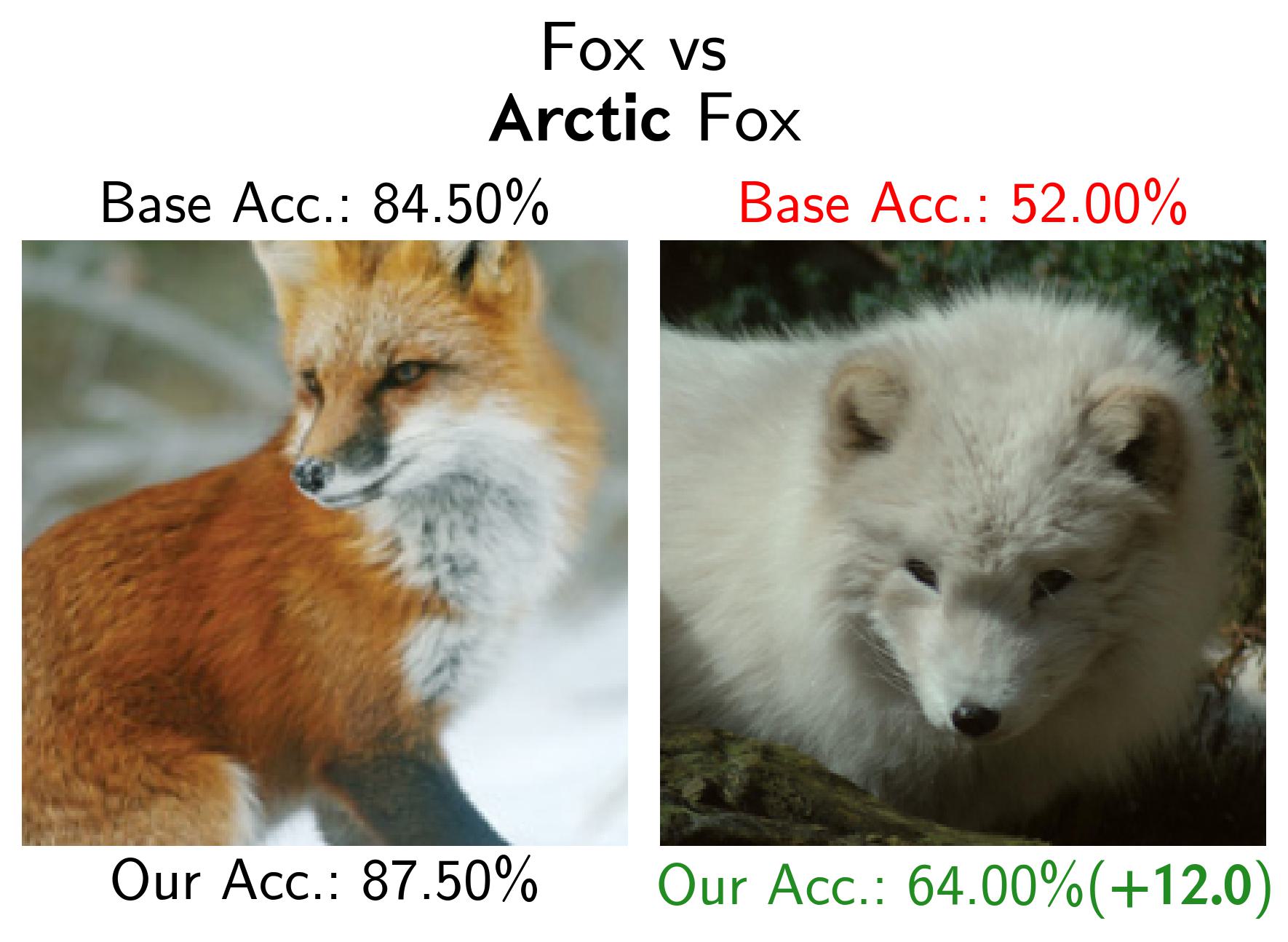}
    \end{subfigure}
    \begin{subfigure}{0.322\linewidth}
    \centering
    \includegraphics[width=\textwidth]{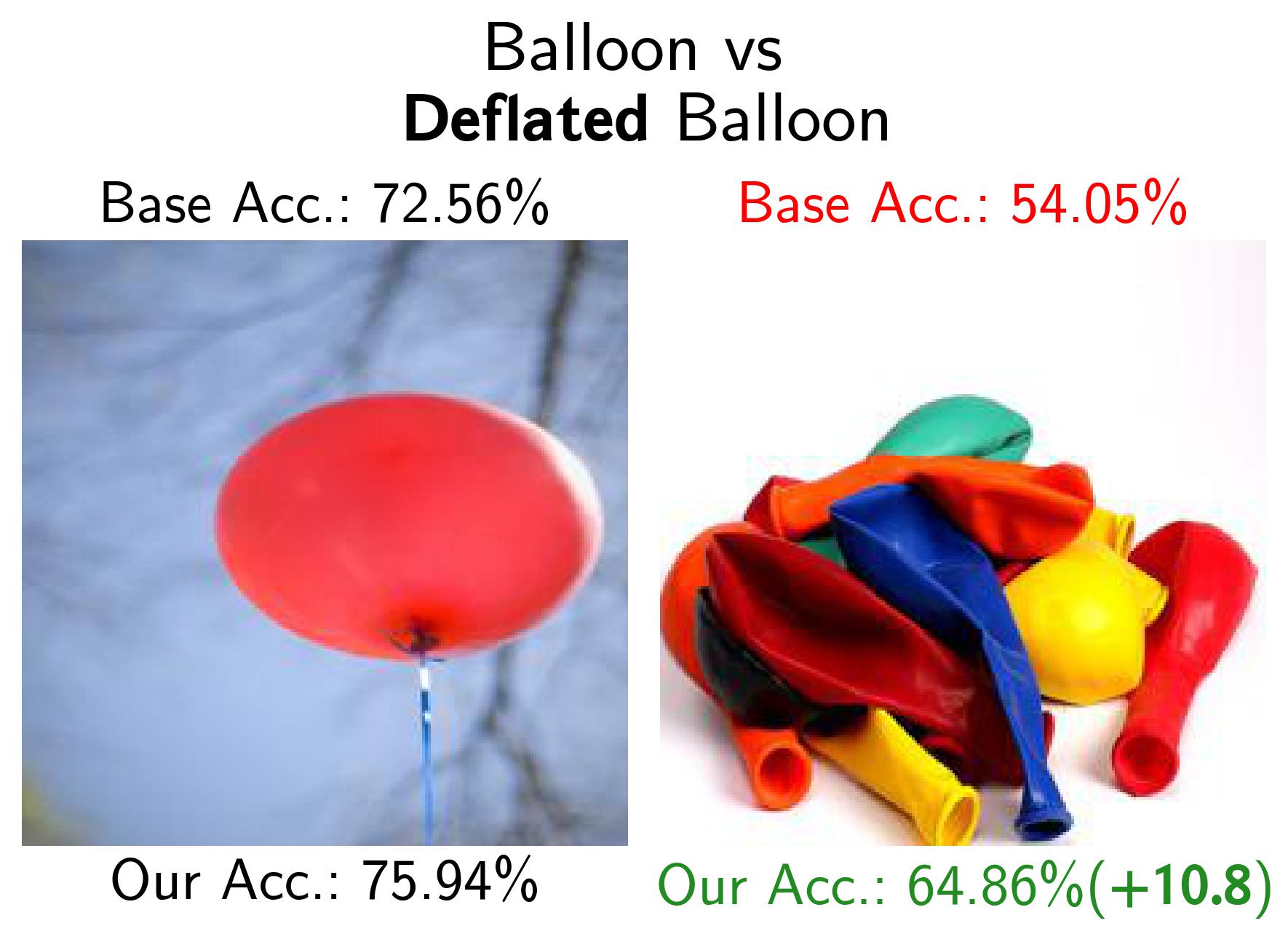}
    \end{subfigure}
    \begin{subfigure}{0.328\linewidth}
    \centering
    \includegraphics[width=\textwidth]{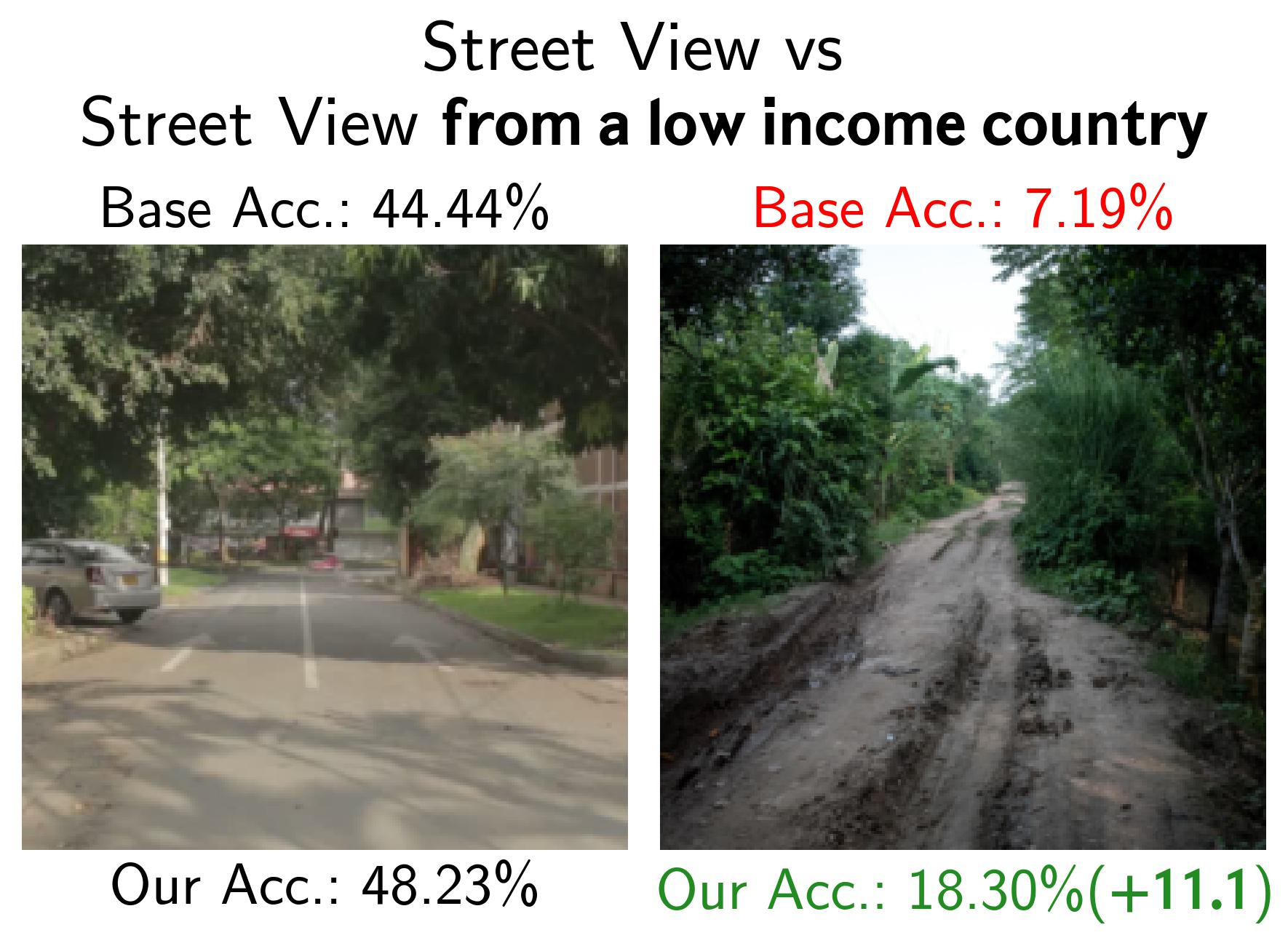}
    \end{subfigure}
    \caption{We test models on datasets that provide groundtruth \textbf{attributes} (shown in bold) annotating hierarchies, diverse states, and real-world shifts (e.g., \citet{rojas2022the} labels the income level and country of origin of each image, towards promoting AI models that reduce bias) within a class. We find that standard zero-shot accuracy (`Base Acc.' above) drops significantly when certain attributes are present, namely when the attribute manifests in visual differences from what the model considers `typical' for the class. We design our method to improve performance on these `atypical' instances.}
    \label{fig:bias_examples}
\end{figure}

Despite their remarkable performance, even today's best models exhibit skewed performance for certain groups of images.
For example, \cite{richards2023does} show models such as CLIP have exacerbated the gap in performance between regions such as Africa and Europe (as well as the gap across income-levels).
We find similar biases arise when an object is visually dissimilar from its typical depiction. 
For example, Figure \ref{fig:fig1} (left) shows CLIP's $97.3\%$ accuracy on typical pears drops dramatically when a pear is peeled ($45.2\%$) or pur\'eed ($30.3\%$).
Addressing such biases is crucial to the reliability of classifiers in the real world, where instances within a class can vary significantly.

\looseness=-1
Zero-shot classifiers, like standard models, use a single vector in deep embedding space to describe an entire class. 
For standard zero-shot classification, a vision-language model 
(i) encodes the image along with 80 hand-crafted prompts per class name (e.g., ``a photo of a pear'' or ``a drawing of a pear''), 
(ii) averages the $80$ embeddings per class to obtain a single vector, 
(iii) predicts the class whose vector maximizes cosine similarity to the image embedding \citep{radford2021learning}.
Prompt averaging encourages all instances of a class to be mapped to the same vector in the model's embedding, inherently limiting the model's ability to infer the innumerable diversity within a class.
A pear can be diced, sliced, whole, in one's hand, or in a bowl. In each case, the image of the pear would be markedly different, and its embedding may not always be well aligned with the single vector that is supposed to represent the \emph{entire} class. Thus, there is a natural tension between the one vector per class  paradigm and performing consistently across a class with high diversity, which we empirically validate. 

While many strategies exist to mitigate performance disparities when labeled-data is available, these methods do not transfer to the data-free setting of zero-shot classification. 
Fortunately, unlike standard classifiers, the open-world nature of VLMs enables them to represent any attribute using the text encoder. 
VLMs can enrich the single per-class vector with attributes to more faithfully capture the variety with which a class can appear, pinpointing whether a pear is peeled or pur\'eed. 
Thus, we argue that instead of learning one vector per class that is invariant to diversity, 
we should leverage the open-world nature of VLMs to \emph{explicitly account for} the diversity within a class (i.e., via multiple vectors). 




Recent works offer promising signs that zero-shot classification can be improved by incorporating attributes beyond the class name, such as subclasses \citep{novack2023chils} or visual descriptors \citep{menon2022visual, pratt2023what}. However, the former is limited to datasets with hierarchical label sets, and the latter reverts back to the one vector per class paradigm via simple averaging, limiting the benefits of incorporating more attributes (Section~\ref{sec:adding_in_attrs}). 
Importantly, diversity comes in many forms that generic descriptors or subclasses alone may not adequately capture.  

In this work, we propose a zero-shot method for enriching classes with open-ended attributes to boost zero-shot classification. 
Our method consists of two steps: (i) an attribute inference step, in which we use generative language modeling (an inherent, under-utilized capability of some modern VLMs) to enumerate relevant attributes along \emph{many} axes of diversity, and (ii) a prediction consolidation step, where we flexibly attend only to subpopulations (i.e., instances within a class sharing an attribute) that are most relevant to the image. 
By enriching and carefully consolidating attributes to describe diversity within a class, our method more faithfully encodes atypical instances. Furthermore, by introducing interpretable intermediate outputs (i.e. the inferred attributes), our method affords greater transparency, as each inference comes with the specific list of fine-grained attributes used to predict the class, and attribute overlaps across classes can help anticipate and articular potential failures, before they happen. 

In experiments over a large suite of datasets encompassing hierarchies, diverse object states, and real-world geographic diversity, we observe our method matches and in most cases exceeds the performance of existing methods, showing that transparency can be achieved without compromising on performance. Our method yields consistent gains on a second dataset suite with finer-grained classes and no labeled diversity, showing that our method still works well when intra-class diversity may be less present. Encouragingly, we find larger improvements occurring for the hardest classes and subpopulations, where atypical instances are usually found, resulting in reduced performance disparities. 
Compared to existing methods, we find that our approach can effectively scale to a much larger number of attributes to cover broader axes of diversity, as shown in the right panel of Figure~\ref{fig:fig1}. Our method also offers a principled trade-off between accuracy overall vs.~on the worst classes, all without additional training. In summary, we (i) identify a limitation of the one-vector-per-class paradigm in adequately representing classes with diverse subpopulations, (ii) propose to go beyond one vector per class, leveraging under-utilized abilities of VLMs to explicitly account for intra-class diversity, and (iii) extensively validate the effectiveness of our method to perform zero-shot classification in \emph{both} a more transparent and accurate way, especially for diverse subpopulations that are often overlooked.

\section{Review of Literature}

\looseness=-1
Despite impressive overall accuracy, modern classifiers still suffer from biases. That is, they under-perform on some parts of the data, often due to spurious correlations or data imbalances in the training set. These biases can result in significant negative real-world impact. For example, \citet{gendershades} exposed significant bias along demographic lines for facial recognition systems, and more recently, \citet{richards2023does} demonstrated that despite steady progress on typical benchmarks, today's best models still generalize poorly to images from lower-income households and certain geographic regions. Namely, VLM-based zero-shot classifiers were shown to have even larger performance disparities across geographic and economic shifts than their supervised counterparts. 
 
However, the promise of open-world zero-shot classification rightfully draws much attention to VLMs, which operate by mapping images and text to a shared latent space. CLIP \citep{radford2021learning}, a seminal VLM, achieves this via joint contrastive training of image and text encoders on 400 million image-caption pairs. Recent models such as BLIP-2 \citep{li2023blip2} bootstrap the training of more powerful VLMs by taking larger pretrained vision and language backbones and fusing their outputs to a single space, which in turn can even be used to generate text; that is, some modern VLMs contain a fully functional LLM with (often under-utilized) generative abilities. To perform zero-shot classification with VLMs, one computes the class that has the highest cosine similarity between a test image's embedding and the embedding of a class name, often averaged over many (80 for CLIP) handcrafted prompt templates. While many efforts have improved VLM-based classification via prompt-tuning \cite{zhou2022learning,zhou2022conditional,zhu2022prompt,derakhshani2023bayesian,huang2022unsupervised,Mirza_Karlinsky_Lin_Kozinski_Possegger_Feris_Bischof_2023,Menghini_Delworth_Bach_2023}, nearly all require some labeled data. Other works focus more closely on the task of debiasing VLM-based classifiers \cite{chuang2023debiasing, dear2023, DrML2023, kim2023biastotext}, though they too utilize labeled data, placing them out-of-scope of the true zero-shot setting. 

\looseness=-1
Compared to previous classifiers, the key novelty of VLMs is their ability to encode \emph{any} text. However, standard zero-shot classifiers only embed classnames, either alone or averaged over prompts. We propose to leverage the open-vocabulary capabilities of VLMs to improve coverage of intra-class diversity by embedding more than just the class name. One effort along these lines is PerceptionCLIP \citep{an2023context}, which infers contextual attributes per image as generative factors and does class inference conditioned on them. Other works utilize LLM-generated class descriptors, towards creating a concept-bottleneck \citep{Yang_2023_CVPR} or rationales for inference \citep{feng2023leveraging}, though these methods use data to train a linear layer atop descriptor similarities. DCLIP \citep{menon2022visual} show including descriptors can also improve performance in the zero-shot setting, and \citet{pratt2023what} extend the gains using additional handcrafted queries. WaffleCLIP \citep{roth2023waffling} shows that appending random characters or words can achieve similar performance to descriptor-based methods like DCLIP, without the need for an external language model. Importantly, although these works obtain more than one vector per class, they ultimately average over them. Thus, decision boundaries remain linear and biases may linger, as atypical instances are still suboptimally uncovered (see Sections \ref{sec:consolidation} and \ref{sec:trade-off}). In contrast, like us, CHiLS \citep{novack2023chils} introduces a \emph{non-linearity} in three steps: they (i) define subclasses with groundtruth label hierarchies or by querying GPT-3, (ii) do zero-shot classification on this extended set of classes (subclasses) and original classes, (iii) reweight the standard zero-shot score for each class with the max score from step (ii) over subclasses within the class. However, CHiLS is designed specifically for hierarchical label sets, which limits 
the types of diversity it can capture (see Section \ref{sec:adding_in_attrs}).
\section{Motivation}

\looseness=-1
We hypothesize that the standard one-vector-per-class paradigm poses a tension for highly diverse classes.
We investigate this by measuring classification performance as a function of class diversity. 
Indeed, we find classes with higher diversity suffer worse performance under the one-vector-per-class classification paradigm.
Then we illustrate how newfound open-vocabulary capabilities of VLMs can enrich the single class vector to 
encompass diverse instances without additional training. That is, we show that incorporating attribute information can substantially improve VLM recognition of atypical subpopulations.

\subsection{A single vector inadequately represents diverse classes}
\label{sec:tension}

\looseness=-1
A standard VLM classifier is most effective when it aligns all instances of a class to their class vector (and away from vectors for other classes). 
Intuitively, aligning instances with high diversity is challenging as their image embeddings are more dispersed---and particularly tough for fixed open-vocabulary VLMs that do not benefit from knowing the specific classes of interest during their pre-training (see Appendix \ref{app-sec:unique_and_invariant}).
We see in Figure~\ref{fig:bias_examples} for example the less typical \emph{Arctic} \texttt{fox} is far harder to recognize than a typical \texttt{fox} (52.0\% versus 84.5\% accuracy).
We observed similar drops in accuracy for a \emph{deflated} \texttt{balloon} versus a regular \texttt{balloon} and an \emph{unpaved} street versus a paved one. 
To systematically quantify this tension, both for VLMs and for the one vector per class paradigm generally, we examine class accuracies on ImageNet \citep{imagenet} relative to the diversity of each class across several models with varying levels of supervision.
To proxy diversity, we measure the variance of image embeddings within a class. 
In all cases, we observe a strong negative correlation between class-wise accuracy and diversity (see Table \ref{tab:diversity_and_acc_imagenet} and details in Appendix \ref{app-sec:tension}). That is, \textbf{classes with higher diversity have lower accuracy in the one vector per class paradigm}.


\begin{figure}
    \centering
    \begin{subfigure}{0.28\linewidth}
    \centering
    \includegraphics[width=\textwidth]{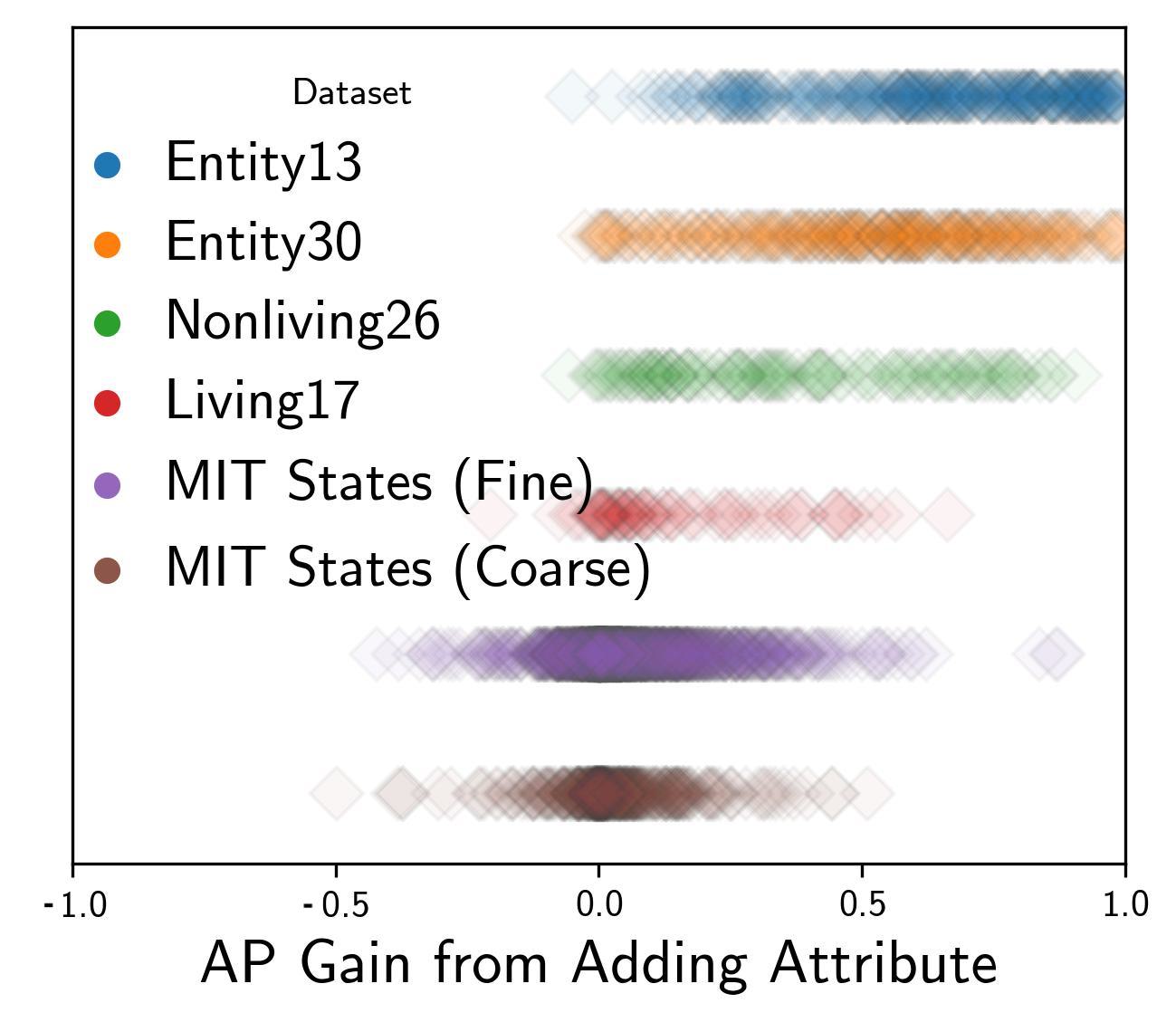}        
    \end{subfigure}
        \begin{subfigure}{0.2\linewidth}
    \centering
    \includegraphics[width=\textwidth]{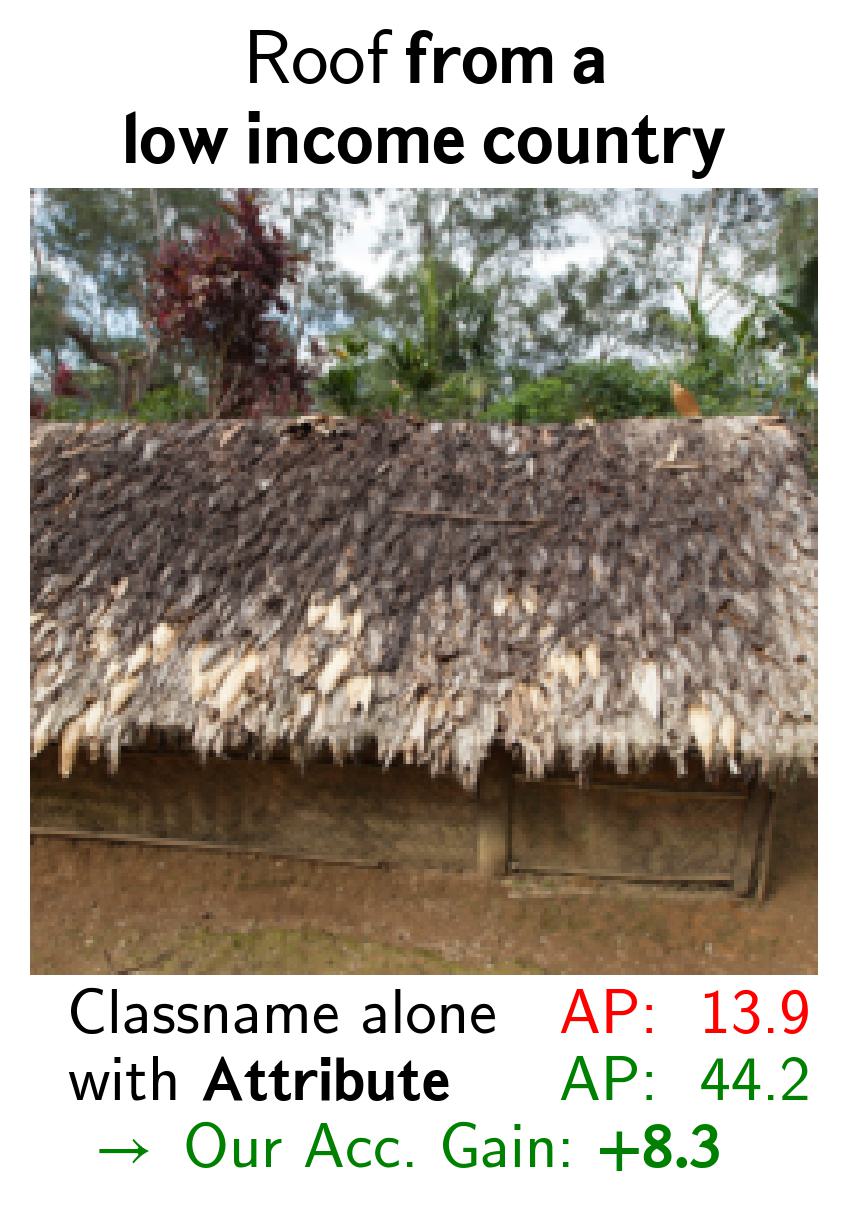}
    \end{subfigure}
        \begin{subfigure}{0.2\linewidth}
    \centering
    \includegraphics[width=\textwidth]{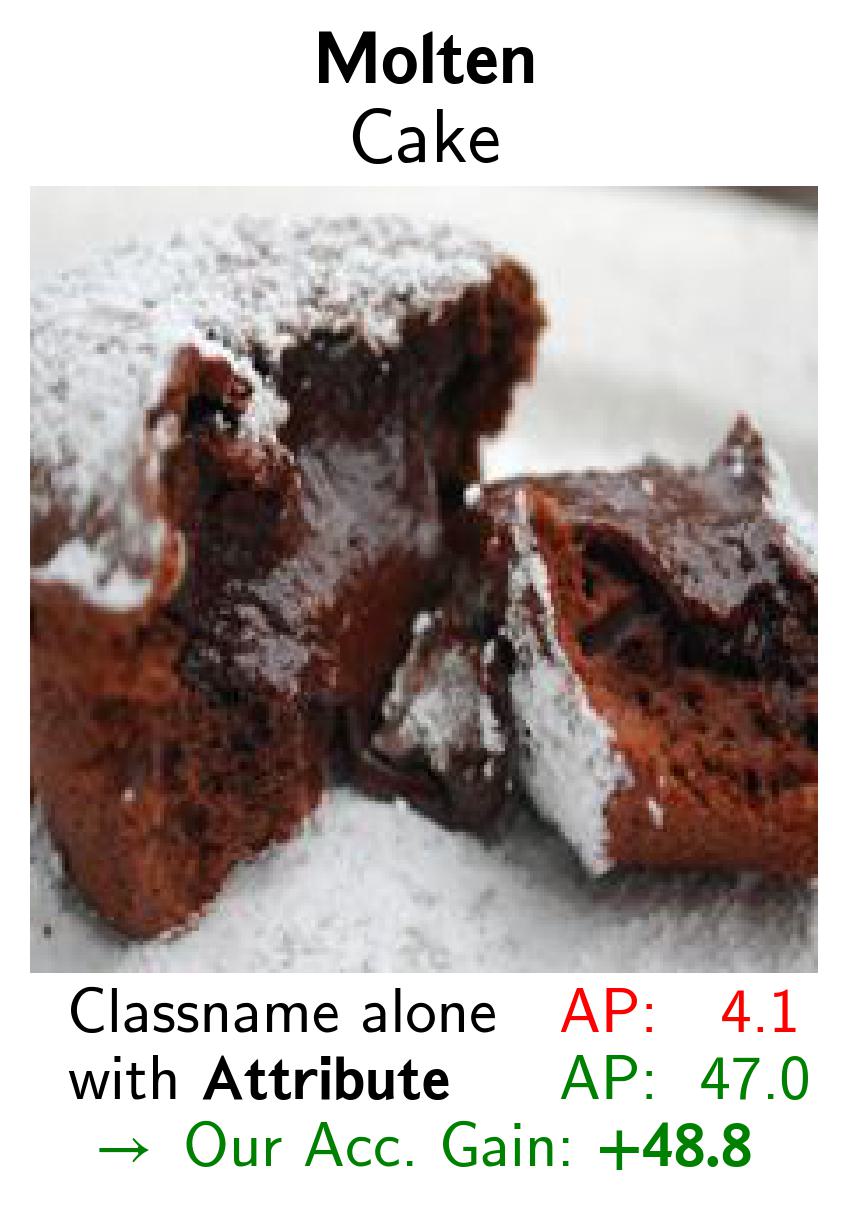}
    \end{subfigure}
    \begin{subfigure}{0.2\linewidth}
    \centering
    \includegraphics[width=\textwidth]{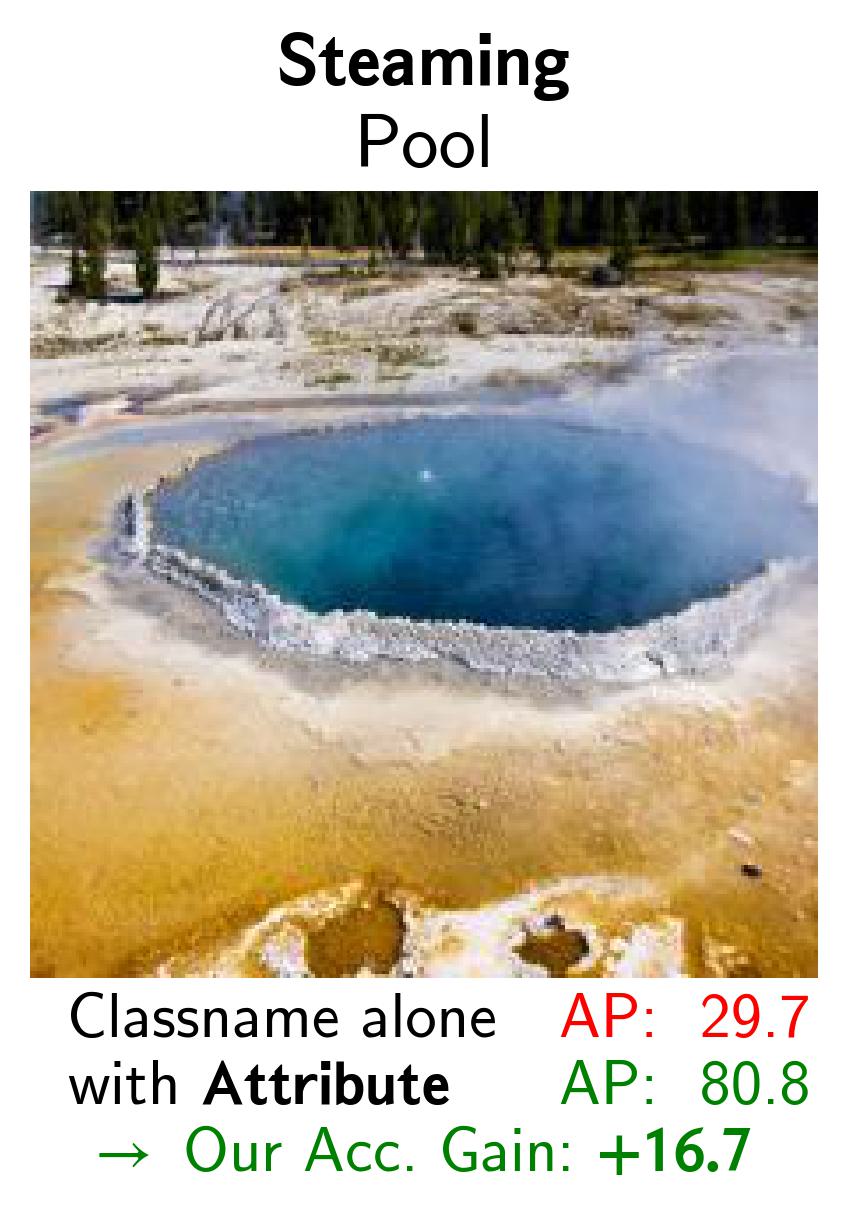}
    \end{subfigure}
    \caption{The average precision (AP) of a classname embedding is often much lower than the average precision of a subpopulation (i.e. classname with attribute) embedding. Subpopulations that see large increases in AP by including the attribute tend to be atypical. We design our method to improve accuracy on these diverse subpopulations, by inferring and explicitly accounting for them.}
    \label{fig:ap_gains}
\end{figure}

\subsection{A path forward: VLMs can recognize diversity with relevant attributes}
\label{sec:vlm_ap_gains}

Although standard VLMs use solely classname in zero-shot classification, their shared embedding space allows to encode relations to any other text. 
In turn we ask:
\textit{can the open-vocabulary encoder of VLMs better situate diverse classes given relevant attributes?}
Specifically, we assess whether enriching classes with attributes can improve zero-shot classification on a suite of datasets with ground-truth attributes per class 
(details in Appendix \ref{app-sec:ap_gains}).
We form a \textbf{subpopulation} by taking instances within a class that share an attribute. 
For each subpopulation, we compute the similarity of image embeddings with the text embedding of (i) the classname and (ii) the classname with the corresponding attribute, using CLIP ViT-B/16. 
We then measure the average precision of the two similarity scores for distinguishing instances within the subpopulation from instances outside of the class.
We find, as shown in Figure \ref{fig:ap_gains}, that for the vast majority of cases, incorporating attributes leads to more precise recognition, and often by large margins: 
adding \texttt{molten} to \texttt{cake} improves average precision by over 40 points. Upon inspection, the highest gains in average precision tend to occur for atypical subpopulations (see Appendix \ref{app-sec:ap_gains}). Thus, VLMs \emph{can} recognize instances in a class even when they are atypical, but this ability is restricted under the one vector per class paradigm. 


\section{Method}
\label{sec:method}



\looseness=-1
We now propose a method to better utilize the ability of VLMs to recognize diverse subpopulations. Our method consists of \textbf{attribute inference} and \textbf{prediction consolidation}. First, we query a large language model (LLM) for diverse per-class attributes that span many (often overlapping) subpopulations. Then, after computing the similarity of an image to each subpopulation, we non-linearly consolidate these similarities to obtain one score per class. We elaborate on these two steps below.

\subsection{Attribute Inference Along Many Axes of Diversity}
\label{sec:diversity_many_forms}

\looseness=-1
To better cover the diverse subpopulations that may exist within a class, we incorporate attribute information. However, \textbf{diversity can come in many forms}. That is, the way in which two instances of a class differ can itself vary. Consider the examples in Figure \ref{fig:bias_examples}. The \emph{Arctic} \texttt{fox} case shows how a class can contain distinct finer-grained categories. In a related manner, the state or condition in which the class instance is in can also substantially change its appearance: a \texttt{balloon} looks much different when it is \emph{deflated}. Further, there exist generic attributes that can lead to substantial visual differences regardless of the class, such as the region or income level of the country where an image is taken, exemplified by the two \texttt{Street} \texttt{View} images. Thus, to capture the many ways in which diversity can arise, we employ multiple distinct queries, in contrast to prior work. Namely, we infer:

\begin{itemize}
\item \emph{Class specific} attributes, such as the possible \textbf{states} of an object (e.g., \emph{diced} or \emph{sliced} for \texttt{pear}). We also obtain \textbf{descriptions} for and different \textbf{kinds} of each class, as in DCLIP and CHiLS respectively.
\item \emph{Class adjacent} attributes, like \textbf{co-occurring objects} or \textbf{backgrounds}, to get useful context.
\item \emph{Class agnostic} attributes that describe how objects vary in general. For example, towards improving geographic fairness, we list potential choices for the \textbf{income-level, region} and \textbf{country} of origin of the image. We also introduce a novel two-step LLM query, where we first ask the LLM to list generic axes of diversity, and then have it populate those axes. We name this \textbf{auto-global} as it automatically generates many global attributes.
\end{itemize}

Appendix \ref{app-sec:attr_inference} contains the exact LLM prompts and example inferred attributes for each query above.

\subsection{Nonlinear Prediction Consolidation}
\label{sec:consolidation}

\looseness=-1
Enumerating attributes along various axes of variation results in descriptions of many diverse subpopulations per class. Since VLMs have open-vocabulary text encoders, we can directly embed these subpopulation descriptions, in addition to the class name. Given a test image, we compute similarities to each of these embeddings. We then must consolidate them to obtain a single score per class. 

\begin{figure}
    \centering
    \includegraphics[width=0.75\linewidth]{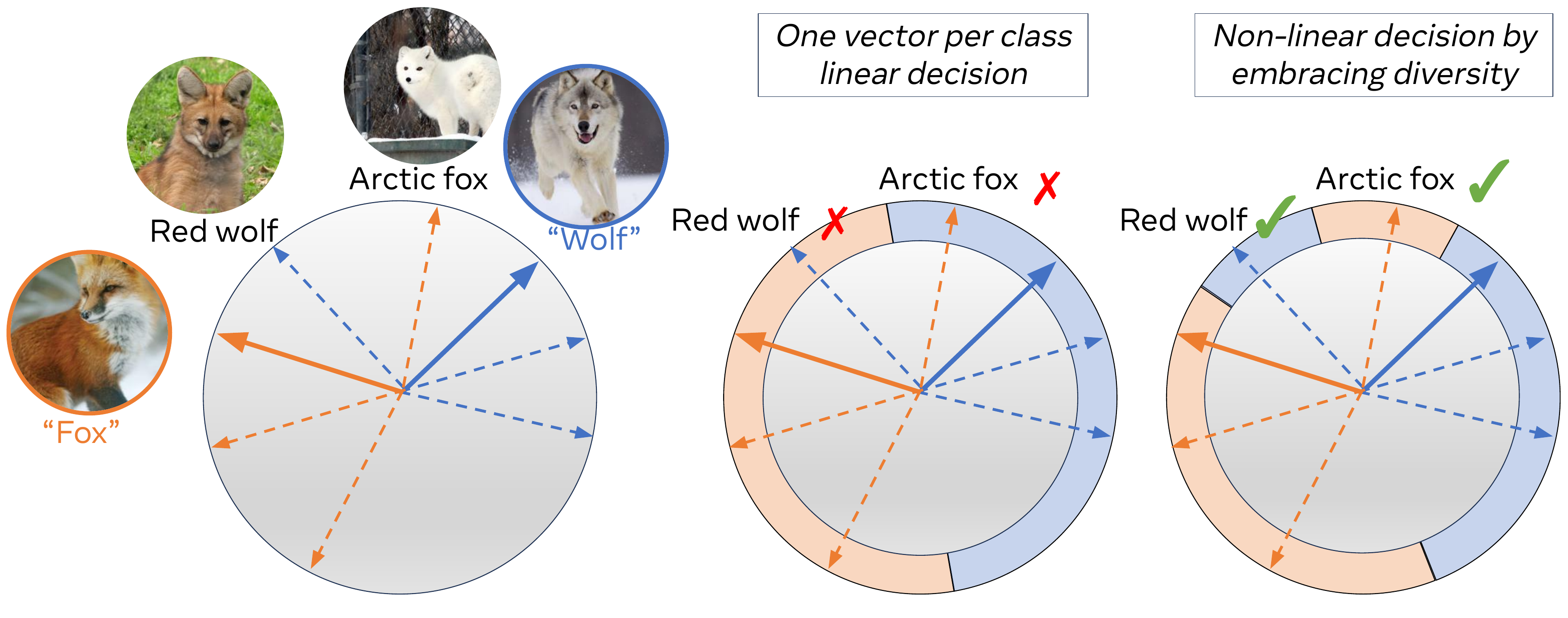}
    \caption{An \emph{Arctic} \texttt{fox} can more closely resemble a typical \texttt{wolf} than a typical \texttt{fox}. Standard zero-shot classification using one vector per class (the classname embedding) is ill suited for this case. We address this issue by nonlinearly consolidating similarities to \emph{multiple} vectors per class that explicitly encode the diverse subpopulations within the class. See section \ref{sec:consolidation} for full explanation.}
    \label{fig:fig_method}
\end{figure}

Figure \ref{fig:fig_method} illustrates the simple case of \texttt{fox} vs \texttt{wolf} classification, where solid/dotted lines correspond to classname/subpopulation embeddings on the hypersphere (shown here in 2D). The left-most panel shows examples from the two classes near where their image embeddings would lie. Text embeddings for the subpopulations (dotted lines) are close to corresponding image embeddings, as VLMs are capable of recognizing even diverse subpopulations (see Section \ref{sec:vlm_ap_gains}). Standard zero-shot inference maps a test-time image to the class of the nearest classname text embedding. Since there is only one vector per class (the classname-based embedding), the decision boundary is linear, as shown in the middle panel. The edge of the hypersphere is colored (orange for wolf, blue for fox) to indicate the predicted class for an image embedding at that location. Notably, the \emph{Arctic} \texttt{fox} is misclassified as \texttt{wolf}, as its appearance more closely resembles a typical wolf than a typical \texttt{fox}, and so, the embeddings of \emph{Arctic} \texttt{fox} images fall closer to the text embedding of ``wolf'' (and vice-versa for the \emph{red} \texttt{wolf}). Methods like DCLIP and WaffleCLIP embed more than just the classname, but they consolidate similarities via averaging, again resulting in a linear decision boundary. Even if atypical subpopulations are included at first, averaging can narrow the initial diverse coverage, as most embeddings for a class may better describe a typical instance. 

\looseness=-1
In contrast, we propose the following \emph{nonlinear} consolidation: we compute the single score per class for a given test image as the average of the similarities of the image embedding to \emph{only the} $k$ \emph{closest} subpopulations embeddings for the class, where $k$ is typically small (we use $k=16$). This way, an image can have a high class score even if it is only similar to a small subset of subpopulations, as is the case for atypical instances. Thus, the \emph{Arctic} \texttt{fox} and \emph{red} \texttt{wolf} can be correctly classified despite being far from the classname and most subpopulation embeddings for their respective classes, as shown on the right panel of Figure \ref{fig:fig_method}, where we use $k=1$ for simplicity (i.e. images are mapped to the class of the closest dotted or solid line, leading to a non-linear boundary). We shed insight on the effect of varying the hyperparameter $k$ in Section \ref{sec:trade-off}, revealing a tunable accuracy-fairness trade-off.

\section{Analysis}

We now empirically validate our method's effectiveness and enhanced interpretability over two dataset suites. Our method performs on par with (and usually surpasses) existing approaches in overall accuracy. Notably, we see larger gains for the hardest classes and subpopulations, which are likely more diverse and atypical respectively (precisely the samples on which our method is intended to improve performance). Furthermore, while matching or exceeds the performance of existing, we offer unique interpretability benefits, such as fine-grained and faithful explanations, as well as the potential for \emph{error anticipation}; we detail these below.


\subsection{Consistent Gains Across Diverse Datasets, Particularly for the Hardest Classes and Subpopulations}
\label{sec:experiments}

\begin{table}
\centering
\resizebox{0.8\textwidth}{!}{%
\begin{tabular}{llcccc}
\toprule
 &  & Accuracy & Avg Worst & Worst 20\% of & Worst 20\% of \\
Dataset Type &  &  & Subpop & Classes & Subpops \\
\midrule
\multirow[t]{5}{*}{States} & Vanilla & \underline{66.71} & 40.66 & 35.46 & 21.73 \\
 & DCLIP & 63.65 & 39.41 & 34.26 & 20.98 \\
 & Waffle & 66.68 & \underline{40.71} & 35.49 & 22.05 \\
 & CHiLS & 66.56 & 40.41 & \underline{36.16} & \underline{22.45} \\
 & Ours & \textbf{67.92} & \textbf{41.53} & \textbf{38.16} & \textbf{23.64} \\
\midrule
\multirow[t]{5}{*}{Hierarchical} & Vanilla & 78.15 & 48.36 & 50.72 & 35.89 \\
 & DCLIP & 77.80 & 48.48 & 51.05 & 34.36 \\
 & Waffle & 78.52 & 49.42 & 49.78 & 35.22 \\
 & CHiLS & \underline{79.44} & \textbf{52.65} & \underline{51.80} & \underline{38.44} \\
 & Ours & \textbf{79.50} & \underline{51.23} & \textbf{52.59}
 & \textbf{38.57} \\
\bottomrule
\end{tabular}%
}
\caption{Zero-shot classification on datasets with known variation types for CLIP with a ViT-B/16 encoder. States averages results over the two categorizations of MIT States data, while Hierarchical averages results over four Breeds datasets. We observe similar results for BLIP-2 (Table \ref{app-tab:known_variation_blip2}).}
\label{tab:known_variation}
\end{table}

\subsubsection{Baselines and Metrics}
\looseness=-1 
We measure performance of zero-shot classifiers using the popular CLIP ViT-B/16 and BLIP-2 VLMs \citep{radford2021learning, li2023blip2}. To infer attributes, we utilize the open source Vicuna-13b-v1.5 language model \citep{vicuna2023}, which notably is already contained in the BLIP-2 model we use. We report accuracy overall as well as averaged over the worst $20\%$ of classes and subpopulations. Note that we only use groundtruth attributes when computing metrics; our method exclusively uses attributes \emph{inferred} via the queries listed in Section \ref{sec:diversity_many_forms}. We also compute the lowest subpopulation accuracy per class and average that, so to obtain the metric denoted as `Avg Worst Subpop'. For the real-world shifts, we also report worst region and worst income group accuracy. Our baselines include: standard zero-shot (only one vector per class, corresponding to the classname embedding) which we call Vanilla, DCLIP (averages over class descriptors), WaffleCLIP (averages over \emph{random} descriptors sampled over ten trials), and CHiLS (reweights standard zero-shot class score with \emph{max} probability of different kinds of the class). Notably, we average all text embeddings over the $80$ prompts crafted for CLIP, so to report best possible baseline results.

\subsubsection{Datasets}

\looseness=-1
We curate a suite of eight \emph{attributed} (so to have groundtruth subpopulations) datasets spanning different axes of diversity. We use the four Breeds datasets \citep{Santurkar2020BREEDSBF} for their hierarchical label sets, as used in the CHiLS paper; in fact, the Breeds datasets were the ones where CHiLS was most effective. Next, we devise two classification tasks (coarse and fine grained) from the MIT States dataset \citep{Isola2015DiscoveringSA} to track performance over labeled states (e.g., \emph{sliced} or \emph{diced} for \texttt{pear}). Importantly, we also include the datasets Dollarstreet \citep{rojas2022the} and GeoDE \citep{ramaswamy2022geode}, which contain images from varied geographic regions and income levels. As the diversity in these datasets is occurs naturally, they can encompass \emph{many} axes of variation, as opposed to our other datasets that only varying along known axes, like object state or kind. 

We also incorporate a second suite of the following $9$ datasets \emph{without} attributes: ImageNet \cite{imagenet}, ImageNet variants (v2, -R, -A, -Sketch) \cite{imagenetv2, inetR, inetA, inetSketch}, Food-101 \cite{food101}, Flowers-102 \cite{flowers102}, FGVC-Aircraft \cite{fgvc}, and Oxford Pets \cite{pets}. These datasets are all somewhat fine-grained\footnote{ImageNet (and its variants) contains 120 out of 1000 categories dedicated only to various dog breeds; Flowers-102 has many highly similar classes; Oxford Pets only has different breeds of dogs and cats; the FG in FGVC-Aircraft stands for fine-grained.}, and as such, are less likely to have intra-class diversity than our original datasets. Thus, this dataset suite provides insight on if embracing diversity is only effective when diversity is to be expected.


\looseness=-1
We note that we strived to minimally fit our method to the evaluation suite. That is, we do not optimize our query set to maximize performance on the datasets we selected, which can be challenging for zero-shot classification methods. One specific measure we took toward this end was fixing our method completely before evaluating on the second dataset suite. Thus, the second dataset suite serves as a held-out challenge set, intended to test the generalizability our method to settings where intra-class diversity may not be present. See Appendix~\ref{app-sec:datasets} for complete details on our dataset suite.

\begin{table}[]
\centering
\resizebox{0.875\textwidth}{!}{%
\begin{tabular}{@{}lcccccc@{}}
\toprule
\textit{DollarStreet} & Accuracy   & \multicolumn{1}{c}{Worst}             & \multicolumn{1}{c}{Worst}             & \multicolumn{1}{c}{Avg Worst} & \multicolumn{1}{c}{ Worst 20\% of} & \multicolumn{1}{c}{Worst 20\% of}                    \\
Method &  &  Region & Income & Subpop & Classes & Subpops \\
\midrule
Vanilla & 51.51 & 42.43 & 34.76 & 37.60 & 18.33 & 11.01 \\
DCLIP & 49.78 & 41.08 & 32.91 & 36.37 & 19.07 & 11.19 \\
Waffle & 51.37 & \underline{42.71} & \underline{34.97} & \underline{37.69} & 18.12 & 10.74 \\
CHiLS & \underline{51.68} & 42.20 & 33.90 & 37.60 & \underline{20.51} & \underline{12.72} \\
Ours & \textbf{52.70} & \textbf{44.04} & \textbf{37.21} & \textbf{40.31} & \textbf{20.88} & \textbf{15.05} \\
\midrule
\textit{GeoDE} &                &                &                &                \\ \midrule
Vanilla & 90.15 & 86.63 & - & 82.57 & 72.24 & 69.95 \\
DCLIP & 91.31 & 88.14 & - & 84.21 & 74.44 & 71.90 \\
Waffle & \underline{91.59} & \underline{89.06} & - & \textbf{85.44} & \underline{75.85} & \underline{74.37} \\
CHiLS & 90.96 & 87.90 & - & 84.48 & 73.27 & 71.64 \\
Ours & \textbf{91.75} & \textbf{89.12} & - & \underline{85.40} & \textbf{76.13} & \textbf{74.64} \\ \bottomrule
\end{tabular}%
}
\caption{Zero-shot classification performance on geographically diverse images from DollarStreet and GeoDE using CLIP with a ViT-B/16 encoder. We observe similar results for BLIP-2 (Table \ref{app-tab:geo}).}
\label{tab:geo}
\end{table}

\begin{table}
\centering
\begin{tabular}{lrrrrrrrrrrr}\toprule
&ImageNet &v2 &-A &-R &Sketch &Food &Flowers &Aircraft &Pets & Average \\\midrule
\multicolumn{11}{c}{Overall Accuracy} \\
\midrule
Vanilla &68.48 &61.98 &30.16 &59.24 &48.37 &88.35 &66.09 &31.26 &\textbf{92.72} &60.74 \\
DClip &\underline{68.85} &\underline{62.37} &\underline{31.35} &60.04 &48.54 &88.05 &\textbf{70.69} &32.67 &92.23 & \underline{61.64} \\
Waffle &68.44 &62.15 &31.09 &\underline{61.17} &\underline{48.58} &88.09 &66.87 &31.05 &92.03 &61.05 \\
CHiLS &0.11 &0.10 &0.00 &0.00 &0.11 &\underline{88.59} &67.49 &34.20 &92.12 &31.41 \\
Ours &\textbf{69.94} &\textbf{63.32} &\textbf{32.19} &\textbf{61.49} &\textbf{49.38} &\textbf{89.06} &\textbf{70.69} &\textbf{34.62} &\underline{92.26} & \textbf{62.55} \\
\midrule
\multicolumn{11}{c}{Accuracy on Worst $20\%$ of Classes} \\
\midrule
Vanilla &34.42 &\underline{25.75} &\underline{7.47} &\underline{29.27} &\underline{9.19} &73.58 &2.08 &0.00 &\textbf{78.84} &\underline{28.96} \\
DClip &\underline{34.58} &25.65 &6.24 &28.90 &8.99 &72.92 &\underline{3.44} &0.00 &\underline{77.50} &28.69 \\
Waffle &32.62 &23.00 &5.79 &28.18 &8.93 &73.36 &2.14 &0.00 &75.65 &27.74 \\
CHiLS &0.00 &0.00 &0.00 &0.00 &0.00 &\underline{75.36} &2.09 &0.00 &76.92 &17.15 \\
Ours &\textbf{37.30} &\textbf{27.60} &\textbf{7.98} &\textbf{33.74} &\textbf{9.33} &\textbf{76.18} &\textbf{4.40} &\textbf{0.25} &77.40 &\textbf{30.46} \\
\bottomrule
\end{tabular}
\caption{Zero-shot classification performance on finer-grained \textbf{held-out} datasets without attributes, using CLIP with a ViT-B/16 encoder. We observe similar results for BLIP-2 (Table \ref{tab:extra_dsets_blip}). We discuss reasons for the failure of CHiLS on ImageNet-scale tasks in Appendix \ref{app-sec:chils_clip}. Our method effectively generalizes to new settings without tuning the set of queries for attribute inference.}
\label{tab:extra_dsets_clip}
\end{table}

\subsubsection{Results}
Table \ref{tab:known_variation} shows results for datasets with diversity along hierarchical and states axes, and table \ref{tab:geo} shows results for geographic diversity. Our method consistently matches (and even improves) accuracy of existing methods, even over CHiLS in the hierarchical setting it was specifically designed for. Notably, CHiLS becomes less effective for other datasets, while our method remains strong. We observe larger gains for worst class and subpopulation metrics, especially over baselines that consolidate via averaging (Vanilla, DCLIP, Waffle), supporting the claim that our method improves coverage of the most atypical instances, and that moving beyond the one vector per class paradigm helps in this regard. For example, compared to baselines that consolidate via averaging to obtain one vector per class, our method improves accuracy for the worst classes and subpopulations by $2-3\%$ in most cases. For Dollarstreet, these gains manifest in a $9\%$ average relative gain over baselines for the accuracy over worst income group metric (and an even larger gain for the worst $20\%$ of subpopulations), showing that our methodology can facilitate progress on real-world fairness indicators.

\begin{figure}[b!]
    \centering
    \begin{minipage}{0.32\textwidth}
        \centering
        \includegraphics[width=\linewidth]{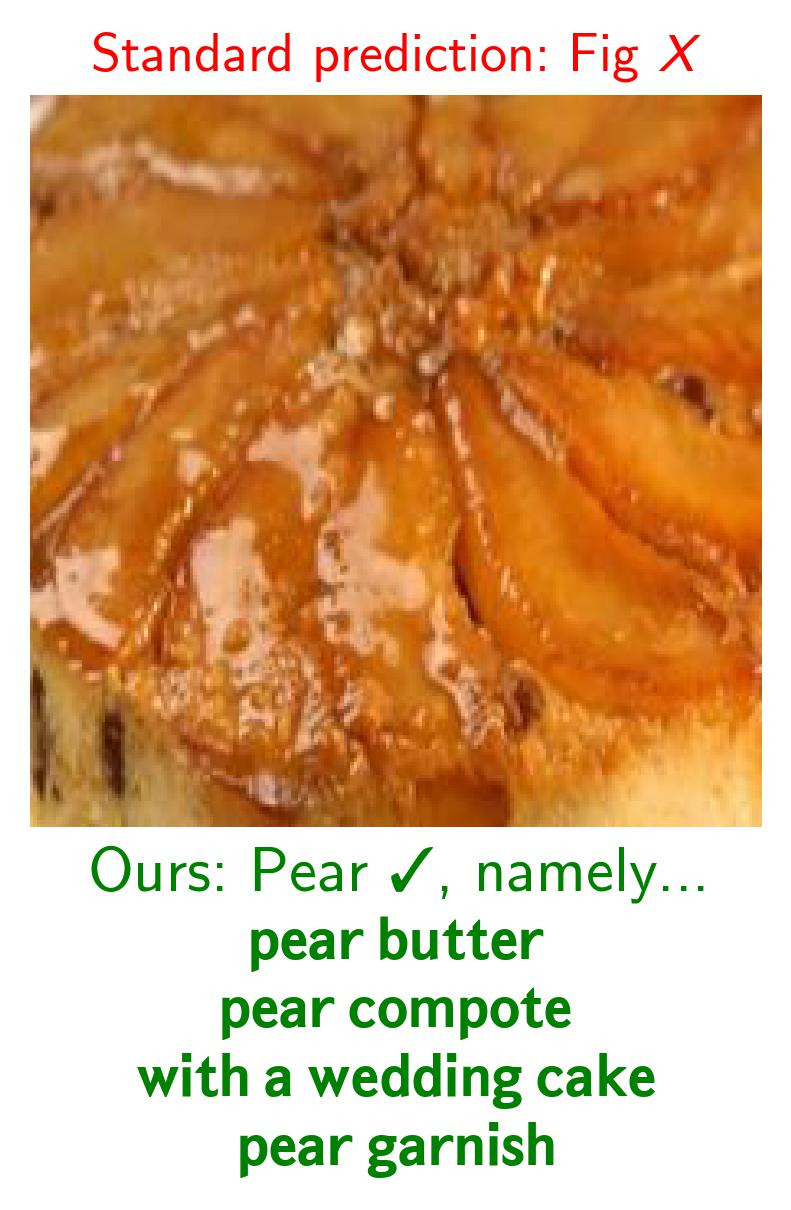}
    \end{minipage}
    \begin{minipage}{0.32\textwidth}
        \centering
        \includegraphics[width=\linewidth]{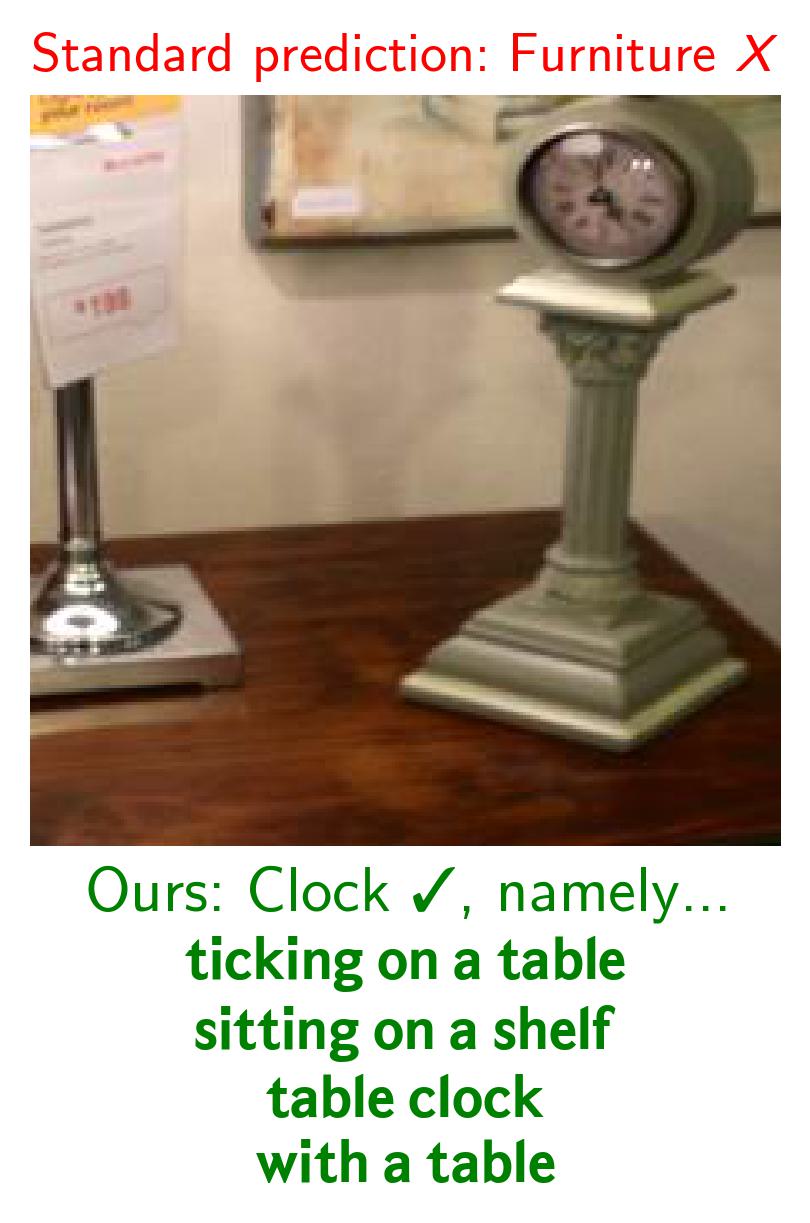}
    \end{minipage}
    \begin{minipage}{0.32\textwidth}
        \centering
        \includegraphics[width=\linewidth]{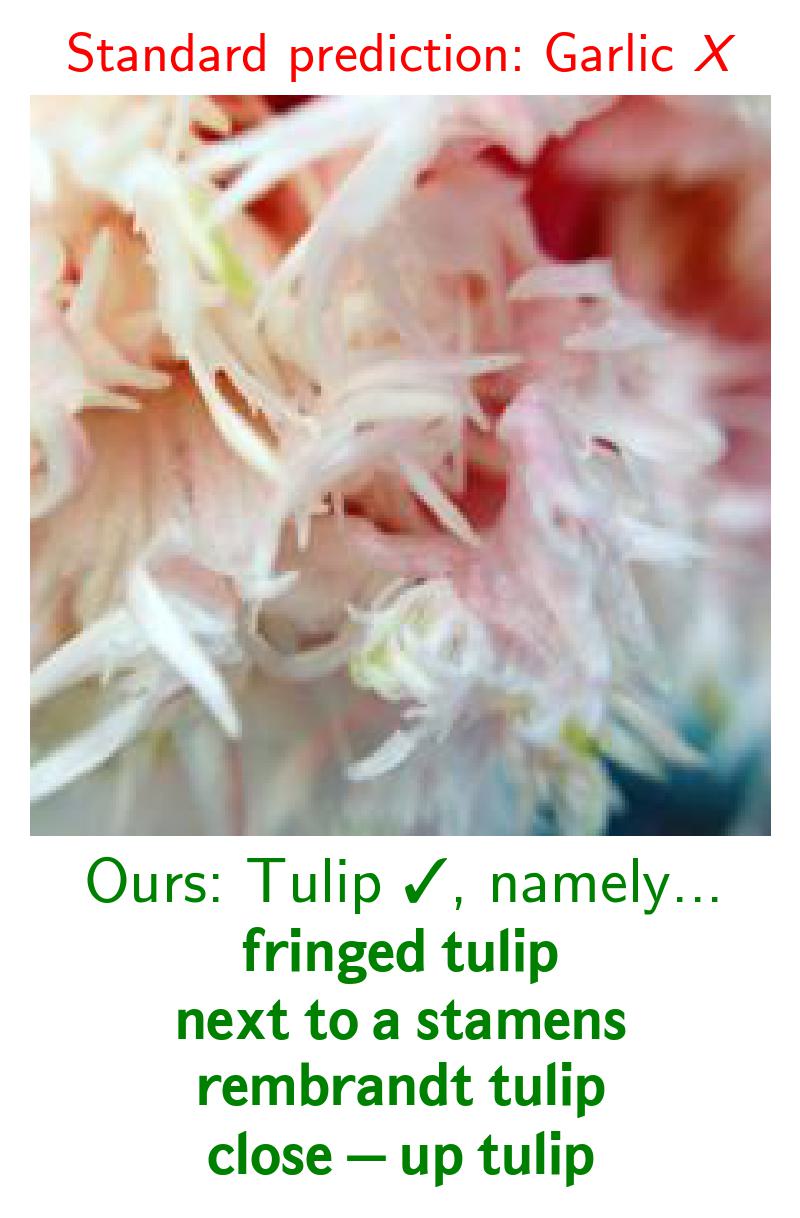}
    \end{minipage}
    \begin{minipage}{0.315\textwidth}
        \centering
        \includegraphics[width=\linewidth]{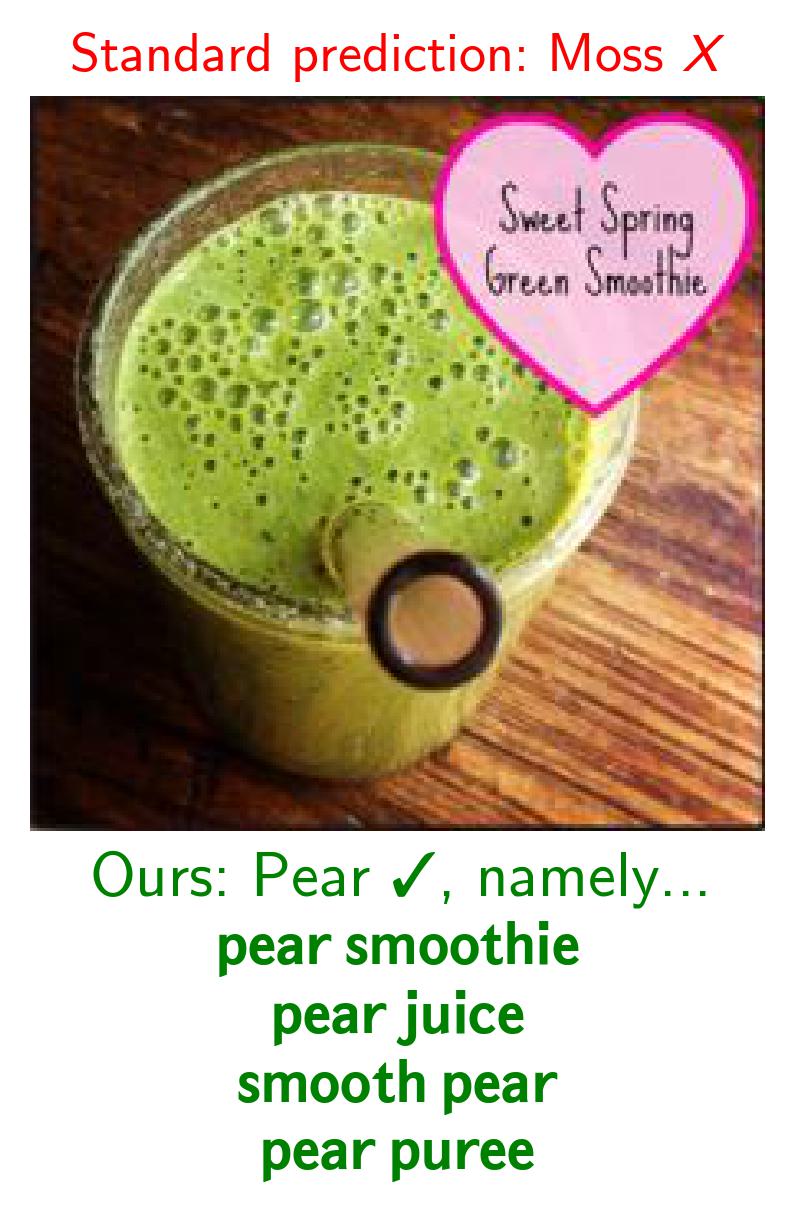}
    \end{minipage}
    \begin{minipage}{0.32\textwidth}
        \centering
        \includegraphics[width=\linewidth]{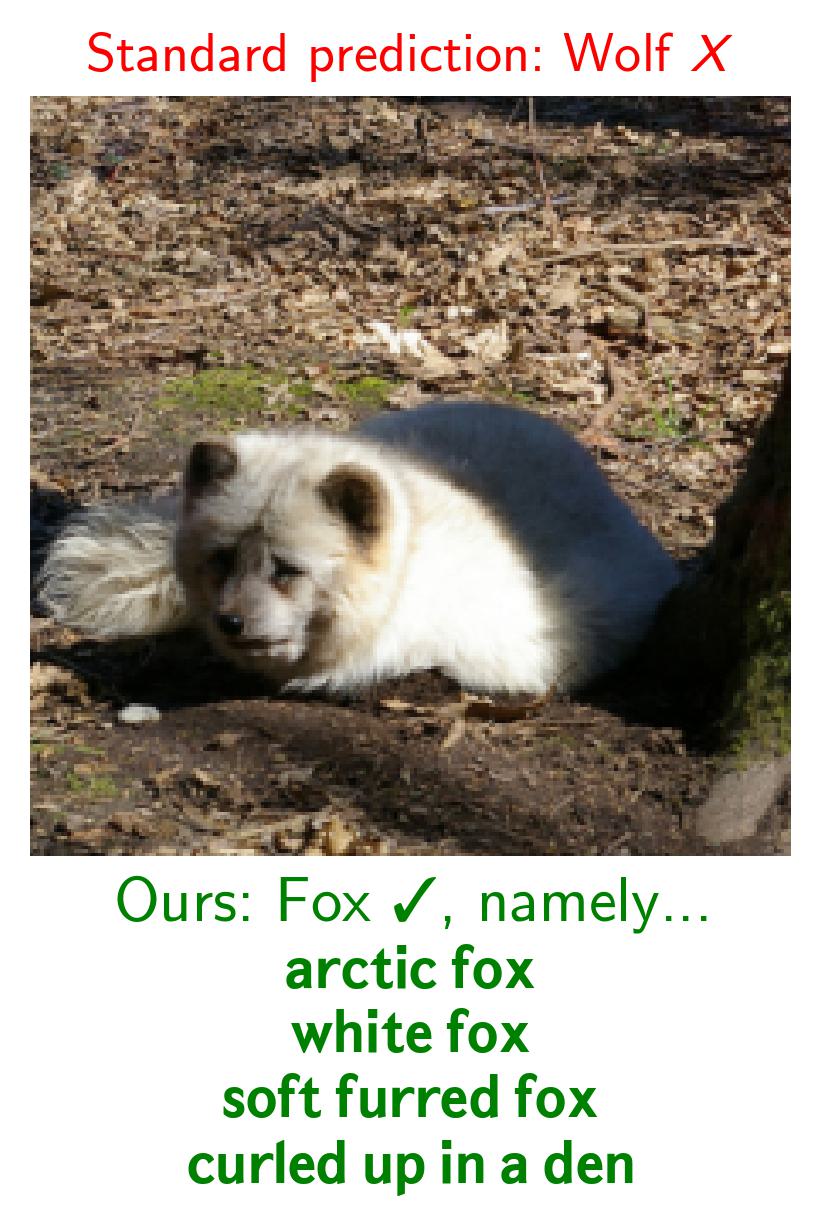}
    \end{minipage}
    \begin{minipage}{0.334\textwidth}
        \centering
        \includegraphics[width=\linewidth]{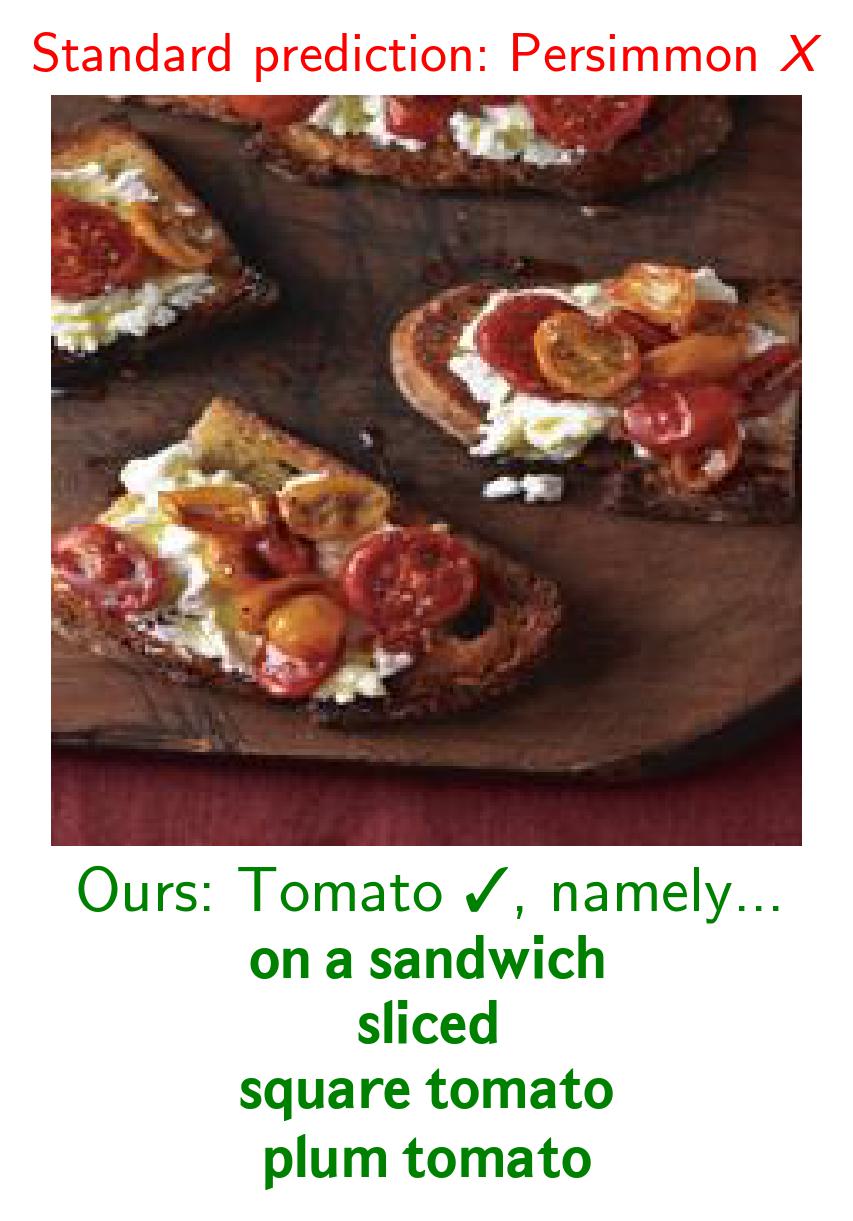}
    \end{minipage}
    \caption{Instances where our method corrects mistakes of the standard approach. The attributes used in inference also serve as faithful fine-grained explanations. Notably, these samples are atypical, suggesting that inspecting samples where our method and standard classification disagree can enable automatic surfacing of atypical cases, towards better understanding the task at hand.}
    \label{fig:eg_inferences}
\end{figure}

Turning our attention to the held-out challenge datasets, Table \ref{tab:extra_dsets_clip} shows our method can generalize effectively to finer-grained classification tasks where intra-class diversity is not explicitly known to be present. Our method improves accuracy on the hardest classes by an average of $1.5\%$ over the closest baseline. Similarly, our method exceeds all baselines by about $1\%$ in overall accuracy in nearly all cases, suggesting that embracing diversity does not come at a cost of overall performance. Moreover, the effectiveness of our method in these new settings show that the queries we select (for inferring attributes) generalize beyond our original dataset suite. That is, we do not need to tune the LLM queries for each new classification task of interest. Nonetheless, the ability to add and remove LLM queries can be seen as a strength, as a practitioner is provided more control than in standard zero-shot classification.

\subsection{Enhanced Interpretability at No Cost to Performance}

\subsubsection{Faithful Fine-grained Interpretations For Free}
Having shown that our method is equally (and usually more) performant than existing approaches, we now discuss the enhanced interpretability of our method. Namely, each inference comes with a list of the $k$ subpopulations specifically relevant to the test image for free. Figure \ref{fig:eg_inferences} shows a few example where our method corrects misclassifications from the standard approach (see Appendix \ref{app-sec:inference_egs} for more). These interpretations are faithful, as they are exactly the subpopulations used to compute the class score. Also, since we include attributes along various axes of diversity, our interpretations are finer-grained than prior work: DCLIP yields the same set general descriptors for any image predicted to a given class, and WaffleCLIP offers no interpretability at all. This interpretability can enable model debugging, as erroneous predictions can be traced back to attributes that either do not match the intended class (i.e. LLM mistake) or cannot be recognized well (i.e. VLM mistake). At a high level, while standard zero-shot classification is a complete black box, the LLM-inferred attributes of our method provide \emph{interpretable intermediate outputs}, increasing the transparency of the system overall. Moreover, our inference strategy results in concise explanations, which have more utility than explanations that are too long for a human to digest \cite{ramaswamy2023overlooked}.

\subsubsection{Anticipating and Articulating Potential Failures}
\label{sec:error_anticipation}
\looseness=-1
In addition to explaining each inference, the interpretable intermediate outputs of our pipeline also allow for \emph{error anticipation}. Namely, by comparing the inferred attributes for each class, one can anticipate and describe similar subpopulations from different classes, which may correspond to inputs where the model is less effective. For example, for the Living-17 task in the Breeds datasets, the LLM lists \emph{gibbon} as a kind of both the \texttt{ape} and \texttt{monkey} classes. While gibbons are apes, they are smaller than most apes, which makes them resemble monkeys. Indeed, standard zero-shot accuracy for gibbons is only $14\%$, where as other apes are classified at an accuracy of $93.5\%$ \footnote{We can compute this because \emph{gibbon} happened to be a groundtruth subpopulation for the \texttt{ape} class, which does not occur in the \texttt{bed} and \texttt{rug} case.}. In another case, \emph{rug} is listed as a co-occuring object for the \texttt{bed} class, when \texttt{rug} itself is another class in the dataset. While anticipating potential failure modes is intuitive for humans, it is challenging to do so \emph{at scale}. By incorporating an auxiliary model (LLM) with interpretable intermediate outputs (inferred attributes), practioners can both more easily audit and verify the zero-shot classification pipeline, and better understand potential challenges with the task of interest. We hope that the greater transparency of our system (importantly, achieved \emph{without compromising on overall accuracy}) can result in increased trust and more responsible use.

\section{Ablations}

We now detail additional ablation studies to shed insight on the source of our method's improvements over existing art and how a practitioner can apply our method with greater control. First, we study how performance varies for both our method and baselines as the number of attributes grows, so to demonstrate the value of our flexible consolidation strategy, specifically for inputs from the hardest classes. Then, we identify a principled trade-off between accuracy overall and on the worst classes using our method, controlled by the hyperparameter $k$. 

\subsection{Scaling with the Many Axes of Diversity}
\label{sec:adding_in_attrs}


\begin{figure}
    \centering
    \includegraphics[width=0.85\linewidth]{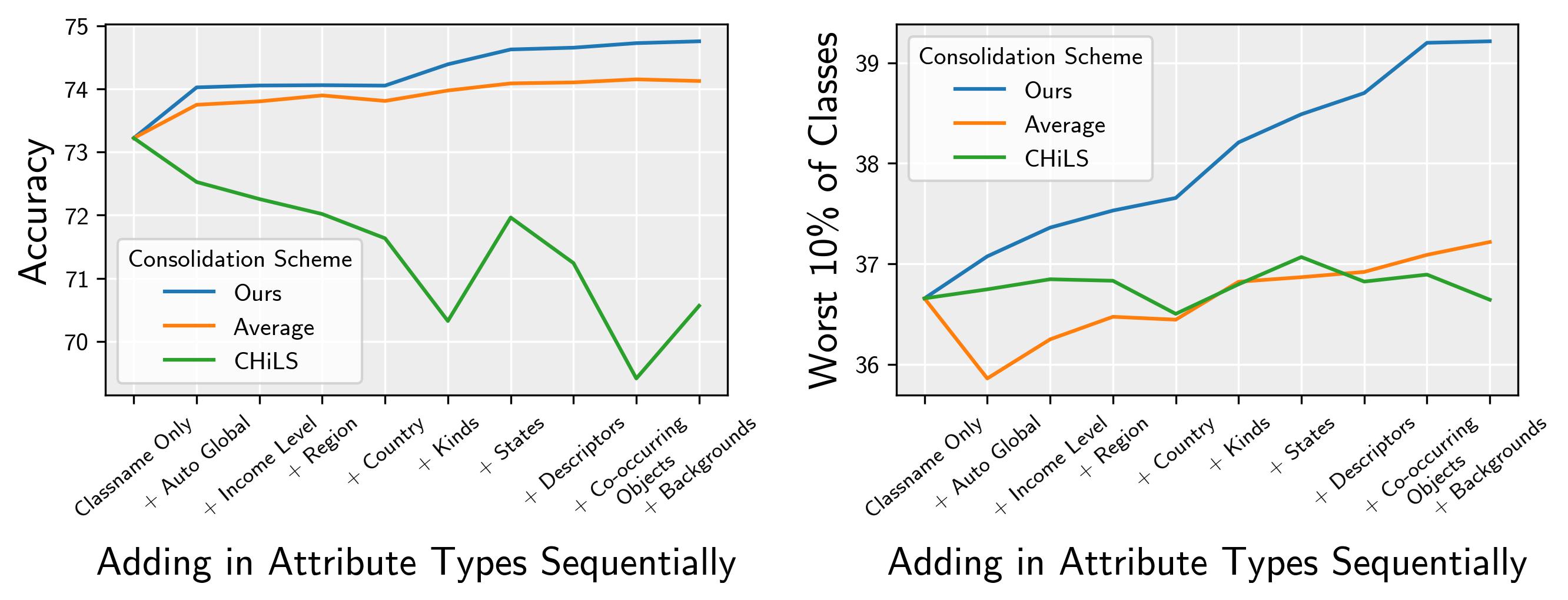}
    \caption{Accuracy, overall and for the worst classes, as new types of attributes are added. Performance for our consolidation scheme continuously improves, while it saturates or deteriorates for others. Figure \ref{app-fig:add_in_attr} shows similar trends for accuracy on the worst 20\% of classes and subpopulations.}
    \label{fig:add_in_attr}
\end{figure}

\looseness=-1
One source of gains for our method is that we infer attributes of \emph{many} types, while prior works only include one. We argue that our flexible consolidation (of subpopulation similarities to a single class score) also provides improvements over naive averaging or the nonlinear consolidation of CHiLS. To test this, we sequentially add each type of attribute, and inspect performance using the three methods. Figure \ref{fig:add_in_attr} shows our consolidation scales best as more attributes are added, with sizable gains for accuracy over the worst classes. In contrast, performance saturates with averaging, and actually deteriorates with CHiLS. The latter occurs since CHiLS assumes that subpopulations are mutually exclusive, as is the case in hierarchical label sets. When adding attributes along the many axes of diversity, resultant subpopulations overlap, making a zero-shot classification over all subpopulations (as done in CHiLS) unreliable. Averaging is also suboptimal, as the impact of each attribute diminishes as the number of attributes added increases: we see this in the left plot, as accuracy barely increases for the final three added attribute types. Also, samples that are close to only a few subpopulations but far from most (i.e., atypical instances) ultimately receive a low score when all scores are averaged. Thus, while averaging over subpopulations can improve accuracy (to an extent), it is less suited to improving performance on atypical instances than our method. We explore this further in the next section.

\subsection{Tunable Trade-off between Accuracy Overall and On Worst Classes}
\label{sec:trade-off}

\begin{figure}
    \centering
    \includegraphics[width=0.85\textwidth]{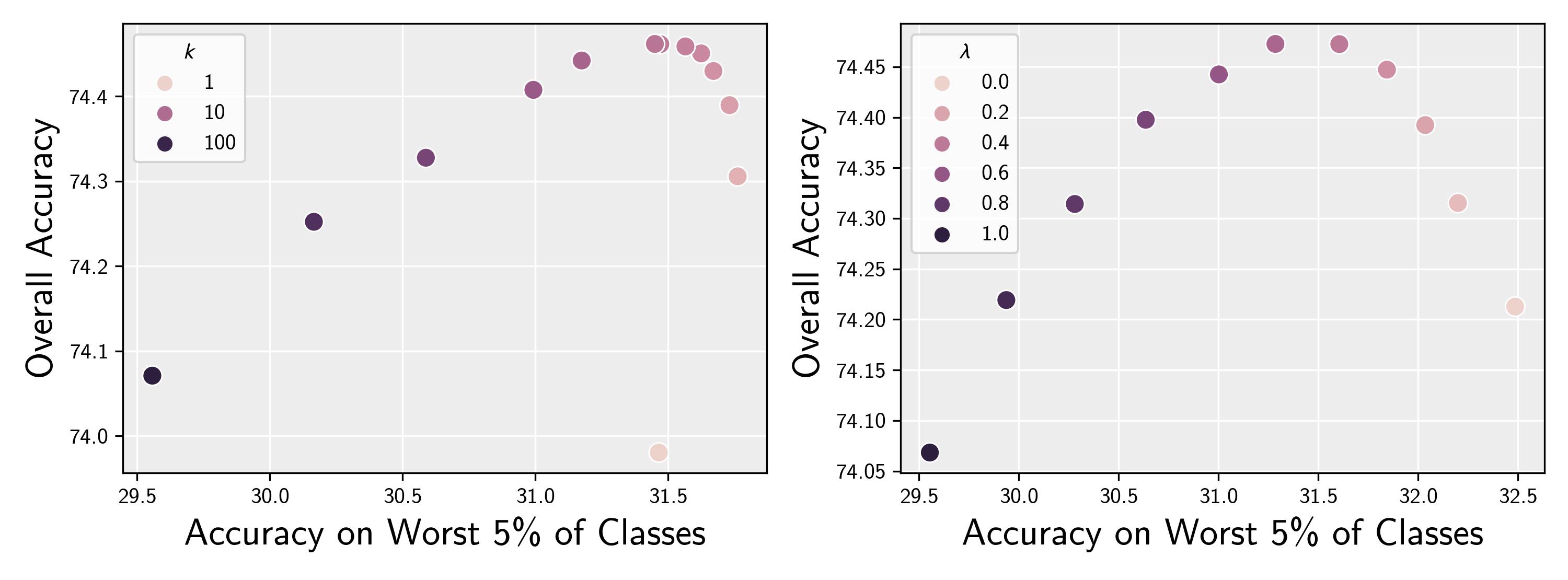}
    \caption{\textbf{Right:} As $k$ decreases, first, accuracy overall and on the worst classes both increase. Then, overall accuracy begins to decrease while accuracy on the worst classes continues to improve. Thus, we can control this trade-off via $k$. \textbf{Left:} $\lambda$, the continuous analog of $k$,  allows for greater control.}
    \label{fig:trade-off}
\end{figure}

\looseness=-1
Recall that our method consists of computing the similarity of a given test image's embedding to the embedding of numerous (on the order of hundreds) subpopulations per class, before averaging over only the top $k$ similarities, where $k$ is small. Note that when $k=\infty$, our consolidation reduces to simple averaging over all vectors per class. To shed insight on how our consolidation differs from averaging, we sweep $k$, while keeping our attribute inference fixed. Additionally, we explore linearly interpolating class scores using our consolidation (top-$k$) and full averaging via a second hyperparameter $\lambda$, so that $\lambda=0$ results in our method and $\lambda=1$ is averaging. We jointly sweep $\lambda$ from $0$ to $1$ and $k$ from $1$ to $128$ to pinpoint the way in which our consolidation improves upon averaging. 


\looseness=-1
Figure \ref{fig:trade-off} shows overall accuracy vs.~accuracy on the worst $5\%$ of classes\footnote{We observe the same trade-off when inspecting the worst $10\%$ and $20\%$ of classes. See Appendix \ref{app-fig:add_in_attr}.} for both $k$ and $\lambda$. The trend is identical for the two parameters: first, both accuracy metrics increase as we move away from full averaging, with much larger gains occurring for the worst classes. Then, accuracy begins to drop, while accuracy on the worst classes continues to improve. To understand this trade-off, consider an instance that has high similarity to one subpopulation embedding for a class, and low similarity to all others. In the $k=1$ case, this instance is given a high score for the class. This can benefit atypical instances of the class, as they may be visually dissimilar from most other instances (recall the \emph{Arctic} \texttt{fox}). However, this can introduce errors, as the correct prediction for an instance mostly close to embeddings from its true class can be flipped with the presence of just one highly similar (perhaps unreliable) subpopulation embedding from a different class. Thus, lower choices of $k$ may benefit more atypical instances, leading to improved accuracy on worst classes 
(which are most diverse; see \ref{sec:tension}), potentially at the cost of overall accuracy. With this insight, practitioners can choose how to tune our method based on their end goals. Also, since $\lambda$ is continuous, it offers closer control of this tradeoff: indeed, accuracy on the worst classes can be improved by a larger margin when varying $\lambda$, and varying $k$ and $\lambda$ together can lead to best numbers for both metrics.\footnote{To be true to the zero-shot setting, no tuning was done to obtain the results in \ref{sec:experiments}. We tried two reasonably small values for $k$ (8 and 16), observed similar results, and went with $k=16$, which was marginally better.}

\section{Limitations}

\textbf{On utilizing auxiliary models.} Our method adds an LLM into the zero-shot classification pipeline, which can increase computational cost and introduce a source of error. We note that the added compute for inferring attributes is only done once per task, so the asymptotic cost per inference only differs marginally (due to computing similarities to more vectors per class, which is a very fast operation) compared to the standard approach. To inspect the reliability of LLM outputs, we manually verify $300$ randomly selected LLM outputs. We find only $2.7\%$ of responses are uninformative\footnote{All of these were found in responses to the `descriptors' query, with responses like `size' or `shape' for the class \texttt{dog}.}, and none of the $300$ to be inaccurate. Moreover, our flexible consolidation scheme offers a kind of robustness to irrelevant LLM outputs: Recall, only the similarity to a small number of subpopulations per class contribute to each logit. Thus, irrelevant (i.e. not appearing in the data) subpopulations are effectively ignored and do not affect the logit. However, we still know that LLMs are capable of providing inaccurate outputs, and even detected one such instance (the \emph{gibbon} example from section \ref{sec:error_anticipation}). We find the automatic detection of unreliable LLM outputs to be an interesting avenue for future work, both to improve accuracy and gain insight of potential complexities in the given task. 

\textbf{VLMs are not always reliable.} Our work assumes that VLMs are capable of recognizing subpopulations within a class when named. While this is often true, VLMs can still fail, especially for composite concepts. We aim to keep our subpopulations simple, ascribing only one attribute to each. Nonetheless, it is currently not possible to know apriori whether a VLM can recognize a subpopulation in a zero-shot manner. We hope more work on uncertainty estimation can enrich our method, by way of automatically flagging and removing subpopulations that the VLM will not be able to reliably detect.

\section{Conclusion}

\looseness=-1
To represent classes with diverse instances, which can come in many forms, one vector per class may not be enough. Moreover, VLMs have amazing abilities that are restricted when we only use one vector per class. Thus, instead of ignoring intra-class diversity, we \emph{embrace} it, by explicitly inferring and encoding as much of it as we can. We propose a simple nonlinear consolidation scheme that flexibly attends to subpopulations present in an image while ignoring those that are irrelevant. We find that our method consistently matches or improves over strong baselines, and careful ablations indicate that our method's gains come from improving performance on the hardest classes and subpopulations. Thus, embracing diversity can help reduce performance disparities, including on real-world fairness benchmarks, towards models that work well \emph{for all}. Our approach allows powerful models to work together in a transparent way via intermediate interpretable outputs, facilitating inferences with explanations, as well as greater tools to understand and debug potential failures. We hope our work spurs further curiosity around how existing paradigms may limit the capabilities of our modern models, towards the development of new AI systems that overcome the fairness and transparency limitations of today.

\newpage
\bibliographystyle{ACM-Reference-Format}
\bibliography{sample-base}

\newpage
\appendix

\section{Example Interpretable Inferences}
\label{app-sec:inference_egs}

\begin{figure}
    \centering
    \begin{minipage}{0.23\textwidth}
        \centering
        \includegraphics[width=\linewidth]{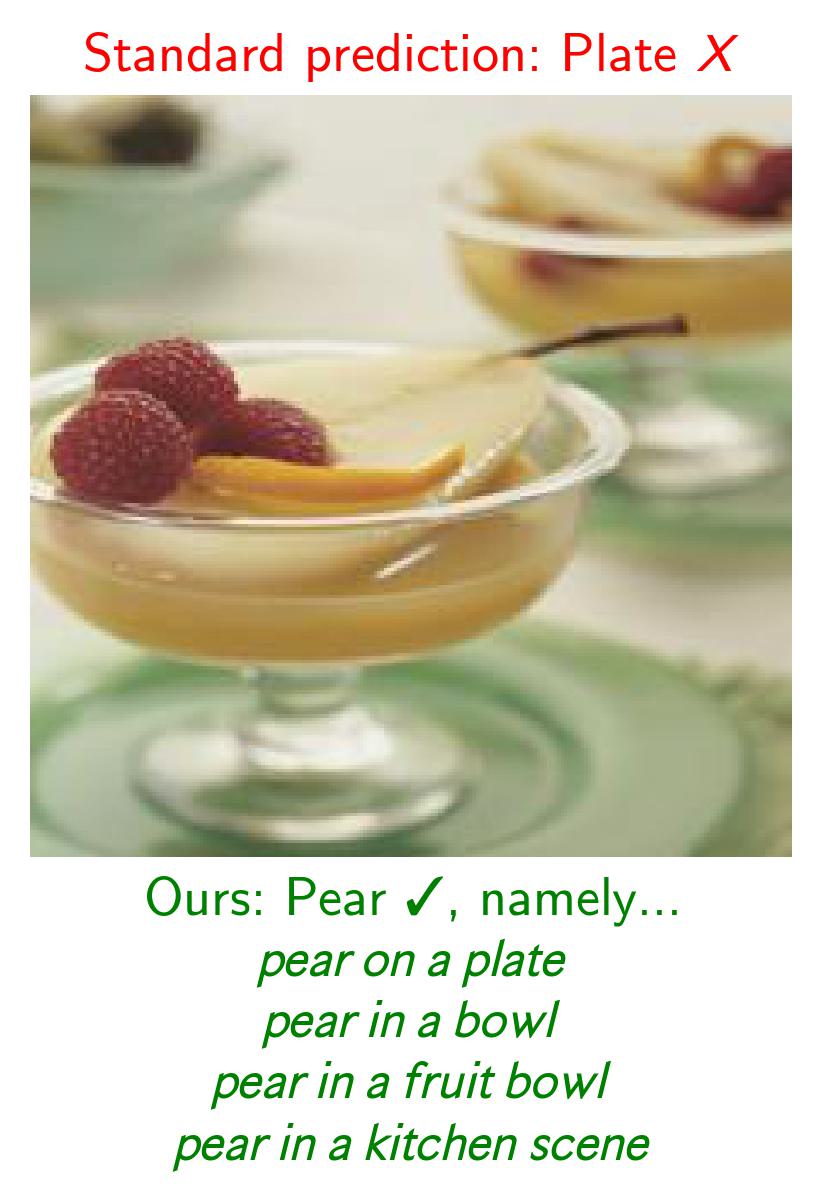}
    \end{minipage}
    \begin{minipage}{0.23\textwidth}
        \centering
        \includegraphics[width=\linewidth]{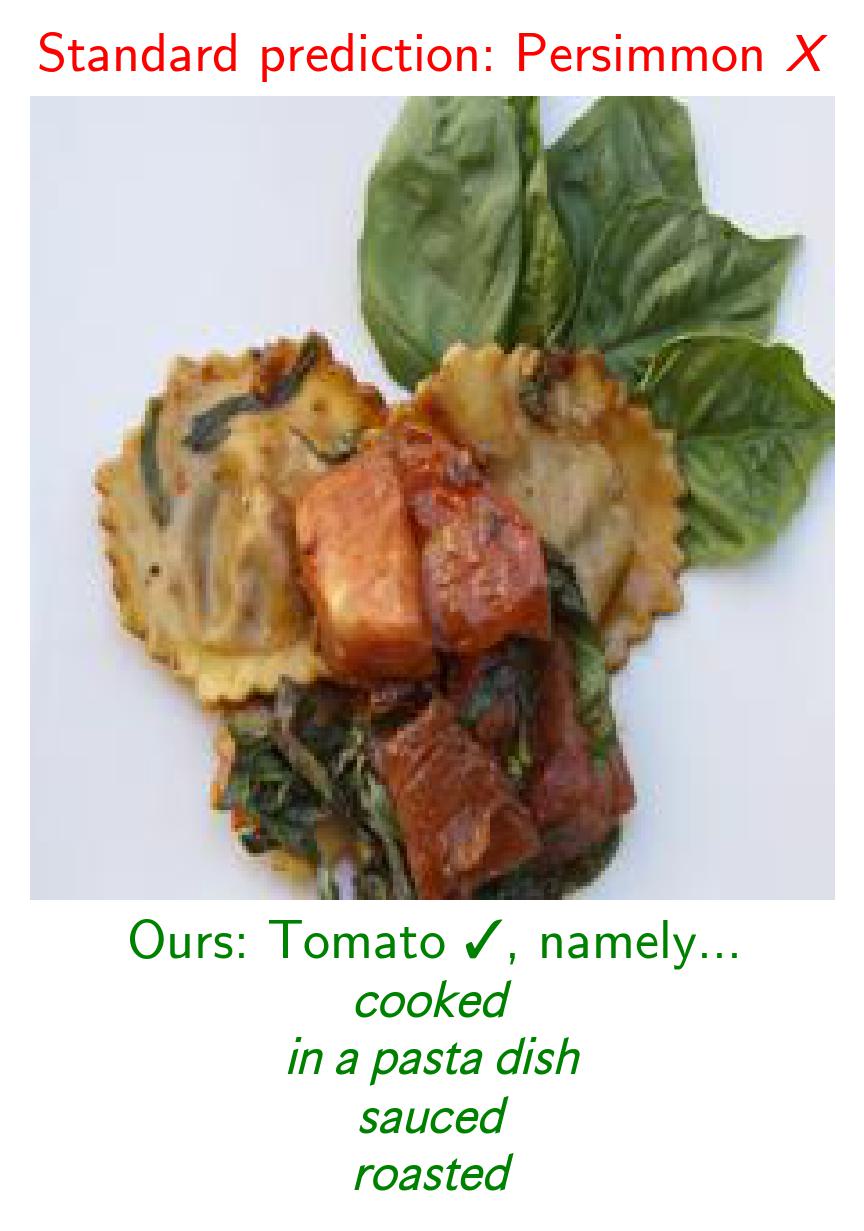}
    \end{minipage} 
    \begin{minipage}{0.23\textwidth}
        \centering
        \includegraphics[width=\linewidth]{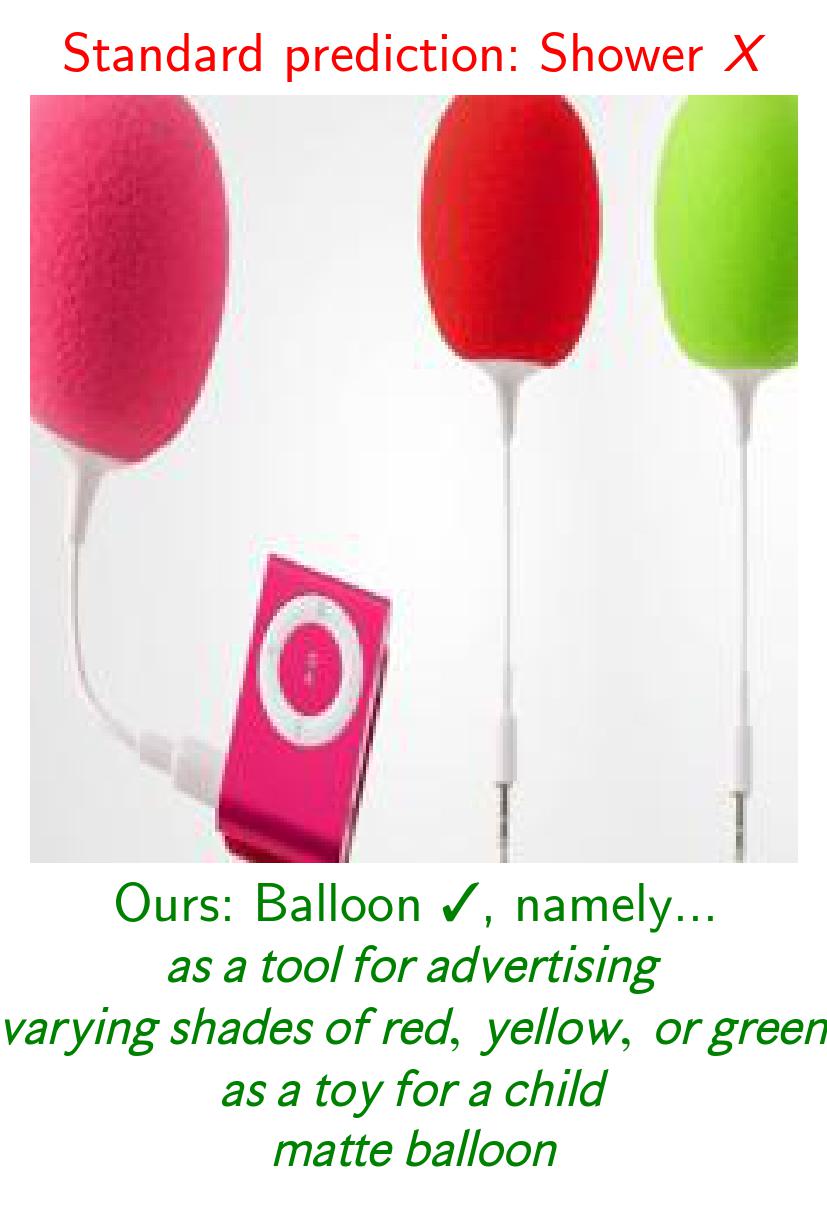}
    \end{minipage}
    \begin{minipage}{0.23\textwidth}
        \centering
        \includegraphics[width=\linewidth]{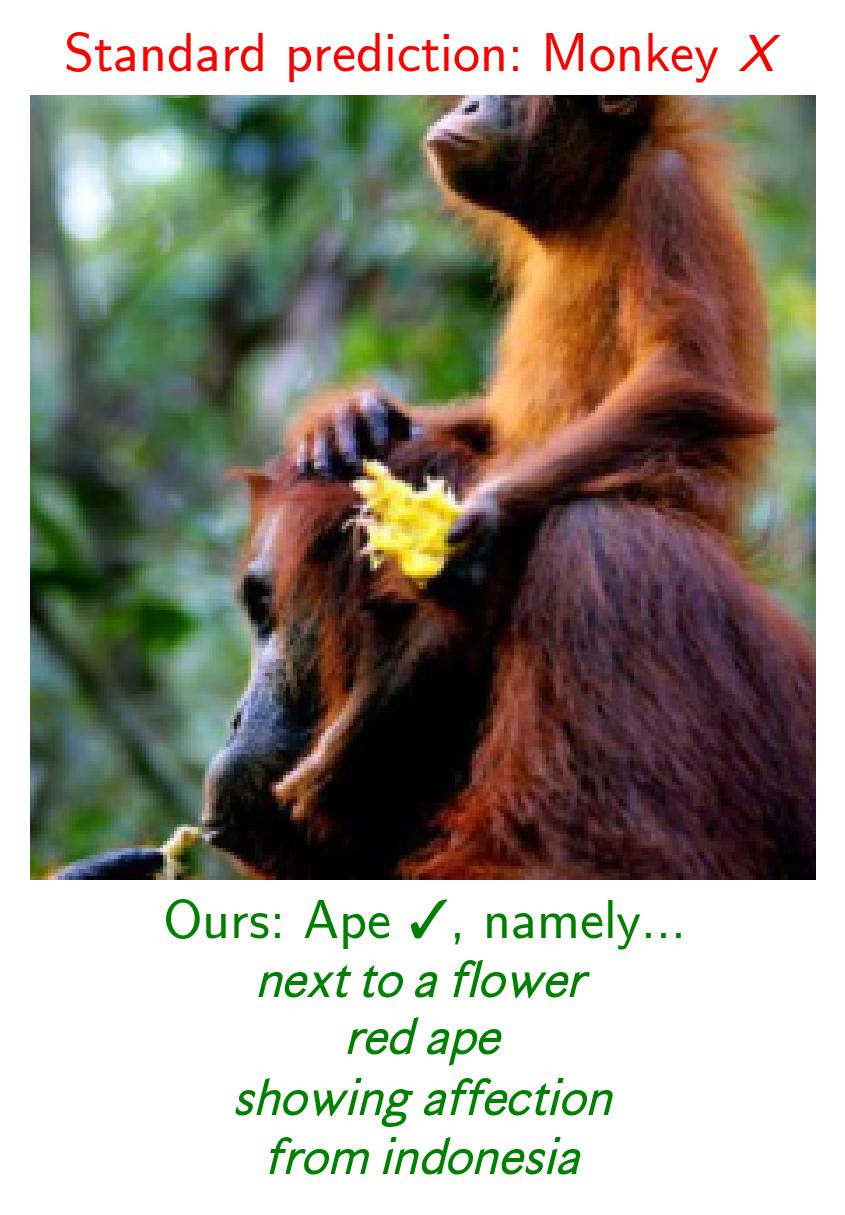}
    \end{minipage}
    \begin{minipage}{0.23\textwidth}
        \centering
        \includegraphics[width=\linewidth]{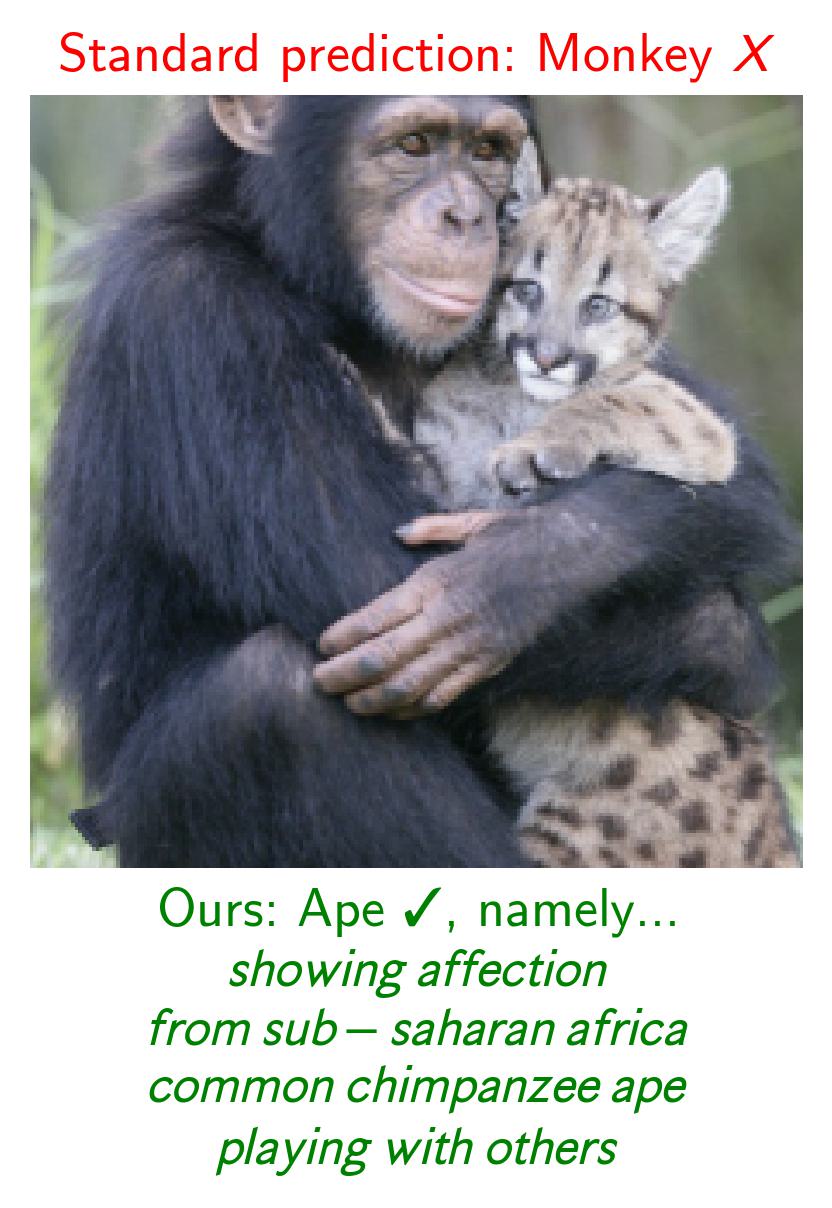}
    \end{minipage} 
    \begin{minipage}{0.23\textwidth}
        \centering
        \includegraphics[width=\linewidth]{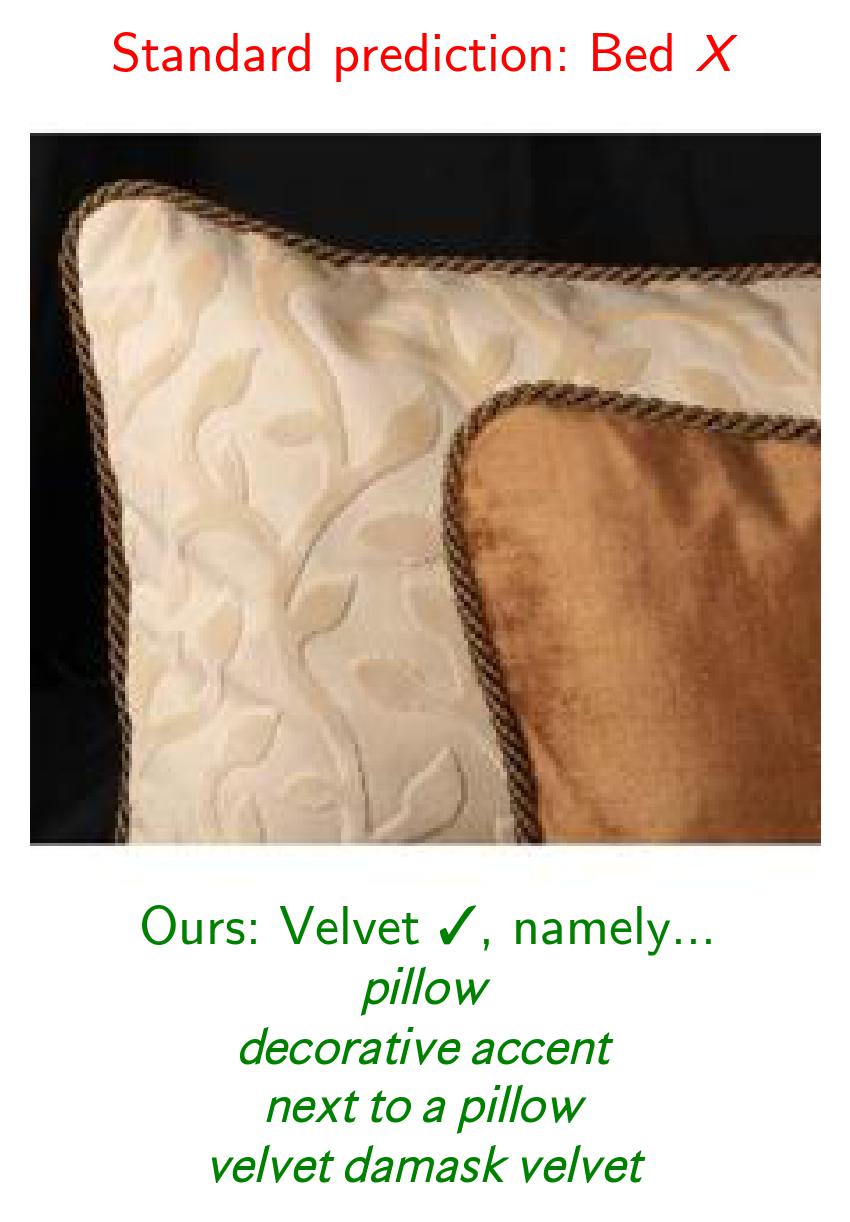}
    \end{minipage}
    \begin{minipage}{0.23\textwidth}
        \centering
        \includegraphics[width=\linewidth]{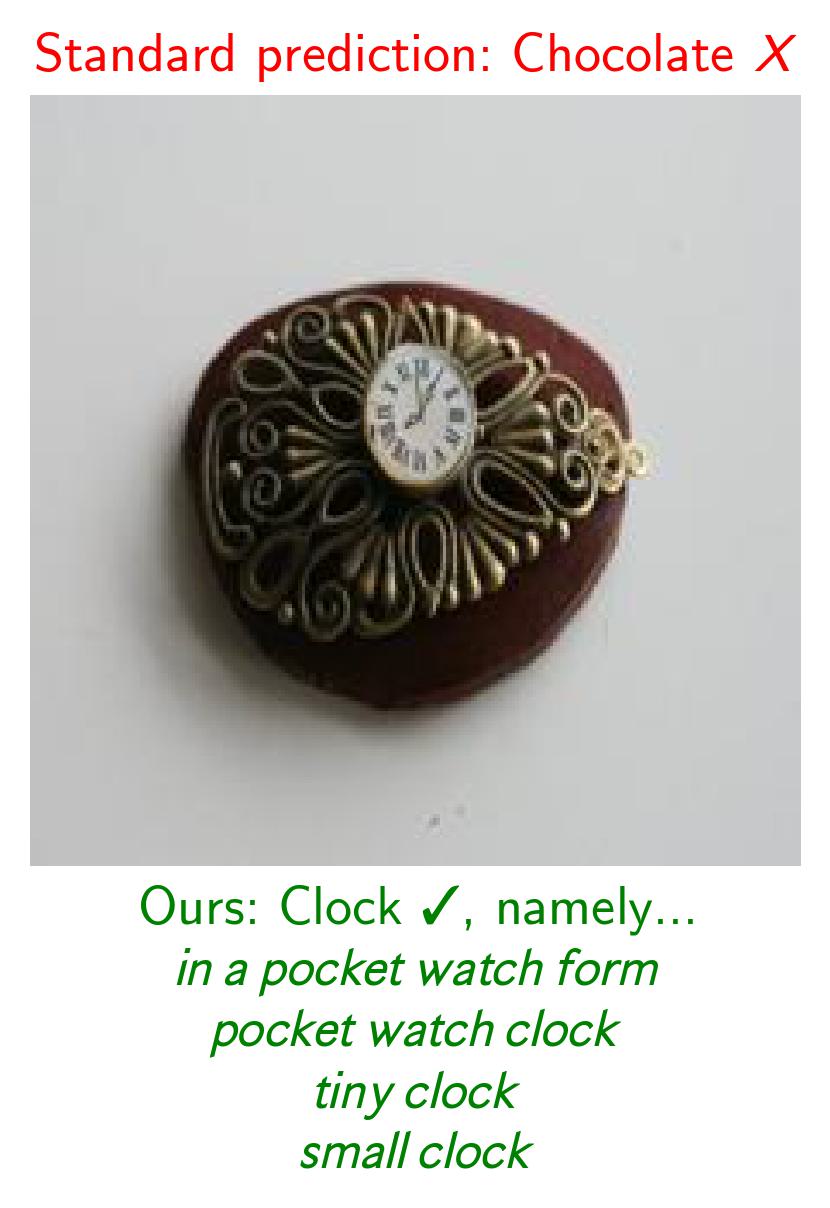}
    \end{minipage}
    \begin{minipage}{0.23\textwidth}
        \centering
        \includegraphics[width=\linewidth]{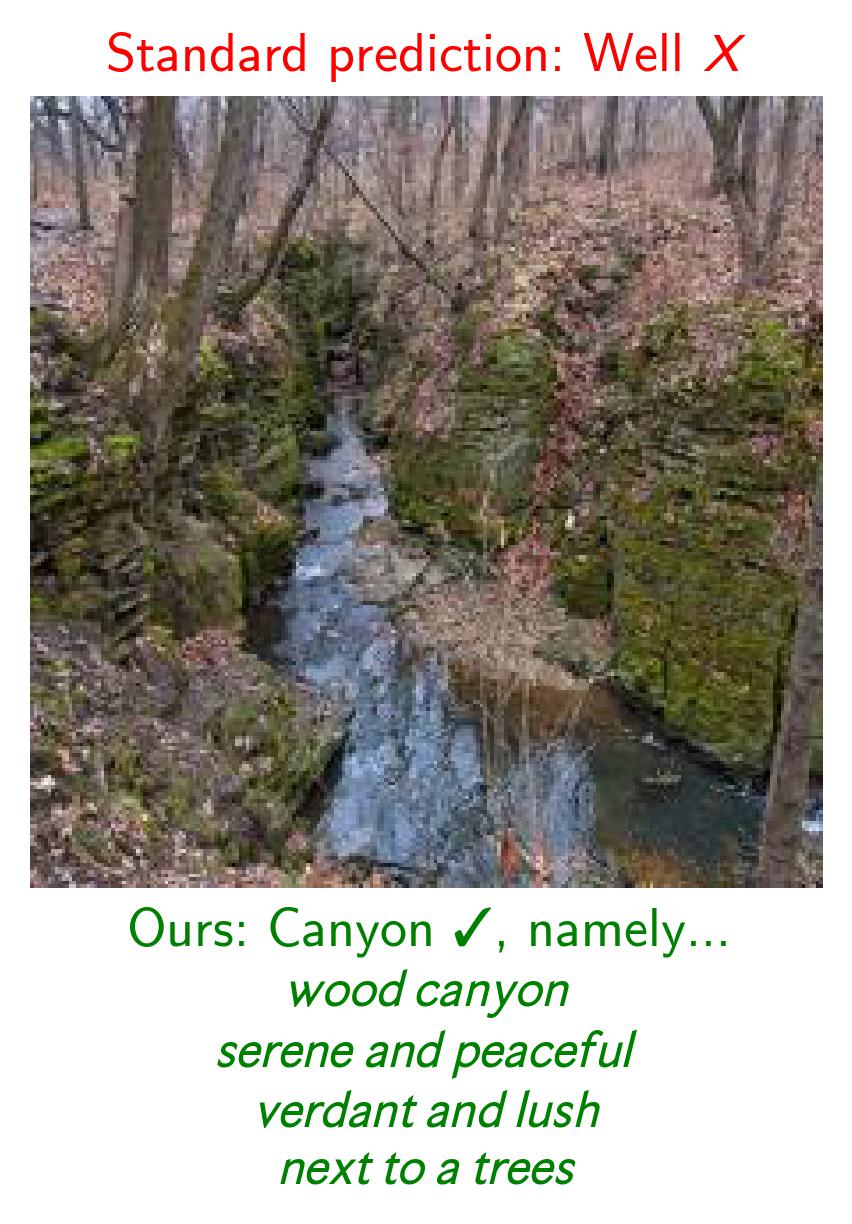}
    \end{minipage}
    \caption{Our method yields faithful, fine-grained interpretations, for free. Top 4 shown for brevity.}
    \label{app-fig:eg_inferences}
\end{figure}

We show additional examples of interpretable inferences in figure \ref{app-fig:eg_inferences}.

\section{Cases Where Attributes Help Most}
\label{app-sec:ap_gains}

\begin{figure}
    \begin{minipage}{0.48\linewidth}
    \centering
    \includegraphics[width=\linewidth]{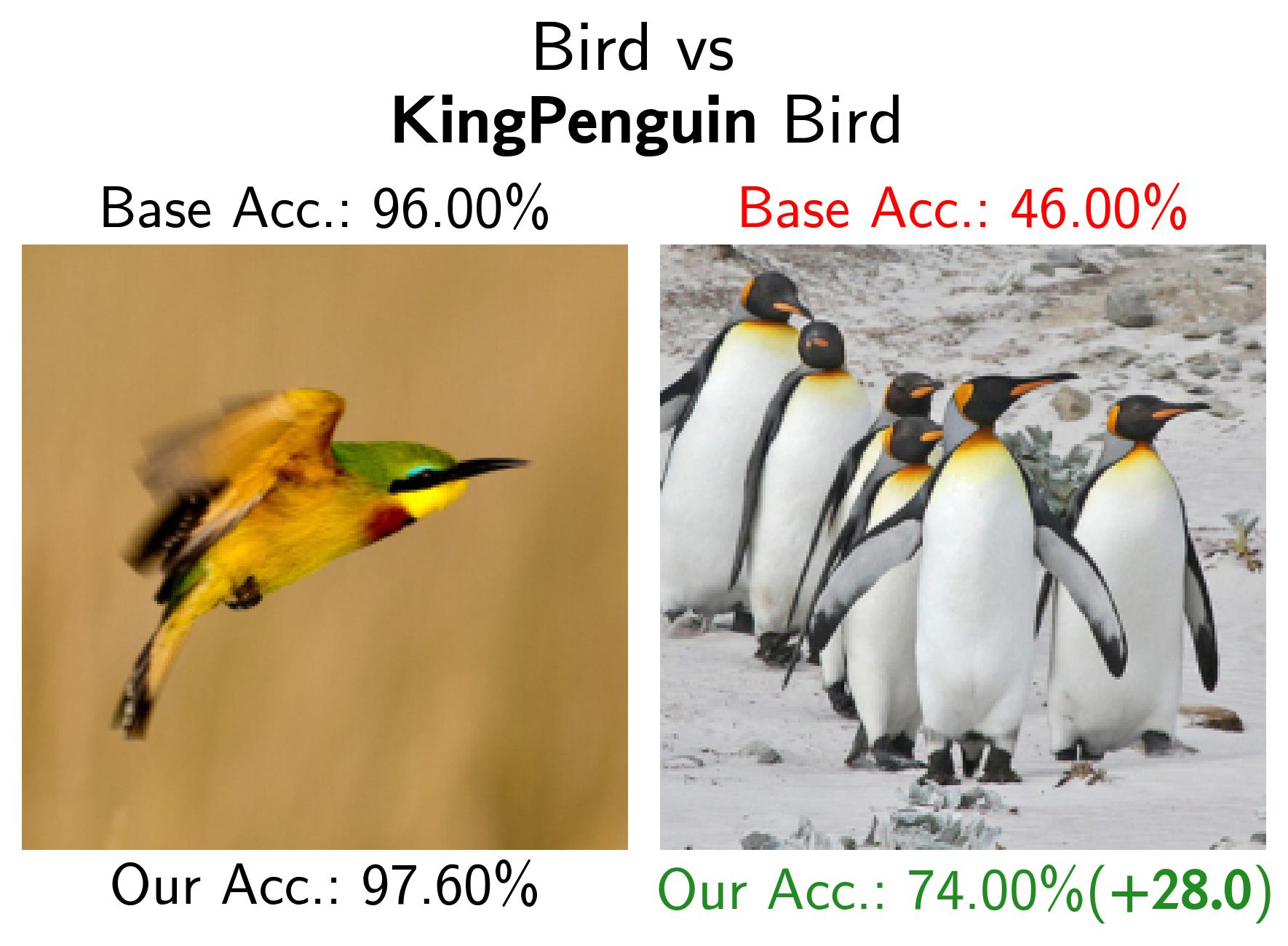}  
    \end{minipage}
    \begin{minipage}{0.48\linewidth}
    \centering
    \includegraphics[width=\linewidth]{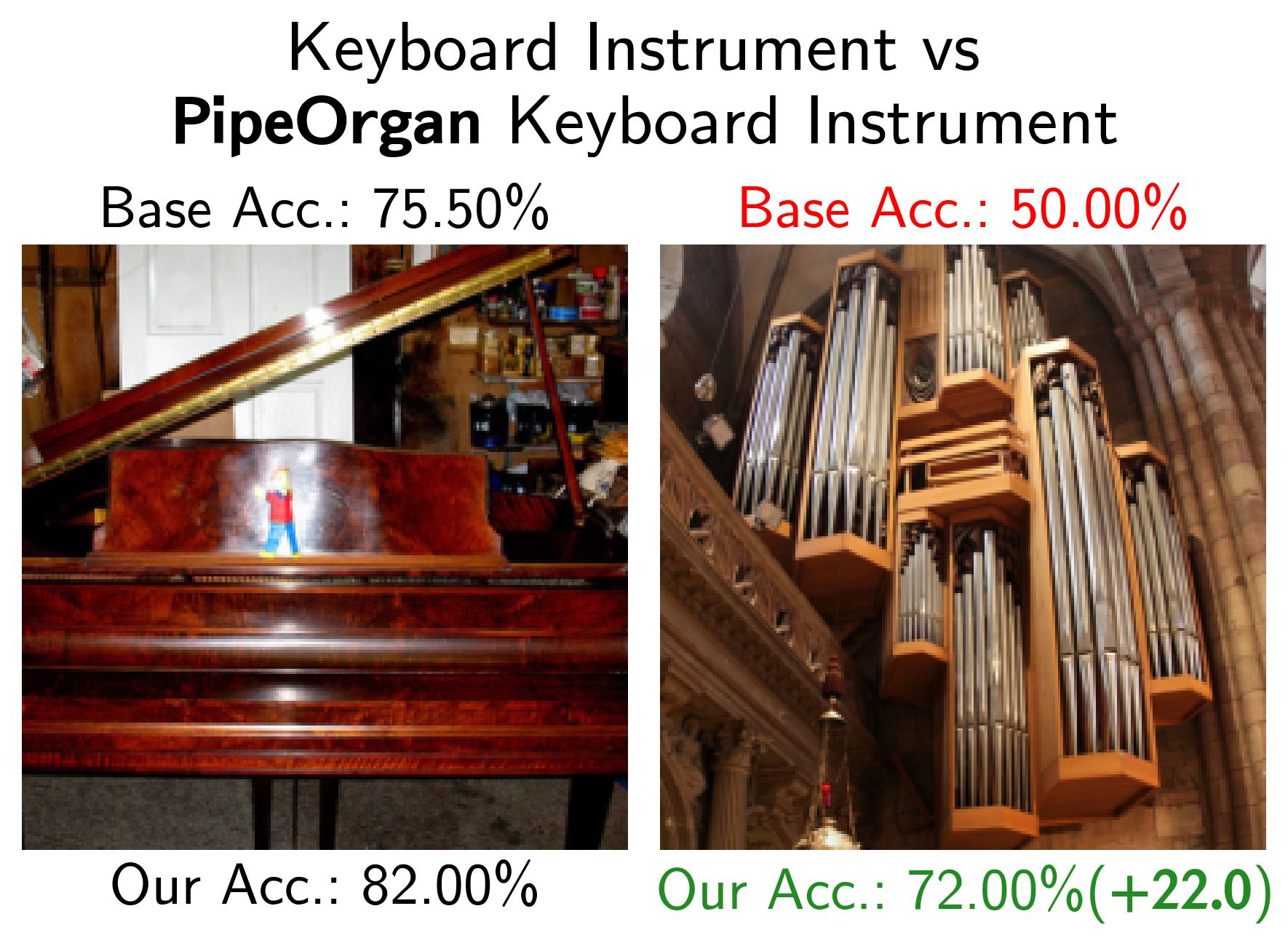}  
    \end{minipage}
    \begin{minipage}{0.48\linewidth}
    \centering
    \includegraphics[width=\linewidth]{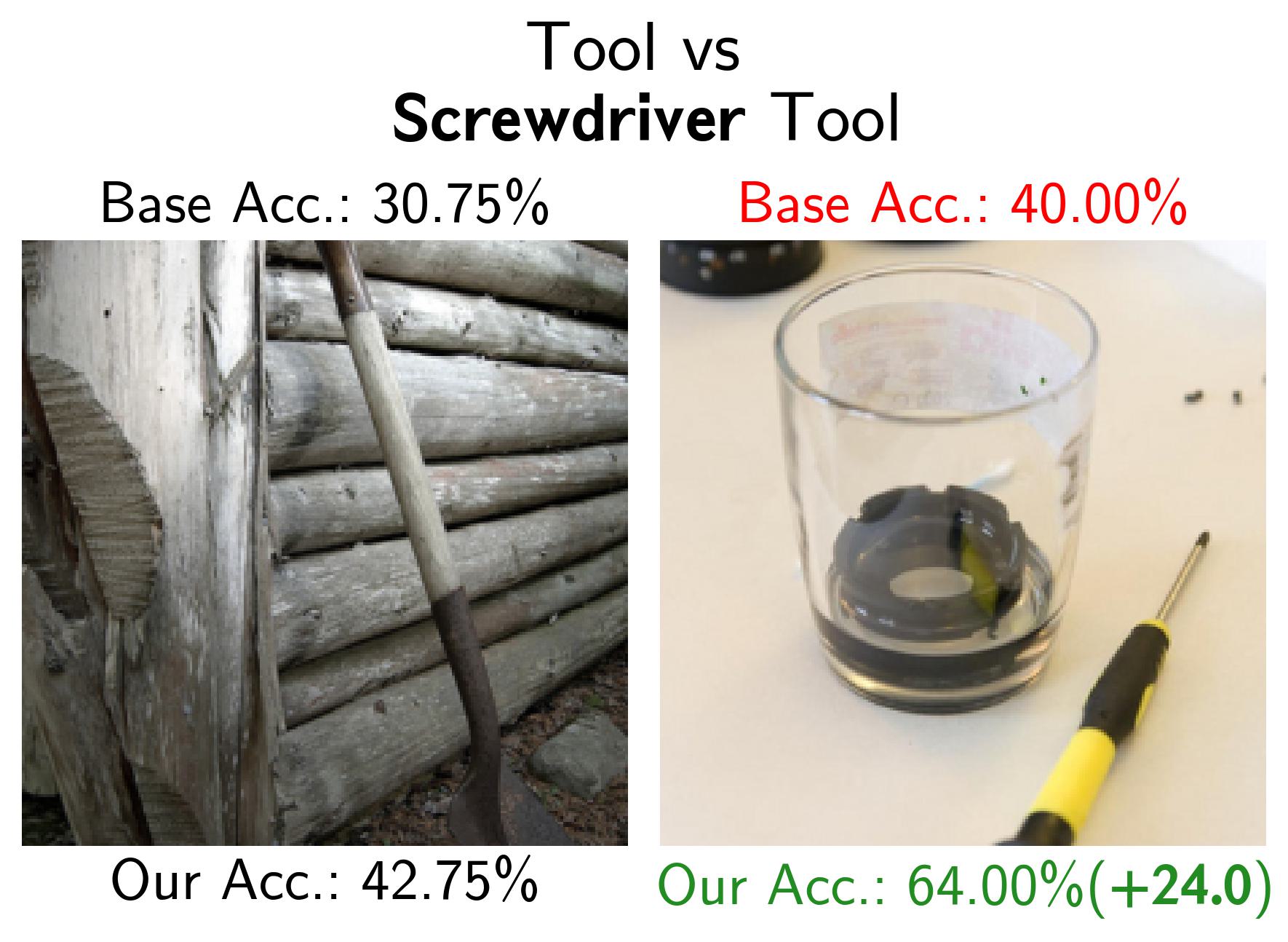}  
    \end{minipage}
    \begin{minipage}{0.48\linewidth}
    \centering
    \includegraphics[width=\linewidth]{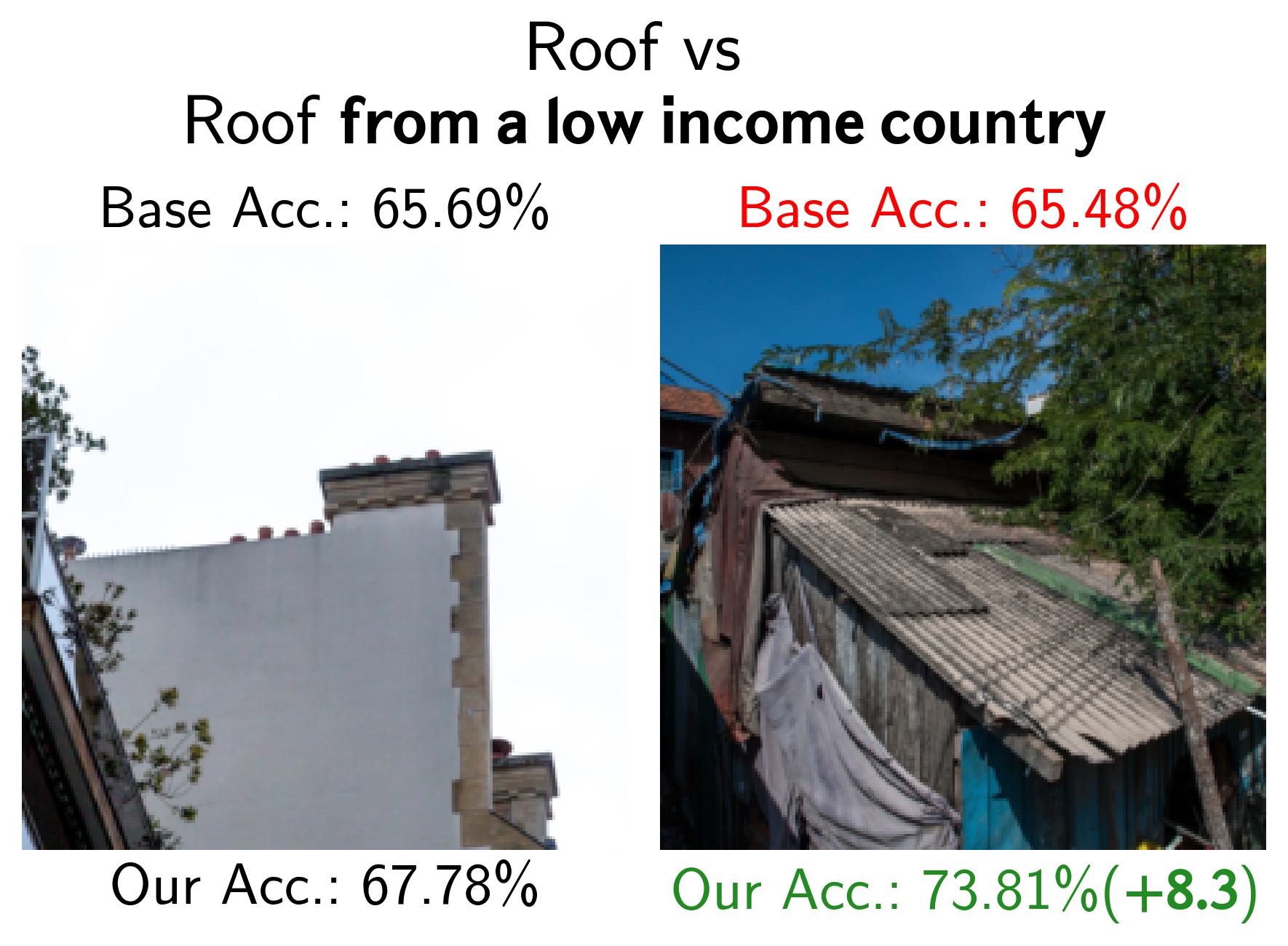}
    \end{minipage}
    \begin{minipage}{0.48\linewidth}
    \centering
    \includegraphics[width=\linewidth]{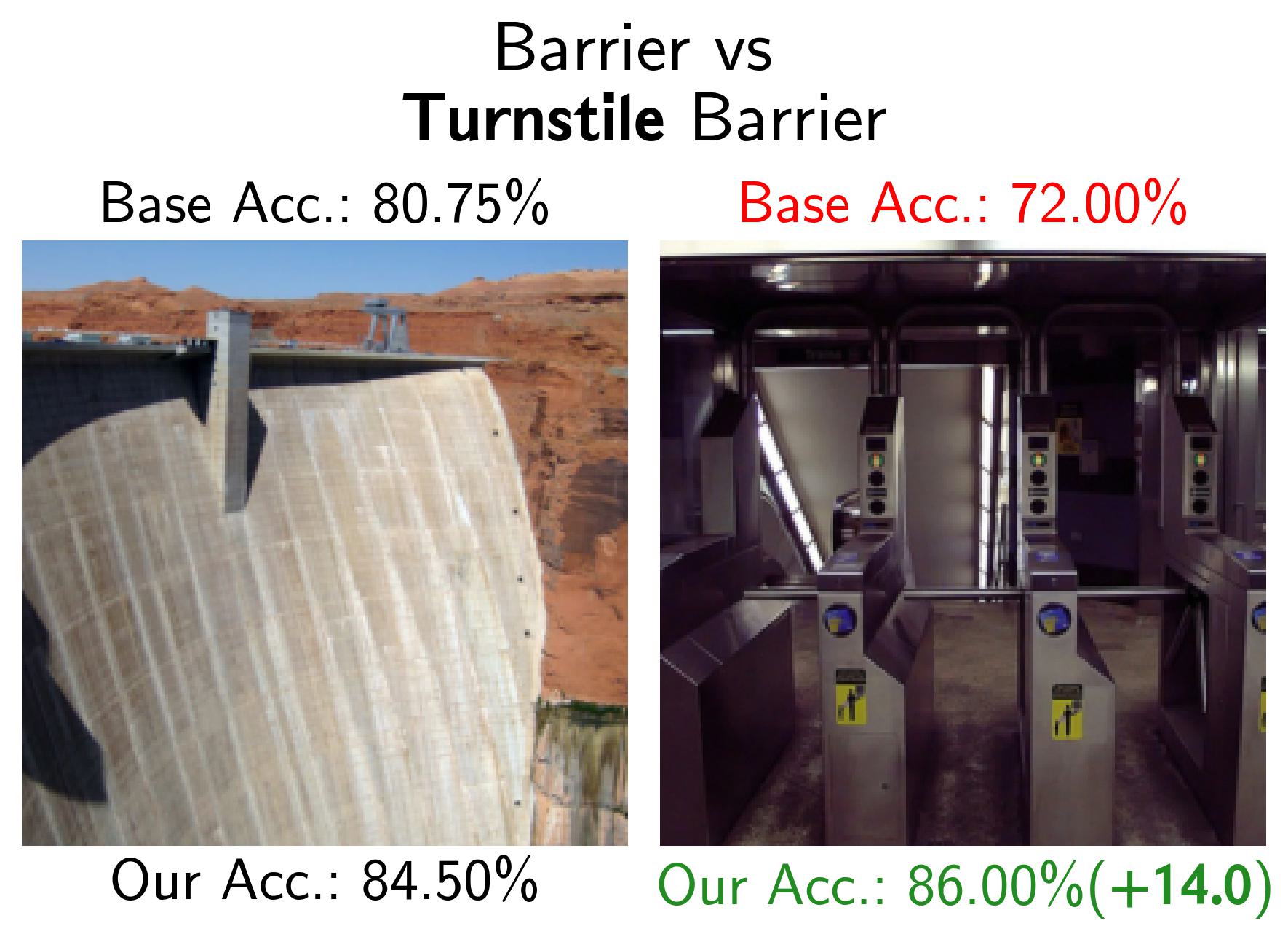}  
    \end{minipage}\hfill
    \begin{minipage}{0.48\linewidth}
    \centering
    \includegraphics[width=\linewidth]{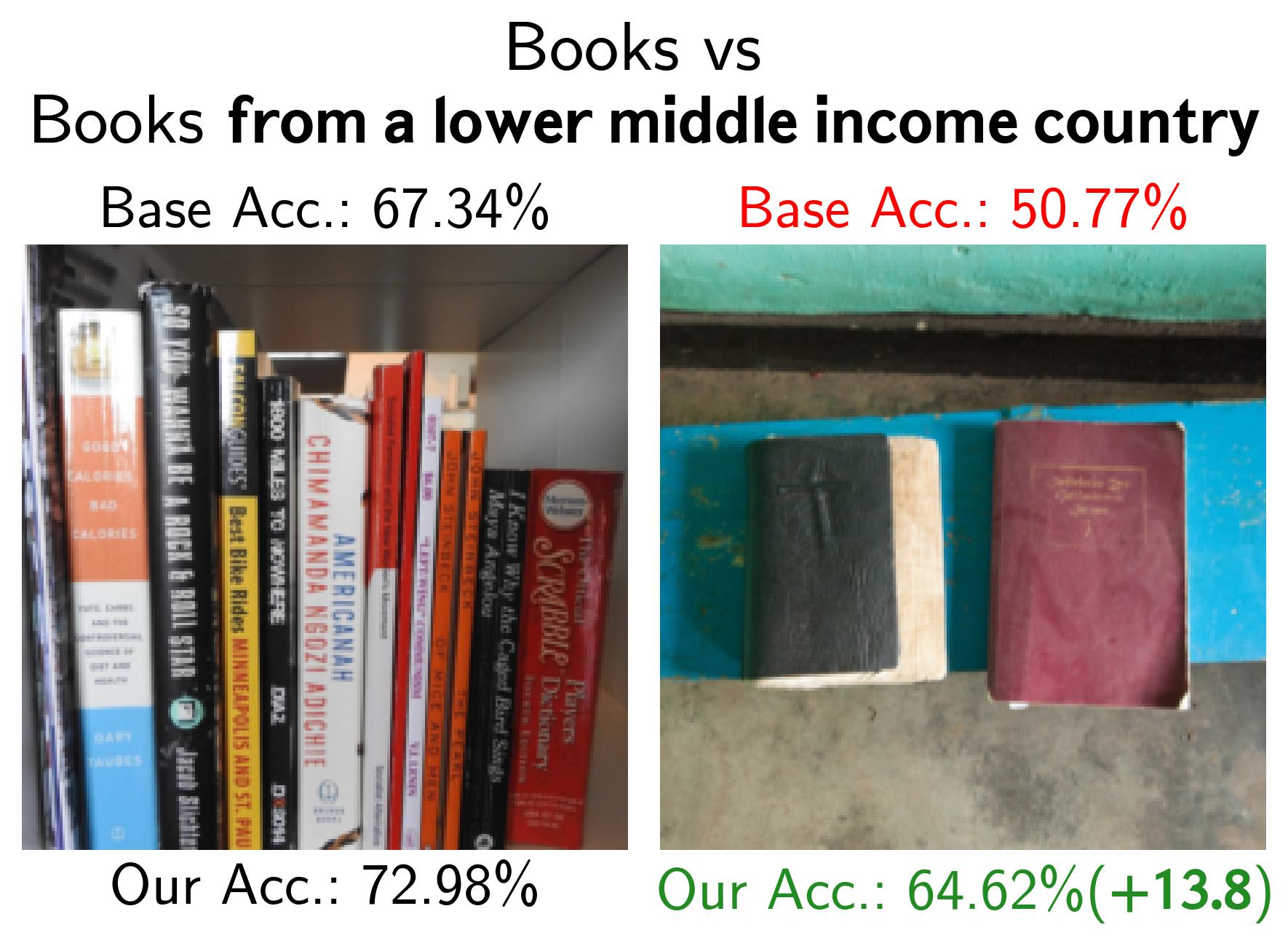}
    \end{minipage}
    \caption{Example subpopulations where our method exhibits sizable accuracy gains compared to standard zero-shot classification (i.e. classname embedding only).}
    \label{app-fig:bias_examples}
\end{figure}

We now provide more examples of instances where the prevailing paradigm for zero-shot classification results in disparate performance, and consequently, where our method yields largest improvements. The crux of the issue with existing paradigm is that the classname embedding struggles to be close to embeddings for images from \emph{all} subpopulations of the class, particularly when the class contains many visually diverse subpopulations. For example, a penguin looks very different than most birds, so embeddings of penguin images will be some distance away from embeddings of most birds. Similarly, the penguin images may not reside close to the text embedding of the caption `a photo of a bird'. Indeed, we find standard zero-shot classification accuracy for King Penguin birds is only $46\%$, while accuracy for the class is $96\%$. Figure \ref{app-fig:bias_examples} shows this example along with other instances where standard zero-shot classification leads to biased performance. We highlight examples that our method leads to improvements. Notice that the subpopulations tend to be atypical. 

How then does our method result in improvements? We leverage the fact that despite poor standard zero-shot accuracy on subpopulations that lie far from their classname embedding, VLMs are still capable of recognizing these atypcial subpopulations. That is, penguin images may be far from the `bird' text embedding, but they are actually quite close to the `penguin' text embedding. That is, standard zero-shot classification does not take advantage of the ability of VLMs to recognize objects at a deeper level than that of the classification task. Thus, by including the right attributes, we can enable accurate recognition of atypical subpopulations. 

\begin{figure}
    \centering
    \begin{minipage}{0.19\linewidth}
    \centering
    \includegraphics[width=\linewidth]{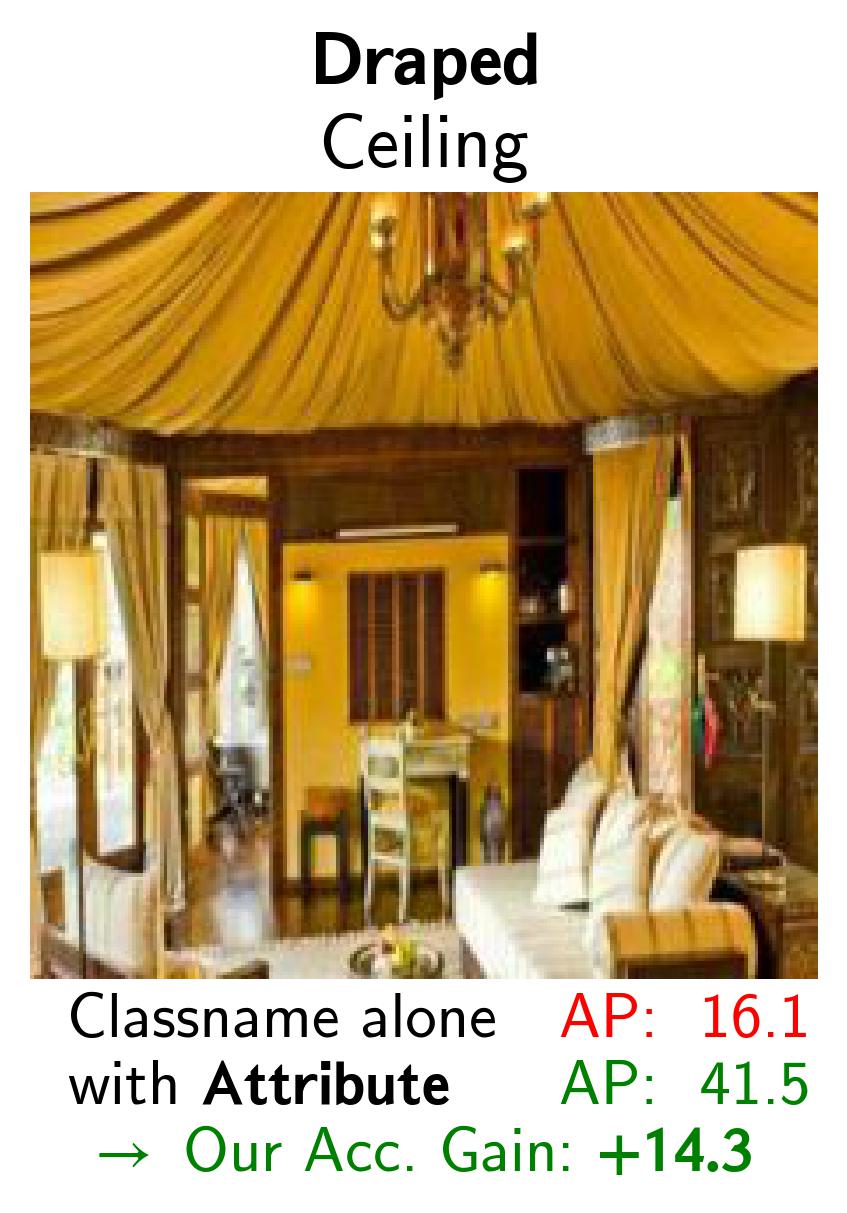}  
    \end{minipage}
    \begin{minipage}{0.19\linewidth}
    \centering
    \includegraphics[width=\linewidth]{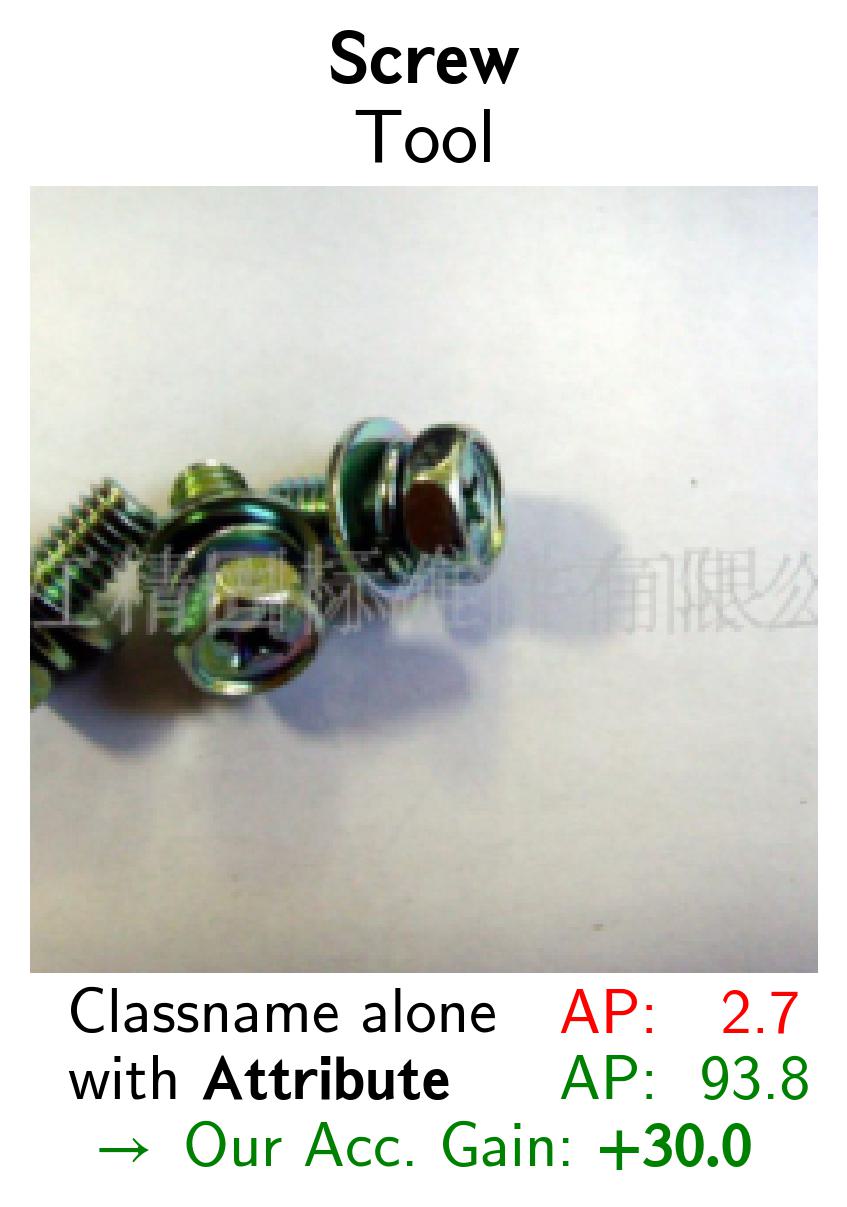}
    \end{minipage} 
    \begin{minipage}{0.19\linewidth}
    \centering
    \includegraphics[width=\linewidth]{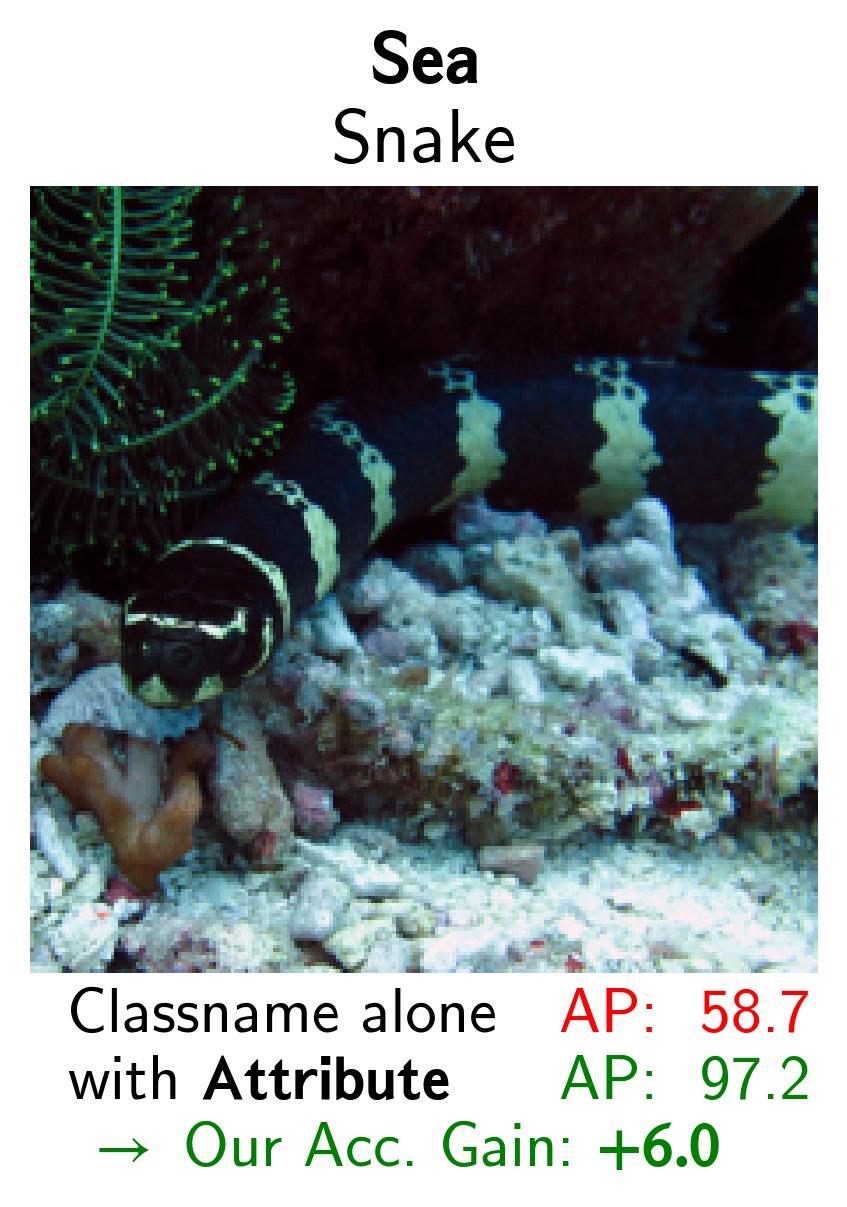}  
    \end{minipage}
    \begin{minipage}{0.19\linewidth}
    \centering
    \includegraphics[width=\linewidth]{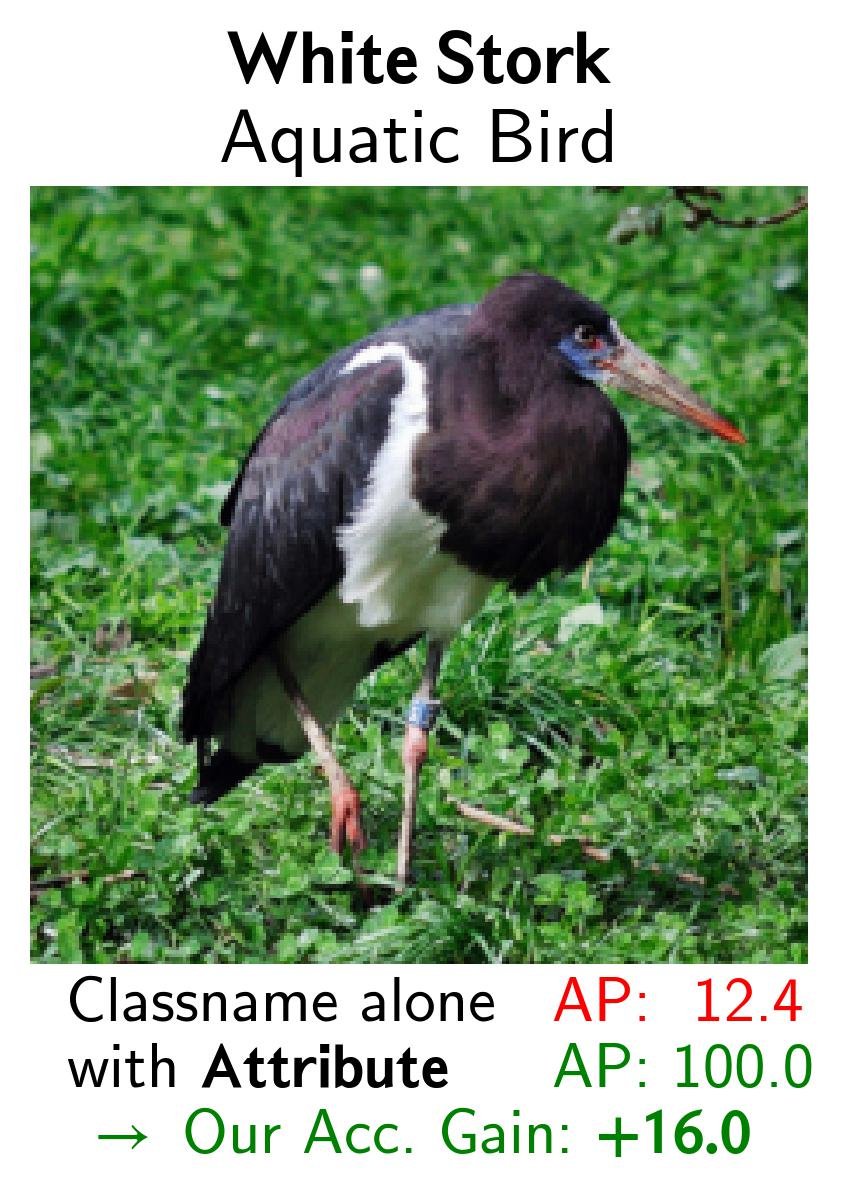}  
    \end{minipage}
    \begin{minipage}{0.19\linewidth}
    \centering
    \includegraphics[width=\linewidth]{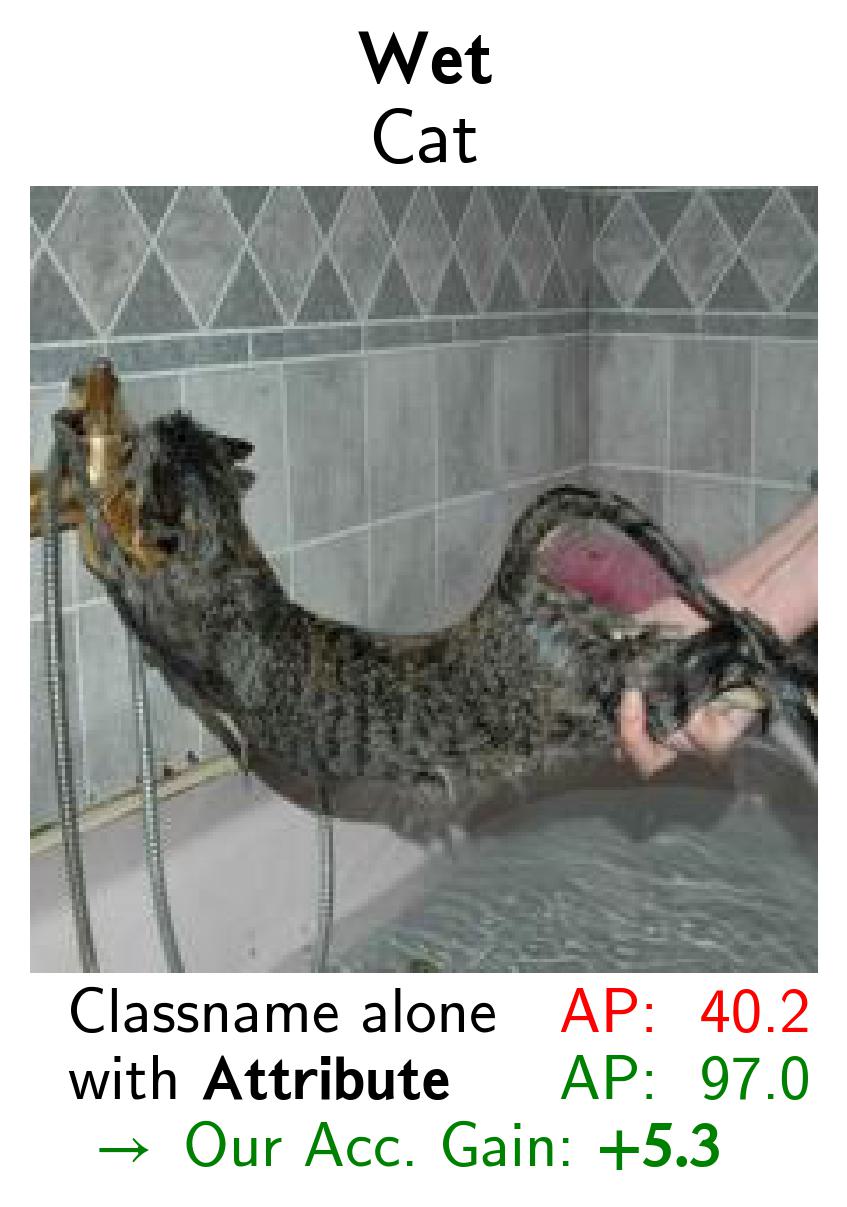}  
    \end{minipage}
    \caption{Example subpopulations where the classname embedding is imprecise, but including the attribute leads to large boosts in average precision. Notably, these subpopulations reflect instances atypical to the class.}
    \label{app-fig:ap_gain_egs}
\end{figure}

Figure \ref{app-fig:ap_gain_egs} shows more examples of subpopulations where including the groundtruth attribute results in significant gains in average precision, indicating that including the attribute allows for recognition of atypical subpopulations. In this figure, AP corresponds to the average precision score obtained when using the similarity of an image to (a) the classname embedding or (b) the embedding of the classname with the attribute (e.g. `bird' vs. `King Penguin bird') to detect images belonging to that subpopulation. Again, these subpopulations generally appear differently than a typical instance from their class, making the classname embedding an imprecise probe for that subpopulation. However, evidently, when given the attribute, VLMs are still capable of recognizing the subpopulation.

\section{Details on Correlation between Diversity and Accuracy per Class}
\label{app-sec:tension}

We compute ImageNet accuracy per class using three models: CLIP ViT-B/16 via standard zero-shot classification, DINO ViT-S/16 with a linear classification head fit to ImageNet over fixed features \cite{dino}, and a ViT-S/16 trained with traditional class-label supervision on ImageNet \cite{deit}. Notably, all these models utilize a linear classification head. That is, they operate under a one vector one class paradigm. To proxy diversity, we measure the variance of embeddings per class. That is, per class, we compute the average squared distance between the mean embedding and the embedding of each class instance. Note that our measure of diversity depends on the image encoder; we explore using each of the three aforementioned models. Table \ref{tab:diversity_and_acc_imagenet} shows the results. All correlations are strongly negative, indicating that across classifiers and using various measures of diversity, classes with higher diversity are predicted at lower accuracies. This supports the intuitive hypothesis that consistently representing an entire class with one vector is made challenging when the class contains diverse instances.

\begin{table}
    \centering
    \begin{tabular}{lrrr}
    \toprule
    Classifier &  CLIP &  DINO &   Sup. \\
    Encoder &       &       &       \\
    \midrule
    CLIP    & -0.28 & -0.51 & -0.43 \\
    DINO    & -0.37 & -0.54 & -0.48 \\
    Sup.     & -0.47 & -0.72 & -0.65 \\
    \bottomrule
    \end{tabular}
    \caption{Correlation between diversity and accuracy by class on ImageNet. We study three models: vision transformers trained with CLIP, DINO, or traditional label supervision. Diversity refers to variance of image embeddings within a class, with embeddings obtained with the `encoder' model. }
    \label{tab:diversity_and_acc_imagenet}
\end{table}

\section{Additional Experimental Details}

Note that we will provide all code, so that further details are easily accessible. 

\subsection{Datasets}
\label{app-sec:datasets}

The four hierarchical datasets we utilize are subsets of ImageNet \cite{imagenet} curated by \cite{Santurkar2020BREEDSBF}. We also utilize the attributed dataset of MIT States \cite{Isola2015DiscoveringSA}, deriving two classification tasks from their annotations. Finally, we utilize the geographic fairness benchmarks of Dollarstreet \cite{rojas2022the} and GeoDE \cite{ramaswamy2022geode}. When reporting subpopulation accuracies, we use income level as the ground truth attribute for Dollarstreet. Note that for MIT States and Dollarstreet, we conduct a filtering of classnames. Namely, we compute cosine similarity of CLIP embeddings for each pair of classnames. For any pair exceeding a threshold, we remove one classname from consideration. We do this because MIT States was not originally intended to be a classification dataset, and we observed highly similar classnames in Dollarstreet (e.g. `toilet' and `bathroom/toilet'). We use a threshold of $0.8$ and $0.9$ to generate the coarse and fine-grained MIT States datasets respectively, and use a threshold of $0.9$ for Dollarstreet. 

\subsection{Inferring Attributes}
\label{app-sec:attr_inference}
\begin{table}[]
    \centering
\resizebox{\textwidth}{!}{
\begin{tabular}{lll}
\toprule
Query & Prompt & Examples \\
\midrule
Kinds & List 16 different kinds of pear & Bartlett, Bosc, D'Anjou \\
States & List 10 different ways in which a pear may appear in an image & Whole pear, Pear slices, Pear chunks \\
Descriptors & List useful features for distinguishing a pear in an image & Round shape, Glossy skin, Green or brown color \\
Co-occurring Objects & In an image of a pear, list 10 other objects that may also appear & Leaves, Stem, Branches \\
Backgrounds & List ten different locations in which a pear may appear in an image & Fruit basket, Still life painting, Candy dish \\
\bottomrule
\end{tabular}
}
    \caption{Example LLM prompts and outputs for class-specific and class-adjacent queries.}
    \label{app-tab:llm_queries}
\end{table}

\looseness=-1
We now provide details on our exact LLM queries. First, for class-specific and class-adjacent queries, table \ref{app-tab:llm_queries} shows the precise prompt shown to the LLM along with example outputs, both for the class \texttt{pear}. For all queries, we append \texttt{Only use up to three words per list item} so that the LLM does not drone on. We sample from the LLM (Vicuna-13b-v1.5) with a temperature of $0.7$, repetition penalty of $1$, and a max number of new tokens of $512$.

We now provide more information on class-agnostic queries. We use continents as regions, and the five most populous countries per continent as our list of countries. These can both be obtained via prompting an LLM or searching the internet. 

\subsection{Auto Global}
\label{app-sec:auto_global}
We now show more details for the auto-global query, which we found quite impressive. It consistently was amongst the attribute type that provided the most accuracy gains across datasets. The first prompt to the LLM was:

\texttt{List 16 common general ways in which two instances of the same object may look different. For example, size, age, or cleanliness. Only use one word per list item.}

The next prompt was:

\texttt{For each of those items, list up to four different general adjectives related to the time. Please use common words.}. 

Then, finally, out of laziness, we included a third prompt of:

\texttt{Thanks. Please organize your output as a python dictionary.}

The resultant axes of variation and attributes per axis can be found in Table \ref{app-tab:auto_global}.

\begin{table}
    \centering
    \begin{tabular}{lcccc}
\toprule
Axis & \multicolumn{4}{c}{Attributes} \\
\midrule
size & small & medium & large & tiny \\
age & young & mature & ancient & old \\
cleanliness & dirty & clean & spotless & grimy \\
color & white & black & red & blue \\
texture & rough & smooth & soft & hard \\
material & plastic & metal & wood & fabric \\
shape & round & square & rectangular & triangular \\
position & upright & horizontal & vertical & diagonal \\
reflection & bright & dull & shiny & matte \\
transparency & clear & opaque & translucent & transparent \\
shine & glossy & matte & shiny & dull \\
pattern & striped & polka-dotted & plaid & solid \\
markings & spotted & striped & checked & speckled \\
surface & rough & smooth & bumpy & even \\
appearance & appealing & unappealing & attractive & unattractive \\
\bottomrule
\end{tabular}
    \caption{Attributes and axes of diversity inferred via the \textbf{auto-global} query. See \ref{app-sec:auto_global} for more information. }
    \label{app-tab:auto_global}
\end{table}

\section{Additional Results}

In the main text, we presented results using CLIP. Results for BLIP-2 can be found in Tables \ref{app-tab:known_variation_blip2} and \ref{app-tab:geo}. Trends are consistent with results CLIP. For a global picture, we present results averaged over both VLMs and all datasets in table \ref{app-tab:avg_performance}. Our method performs best over all metrics, again with largest gains occurring over the worst classes and subpopulations. 

We also show results for each dataset individually in table \ref{app-tab:per_dset}. We find it encouraging that our results are consistent across both VLMs and for each of our eight datasets. 

\begin{table}[]
    \centering
    \resizebox{\textwidth}{!}{
\begin{tabular}{lcccccc} \toprule
 & \multicolumn{1}{l}{Accuracy}  & \multicolumn{1}{c}{Avg Worst} & \multicolumn{1}{c}{Worst 20\% of}  & \multicolumn{1}{c}{Worst 20\% of} & \multicolumn{1}{c}{Worst 10\% of} & \multicolumn{1}{c}{Worst 10\% of}                    \\

Method &  & Subpop & Classes & Subpops & Classes & Subpops \\
\midrule
Vanilla & 73.22 & 50.17 & 44.90 & 33.10 & 36.66 & 22.05 \\
DCLIP & 72.65 & 49.72 & 45.35 & 32.72 & \underline{37.16} & 21.92 \\
Waffle & 73.36 & 50.23 & 44.97 & 33.34 & 36.66 & 22.43 \\
CHiLS & \underline{74.13} & \underline{51.84} & \underline{46.00} & \underline{34.80} & 37.07 & \underline{23.24} \\
Ours & \textbf{74.75} & \textbf{52.04} & \textbf{47.52} & \textbf{35.77} & \textbf{39.21} & \textbf{24.40} \\ \bottomrule
\end{tabular}
}
    \caption{Average performance over eight datasets and two VLMs.}
    \label{app-tab:avg_performance}
\end{table}

\begin{table}
\centering
\begin{tabular}{llcccc}
\toprule
 &  & Accuracy & Avg Worst & Worst 20\% of & Worst 20\% of \\
Dataset Type &  &  & Subpop & Classes & Subpops \\
\midrule
\multirow[t]{5}{*}{States} & Vanilla & 70.60 & \underline{42.65} & 43.44 & 26.28 \\
 & DCLIP & 69.80 & 41.42 & 41.54 & 24.25 \\
 & Waffle & 70.10 & 42.18 & 41.99 & 25.76 \\
 & CHiLS & \underline{70.83} & 42.51 & \textbf{44.31} & \underline{26.75} \\
 & Ours & \textbf{71.30} & \textbf{42.84} & \underline{43.92} & \textbf{27.21} \\
 \midrule
\multirow[t]{5}{*}{Hierarchical} & Vanilla & 75.29 & 50.33 & 44.30 & 32.18 \\
 & DCLIP & 75.60 & 49.41 & \underline{46.35} & 32.25 \\
 & Waffle & 75.25 & 48.84 & 44.48 & 31.67 \\
 & CHiLS & \underline{77.17} & \underline{52.00} & 45.86 & \underline{34.59} \\
 & Ours & \textbf{77.95} & \textbf{52.47} & \textbf{48.66} & \textbf{35.46} \\
\bottomrule
\end{tabular}
\caption{Zero-shot classification on datasets with known variation types for BLIP-2. Hierarchical datasets from \citet{novack2023chils} and States are the average of coarse and fine-grained categorizations of MIT States. See table \ref{tab:known_variation} for results using CLIP ViT-B/16.}
\label{app-tab:known_variation_blip2}
\end{table}

\begin{table}[]
\centering
\begin{tabular}{@{}lcccccc@{}}
\toprule
\textit{DollarStreet} & \multicolumn{1}{l}{}         & \multicolumn{1}{c}{Worst}             & \multicolumn{1}{c}{Worst}             & \multicolumn{1}{c}{Avg Worst} & \multicolumn{1}{c}{ Worst 20\% of} & \multicolumn{1}{c}{Worst 20\% of}                    \\
Method & Accuracy &  Region & Income & Subpop & Classes & Subpops \\
\midrule
Vanilla & 50.91 & 39.76 & 31.89 & 36.76 & 18.87 & 11.33 \\
DCLIP & 49.81 & 39.05 & 32.03 & 37.01 & 18.22 & 12.14 \\
Waffle & 51.07 & \textbf{41.00} & \textbf{33.05} & 36.67 & 19.43 & \underline{12.53} \\
CHiLS & \underline{51.56} & 40.26 & 32.37 & \textbf{38.35} & \underline{19.56} & 12.45 \\
Ours & \textbf{51.96} & \underline{40.63} & \underline{32.78} & \underline{37.91} & \textbf{21.04} & \textbf{13.61} \\ \midrule
\textit{GeoDE} &                &                &                &                \\ \midrule
Vanilla & 90.48 & 87.95 & - & 84.41 & 71.01 & 69.06 \\
DCLIP & 90.98 & 88.19 & - & 84.78 & 72.71 & \underline{71.32} \\
Waffle & \underline{91.10} & \underline{88.85} & - & \underline{84.97} & \textbf{74.11} & \textbf{72.56} \\
CHiLS & 90.75 & 87.99 & - & 84.63 & 71.11 & 69.46 \\
Ours & \textbf{91.40} & \textbf{89.07} & - & \textbf{85.44} & \underline{73.08} & 71.22 \\ \bottomrule
\end{tabular}
\caption{Zero-shot classification performance on geographically diverse household object from DollarStreet and GeoDE using BLIP-2. See table \ref{tab:geo} for results with CLIP ViT-B/16.}
\label{app-tab:geo}
\end{table}

\begin{table}[h]
\centering
\begin{tabular}{lrrrrrrrrrrr}\toprule
&ImageNet &v2 &-A &-R &Sketch &Food &Flowers &Aircraft &Pets &avg \\\midrule
\multicolumn{11}{c}{Overall Accuracy} \\
\midrule
Vanilla &55.68 &51.26 &26.53 &63.06 &55.35 &83.73 &53.05 &25.65 &\textbf{65.33} &53.29 \\
DClip &56.15 &51.44 &\underline{26.60} &62.53 &54.79 &84.02 &\textbf{54.12} &26.43 &62.36 &53.16 \\
Waffle &\underline{57.10} &\underline{52.11} &\textbf{26.83} &\underline{64.78} &\underline{55.88} &83.61 &52.03 &\textbf{27.00} &62.78 &\underline{53.57} \\
CHiLS &56.26 &51.56 &25.63 &63.43 &55.54 &\underline{84.29} &53.55 &\underline{26.94} &\underline{64.95} &\underline{53.57} \\
Ours &\textbf{57.28} &\textbf{52.27} &26.25 &\textbf{65.13} &\textbf{56.21} &\textbf{84.79} &\underline{54.03} &25.23 &63.42 &\textbf{53.85} \\
\midrule
\multicolumn{11}{c}{Accuracy on Worst $20\%$ of Classes}
\\\midrule
Vanilla &6.74 &5.40 &\textbf{3.39} &24.12 &3.99 &57.36 &0.00 &0.00 &\textbf{27.52} &14.28 \\
DClip &7.62 &\underline{6.55} &2.80 &22.83 &4.20 &58.52 &0.00 &0.00 &17.34 &13.32 \\
Waffle &\textbf{8.22} &5.98 &\underline{3.25} &\underline{24.16} &\textbf{5.14} &57.21 &0.00 &0.00 &25.84 &14.42 \\
CHiLS &8.05 &6.25 &2.90 &24.02 &\underline{4.87} &\underline{60.90} &0.00 &0.00 &\underline{27.13} &\underline{14.90} \\
Ours &\underline{8.19} &\textbf{6.90} &2.87 &\textbf{25.65} &4.53 &\textbf{61.44} &0.00 &0.00 &25.28 & \textbf{14.98} \\
\bottomrule
\end{tabular}
\caption{Accuracy on extra datasets for BLIP-2.}
\label{tab:extra_dsets_blip}
\end{table}

\begin{table}
\begin{center}
\resizebox{\textwidth}{!}{
\begin{tabular}{l|cc|cc|cccc}
\toprule
 & \multicolumn{2}{|c|}{\emph{Geographic}} & \multicolumn{2}{|c|}{(MIT) \emph{States}} & \multicolumn{4}{|c}{\emph{Hierarchical}} \\
Method & Dollarstreet & Geode & Coarse & Fine & Entity13 & Entity30 & Nonliving26 & Living17 \\
\midrule
\multicolumn{9}{c}{Accuracy} \\
\midrule
Vanilla & 51.21 & 90.34 & 78.24 & 59.07 & 68.22 & 68.43 & 77.27 & 92.96 \\
DCLIP & 49.80 & 91.14 & 77.80 & 55.65 & 68.64 & 68.49 & 76.32 & 93.35 \\
Waffle & 51.22 & 91.34 & 78.31 & 58.47 & 68.95 & 68.66 & 77.13 & 92.80 \\
CHiLS & 51.62 & 90.85 & 78.83 & 58.55 & 69.33 & \textbf{70.69} & \textbf{79.55} & \textbf{93.65} \\
Ours & \textbf{52.33} & \textbf{91.58} & \textbf{79.33} & \textbf{59.90} & \textbf{71.47} & 70.59 & 79.25 & 93.59 \\
\midrule
\multicolumn{9}{c}{Average Worst Subpopulation Accuracy} \\
\midrule
Vanilla & 37.18 & 83.49 & 53.83 & 29.49 & 21.77 & 36.27 & 57.12 & 82.24 \\
DCLIP & 36.69 & 84.50 & 53.01 & 27.81 & 22.54 & 36.87 & 54.54 & 81.82 \\
Waffle & 37.18 & 85.20 & 53.94 & 28.96 & 21.88 & 36.87 & 56.37 & 81.41 \\
CHiLS & 37.98 & 84.56 & 53.69 & 29.22 & 23.77 & \textbf{42.50} & 59.50 & \textbf{83.53} \\
Ours & \textbf{39.11} & \textbf{85.42} & \textbf{54.26} & \textbf{30.10} & \textbf{25.31} & 39.03 & \textbf{59.54} & \textbf{83.53} \\
\midrule
\multicolumn{9}{c}{Accuracy for Worst $20\%$ of Classes} \\
\midrule
Vanilla & 18.60 & 71.63 & 52.12 & 26.79 & 34.38 & 32.50 & 49.15 & 74.00 \\
DCLIP & 18.64 & 73.57 & 51.35 & 24.46 & 36.48 & 33.21 & 46.60 & \textbf{78.50} \\
Waffle & 18.78 & \textbf{74.98} & 52.31 & 25.17 & 31.41 & 33.46 & 48.40 & 75.24 \\
CHiLS & 20.04 & 72.19 & 53.65 & 26.82 & 36.07 & 31.71 & 52.05 & 75.50 \\
Ours & \textbf{20.96} & 74.61 & \textbf{54.03} & \textbf{28.05} & \textbf{37.55} & \textbf{34.94} & \textbf{53.10} & 76.92 \\
\midrule
\multicolumn{9}{c}{Accuracy for Worst $10\%$ of Classes} \\
\midrule
Vanilla & 11.92 & 59.30 & 41.63 & 18.09 & 29.75 & 21.75 & 40.58 & 70.25 \\
DCLIP & 11.82 & 64.22 & 41.16 & 15.90 & 26.80 & 22.71 & 38.08 & \textbf{76.62} \\
Waffle & 10.69 & \textbf{64.74} & 42.00 & 16.68 & 23.03 & 23.79 & 39.78 & 72.59 \\
CHiLS & 13.64 & 58.82 & \textbf{44.74} & 18.41 & 25.60 & 21.04 & 42.92 & 71.38 \\
Ours & \textbf{14.35} & 62.61 & 44.24 & \textbf{19.29} & \textbf{31.10} & \textbf{25.33} & \textbf{43.50} & 73.25 \\
\midrule
\multicolumn{9}{c}{Accuracy for Worst $20\%$ of Subpopulations} \\
\midrule
Vanilla & 11.17 & 69.50 & 36.23 & 11.78 & 14.54 & 15.62 & 33.90 & 72.07 \\
DCLIP & 11.67 & 71.61 & 35.08 & 10.16 & 14.54 & 14.92 & 30.19 & 73.57 \\
Waffle & 11.64 & \textbf{73.47} & 36.89 & 10.93 & 13.24 & 16.27 & 32.77 & 71.49 \\
CHiLS & 12.58 & 70.55 & 37.44 & 11.76 & 15.23 & 16.88 & \textbf{39.67} & 74.29 \\
Ours & \textbf{14.33} & 72.93 & \textbf{38.21} & \textbf{12.64} & \textbf{17.33} & \textbf{16.94} & 38.86 & \textbf{74.93} \\
\midrule
\multicolumn{9}{c}{Accuracy for Worst $10\%$ of Subpopulations} \\
\midrule
Vanilla & 6.10 & 57.47 & 23.27 & 4.95 & 5.35 & 5.67 & 18.90 & 54.71 \\
DCLIP & 6.08 & 61.38 & 21.71 & 3.74 & 5.77 & 5.04 & 15.20 & 56.43 \\
Waffle & 5.82 & \textbf{63.27} & 23.27 & 4.26 & 4.96 & 6.56 & 17.00 & 54.30 \\
CHiLS & 7.40 & 57.62 & \textbf{24.93} & 4.88 & 5.96 & 6.21 & 21.50 & 57.43 \\
Ours & \textbf{8.62} & 61.26 & 24.72 & \textbf{5.53} & \textbf{6.88} & \textbf{6.88} & \textbf{22.00} & \textbf{59.29} \\
\bottomrule
\end{tabular}
}
\caption{Metrics for each dataset. Results are averaged over CLIP and BLIP-2. Our method's gains are consistent over the eight dataset suite.}
\label{app-tab:per_dset}
\end{center}
\end{table} 

\begin{figure}
    \centering
    \includegraphics[width=\linewidth]{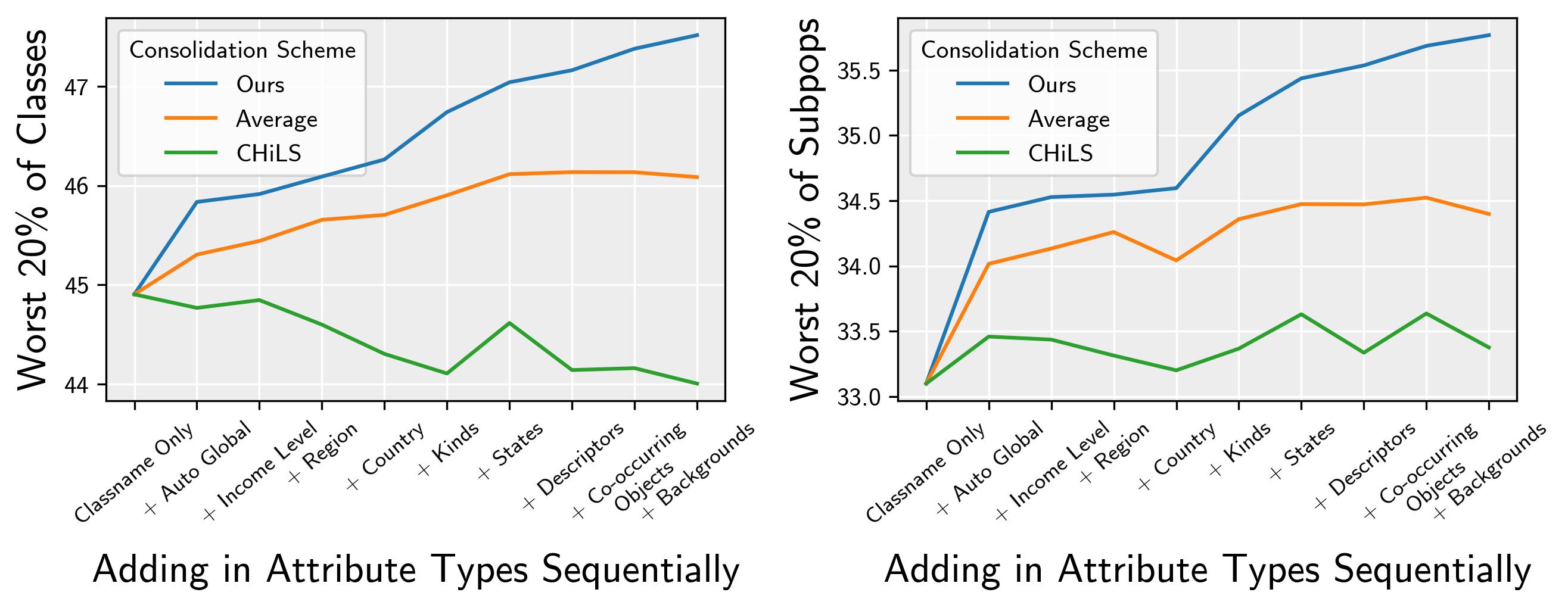}
    \caption{Accuracy for the worst 20\% of classes and subpops, averaged over our dataset suite as we sequentially add new types of attributes using different consolidation schemes. See figure \ref{fig:add_in_attr} in the main text for accuracy overall and over the worst 10\% of classes, along with more discussion. As shown in the main text, our method scales the best as attributes are added sequentially. }
    \label{app-fig:add_in_attr}
\end{figure}

Further, for the analysis in Section \ref{sec:adding_in_attrs}, we show performance using the similar metrics of accuracy over the worst $20\%$ of classes and subpopulations, as shown in most tables. See figure \ref{app-fig:add_in_attr}. Trends are the same as in the main text, though slightly less pronounced. To be clear, our consolidation yields best performance, while others either saturate or deteriorate.   

\begin{figure}
    \centering
    \begin{minipage}{0.85\textwidth}
    \centering
    \includegraphics[width=\linewidth]{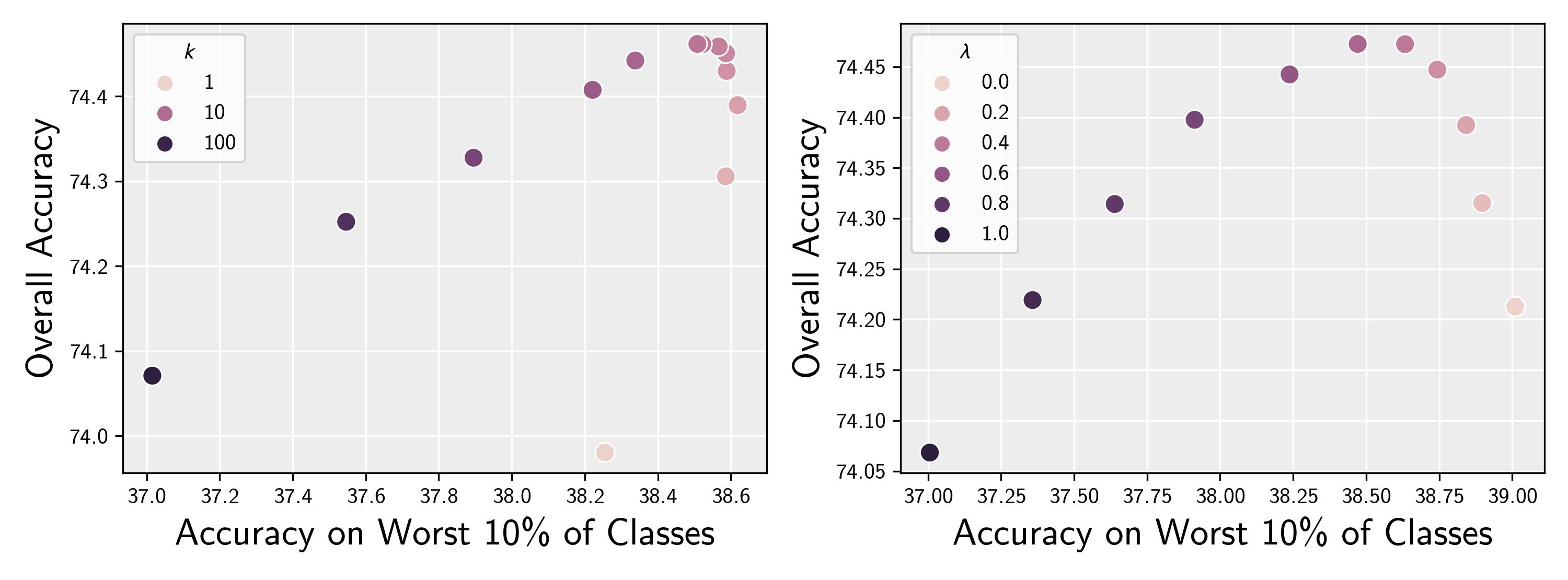}
    \end{minipage}
    \begin{minipage}{0.85\textwidth}
    \centering
    \includegraphics[width=\linewidth]{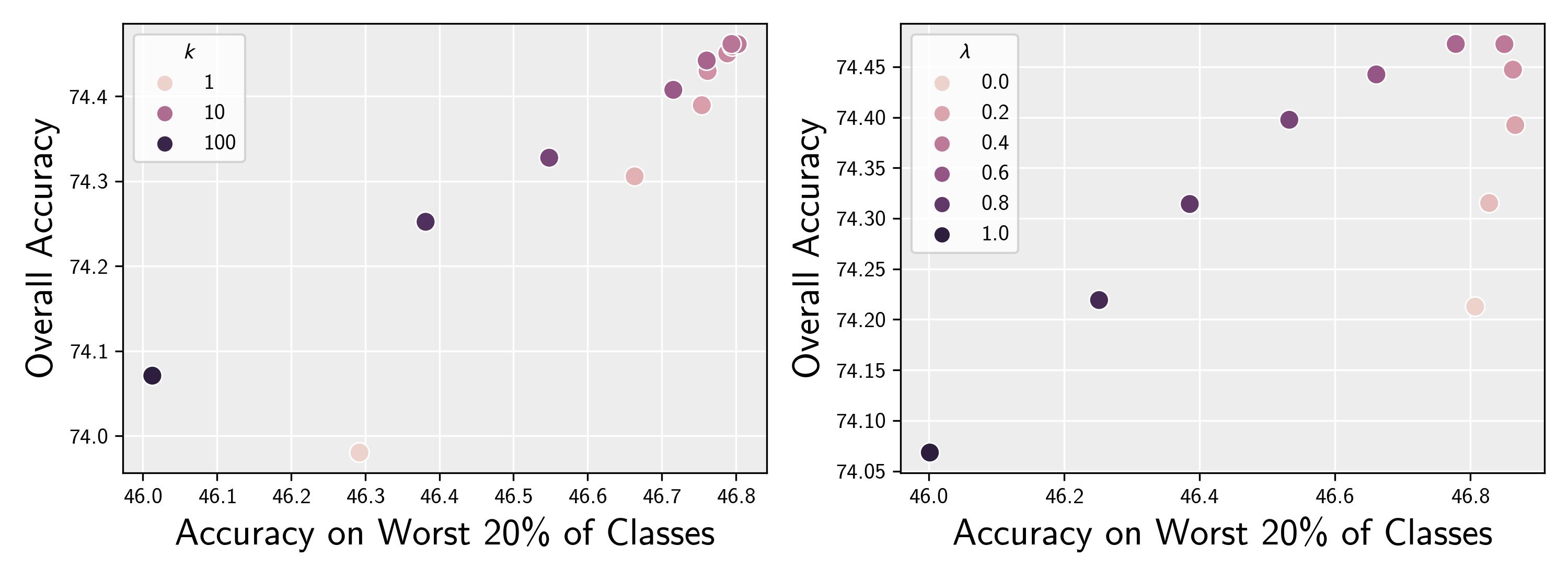}
    \end{minipage}
    \caption{We replicate figure \ref{fig:trade-off} using metrics that look at a larger portion of the worst classes. A similar tradeoff emerges, though in a slightly less pronounced way. We note that this is expected, as increasing the number of classes considered likely also increases the number of less diverse classes included.}
    \label{app-fig:tradeoffs}
\end{figure}

Lastly, we also show additional plots for the analysis in Section \ref{sec:trade-off}. In the main text, we plotted accuracy overall vs. over the worst $5\%$ of classes. We choose to show accuracy over the worst $5\%$ because it most clearly conveys the tradeoff we observe. Figure \ref{app-fig:tradeoffs} shows this tradeoff still exists when looking at other percentiles, though it is less pronounced, which is expected.

\section{Why did CHiLS fail on ImageNet for CLIP?}
\label{app-sec:chils_clip}

In Table \ref{tab:extra_dsets_clip}, we observe CHiLS to fail catastrophically for ImageNet using a CLIP. We conjecture the issue arises due to the high number of classes ($1k$), and even larger set of subclasses (about $10k$). Each logit in CHiLS is the product of a softmax output over $1k$ options and a softmax output over about 11k options. Furthermore, because CLIP similarities usually fall within a small range (about $0.1-0.3$), the difference in final logits may be so small that noise from rounding errors dominates the signal. Notably, \textbf{CHiLS does not fail on ImageNet when using BLIP-2 and our other results on CHiLS (i.e. on hierarchical datasets) closely match the results reported in the original CHiLS paper, suggesting that our implementation is correct}, and that the problem arises due to small differences in CLIP similarities. On BLIP-2, while CHiLS does not fail catastrophically, it still underperforms compared to our method, with accuracy about $1\%$ lower.

One could likely fix this problem by changing the temperature of the softmax, but we opt to faithfully follow the original method. Indeed, a modified version of CHiLS without the softmax (which amounts to our method using only the \emph{Kinds} query (see table \ref{app-tab:llm_queries} and with $k=1$) does not fail catastrophically on ImageNet, though the overall accuracy and accuracy on worst classes for this `fixed' CHiLS does not exceed our method's results. 

While this change seems small, we believe it encapsulates a difference in philosophy between CHiLS and our method. CHiLS is designed for datasets with clear hierarchy, where each input can fit neatly into one of many mutually exclusive subpopulations. In contrast, we argue that diversity emerges in many ways, with overlapping subpopulations arising from attributes drawn along various axes. By taking a softmax, CHiLS requires that an input is not only similar to one subpopulation within a class, but that it is also dissimilar from the other subpopulations. In our method, instead of seeking to explicitly name all subpopulations in a mutually exclusive way, we enumerate many potential attributes, and create a flexible consolidation that only requires an input to be close to a few subpopulations within its class for it to be classified correctly. 


\section{When can we cram an entire class in one vector, and when can we not?}

Arguably, diversity within classes is unavoidable, as two instances can vary in numerous ways (discussed further in Section \ref{sec:diversity_many_forms}). How then, have classifiers enjoyed success under the one-vector-per-class paradigm, despite its tension with intra-class diversity? First, we note these performance disparities are often obfuscated in metrics like overall accuracy; indeed, the supervised classifiers studied above each achieve impressive overall accuracies. Nonetheless, the tension can be somewhat resolved if (i) one learns embeddings that reduce the diversity that is present in input space, and/or (ii) the single vector learned per class contains features that are unique to the class and present across class instances, despite intra-class variance that persists in the embedding space. We expand on these below.

\subsection{Ideal conditions for the one-vector-one-class paradigm}
\label{app-sec:unique_and_invariant}

\looseness=-1
Most modern vision classifiers consist of a deep feature encoder, mapping images to a rich embedding space, followed by a linear classification head, mapping embeddings to class logits. The linear classification head consists of a single vector (and a scalar bias) per class. A linear classification head is accurate if, for any instance from the $i^{th}$ class, the activation on the $i^{th}$ class vector must be higher than the activation for any other class vector. We express this mathematically below, with $\mathbf{x}$ denoting the embedding of an image from class $i$, and $\mathbf{c_i, c_j}$ denoting vectors in the classification head. 

\begin{align}
    &\forall \mathbf{x} \in \mathcal{C}_i, \forall j\not=i, \; \text{we require that }\mathbf{x} \cdot \mathbf{c_i} - \mathbf{x} \cdot \mathbf{c_j} > 0 \\
    &\text{Note that } \mathbf{x} \cdot \mathbf{c_i} - \mathbf{x} = \mathbf{x} \cdot (\mathbf{c_i} - \mathbf{c_j}) = \|\mathbf{x}\|\|\mathbf{c_i}-\mathbf{c_j}\|\cos(\mathbf{x}, \mathbf{c_i}-\mathbf{c_j}) \\
    &\text{Thus, } \forall\mathbf{x} \in \mathcal{C}_i, \forall j\not=i,\text{ we require that }\cos(\mathbf{x}, \mathbf{c_i}-\mathbf{c_j}) > 0  
\end{align}

The last step arises because norm is always non-negative. Now, let us focus on different components of this required condition (by definition) for an accurate one-vector-per-class classification head. First, the single vector $\mathbf{c_i}$ must contain contain a set of features that are \emph{unique} to that class. That is, these features remain when considering the residual $\mathbf{c_i - c_j}$ for any $i \not = j$. Secondly, the unique features that discriminate the class from all others must also be aligned with every instance of the class. In other terms, these unique features must be \emph{invariant} to any diversity within the class. Also, note that the quantity we expand upon above is simply the margin for classification. In the ideal case, this margin would be maximized. 

\subsection{Class-supervised training is well suited for the one vector per class paradigm, but VLM pretraining is not}

In traditional class-label supervised training, the feature encoder is jointly optimized with the classification head to minimize a classification loss. Let us consider how this effects the linear classification head and the feature encoder individually. First, fixing the classification head, we see the supervised objective encourages all embeddings from one class to be drawn close to their respective single vector, and consequently, close to one another. In other words, invariance of embeddings within a class is promoted. Next, with the feature encoder fixed, classification head vectors align with embeddings within their class and de-align with embeddings from outside their class. Thus, the classification head vectors are optimized to solely contain the features unique to their class embeddings. Therefore, training with traditional class-label supervision directly promotes the invariance and uniqueness properties required for the success of the one-vector-per-class paradigm.

On the other hand, VLMs are optimized with markedly different objectives. Many VLMs employ contrastive image-text matching, in which negative examples are far weaker and classes are no longer defined; in some ways, the training is analogous to optimizing a classification task with an infinite number of classes. Indeed, two instances that belong to the same class in a downstream task may have embeddings pushed apart during VLM pretraining, directly going against the aforementioned notion of class-wise invariance. Other common VLM objectives like captioning or question answering promote the descriptiveness of the embedding. Thus, instead of honing in on unique features, embeddings are likely to describe as much as possible. We note that having maximally descriptive embeddings is typically a good thing, as it allows for re-use of the same feature encoder for many downstream tasks, as is done in linear probing with self-supervised encoders. The key caveat is that in those cases, the linear classification head is still exposed to instances from all classes, and thus, each classification head vector can learn to align only with the unique features for its class. In contrast, in the zero-shot setting, the classification head vectors are obtained independently of one another via embedding the names of classes via the text encoder, and thus, it is unreasonable to expect that these vectors satisfy the uniqueness condition.

\begin{figure}
\centering
    \centering
    \includegraphics[width=0.45\linewidth]{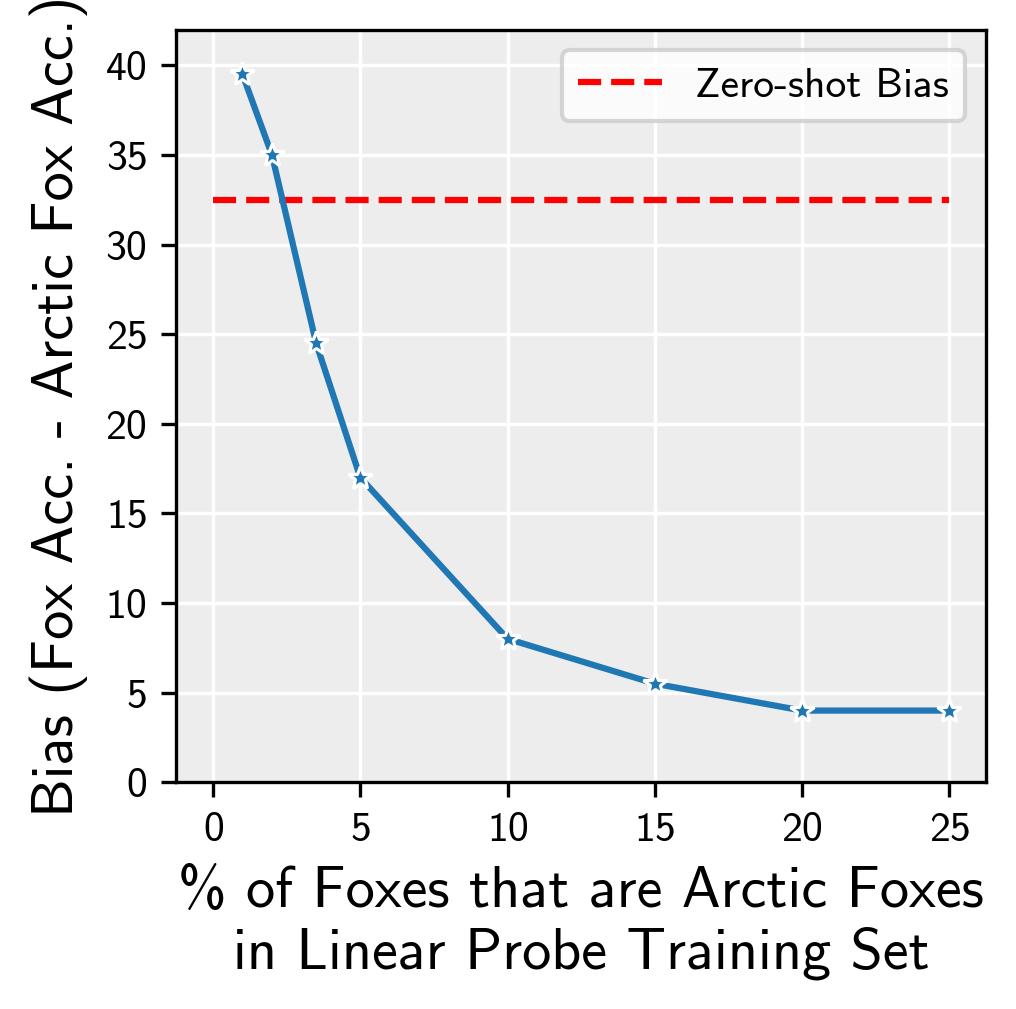}
    \caption{Arctic Fox bias is amplified in zero-shot classifier vs. to supervised linear probes.}
    \label{app-fig:fox_bias_amplified}
\end{figure}

\subsection{Arctic Fox Case Study: Bias can be amplified when using one vector per class paradigm for zero-shot classification}
\label{app-sec:arctic_fox}



Staying in the one-vector-per-class setting, we now compare class vectors obtained directly in a zero-shot manner to those obtained with supervision. Specifically, we focus on the \texttt{Arctic Fox} bias, shown in Figure \ref{fig:bias_examples}. We train a linear classification head over fixed CLIP embeddings used a skewed training set that under-represents \texttt{Arctic foxes} in the training set. We find that the bias of the zero-shot vector is on par with having only $3\%$ of the training images in the \texttt{fox} class be \texttt{Arctic foxes} in the supervised setting, suggesting that limitations of the one vector per class paradigm may be exacerbated in the zero-shot setting.  

\section{One final Trade-off}

In section \ref{sec:trade-off}, we should two hyperparameters that could trade overall accuracy for accuracy over the worst classes. We now present one more, along with a \emph{theoretical} explanation. Throughout the paper, we consider `averaging' to mean computing similarities to multiple vectors and then averaging those similarities; this is how DCLIP and WaffleCLIP average, and will refer to this as \textbf{Average Sims}. However, averaging over prompts as done in originally in CLIP consists of averaging vectors first and then computing similarity to one average vector; we call this \textbf{Average Vecs}. The difference is subtle: in the latter case, an additional normalization occurs when cosine similarity is taken.

We now show theoretically that when all embeddings are normalized (i.e. for CLIP), \textbf{Average Vecs} simply rescales the class score yielded by \textbf{Average Sims} by a factor that measures how \emph{diffuse} the vectors for the class are. Let $x$ be an image embedding and $\{v_1, v_2, \dots, v_k\}$ be subpopulation vectors for a given class. We assume all vectors are normalized to the hypersphere, as is the case for CLIP. That is, $\|v_i\| = 1$ for all $i$ and $\|x\|=1$. Let $\overline{v} := \frac{1}{k}\sum_{i=1}^k{v_i}$ denote the average vector. We compute the class score for \textbf{Average Vecs} below. 

\begin{align*}
\textbf{Average Vecs} &= \cos\left(x, \overline{v} \right) 
= \frac{x\cdot \overline{v}}{\|x\|\|\overline{v}\|} = \frac{x\cdot \frac{1}{k}\sum_{i=1}^k{
 v_i}}{\|\overline{v}\|}= \frac{\frac{1}{k}\sum_{i=1}^k{ x\cdot 
 v_i}}{\|\overline{v}\|} \\
 &= \frac{\frac{1}{k}\sum_{i=1}^k{\cos(x,v_i)}}{\|\overline{v}\|} = \frac{\textbf{Average Sims}}{\|\overline{v}\|}
\end{align*}

To get from line 1 to 2, we utilize the fact that cosine similarity is equivalent to the dot product when both arguments are unit norm. Let us now consider what this result entails. The denominator is the norm of the average vector. This quantity is always between $0$ and $1$. It is lowest when the vectors are most diffuse. Thus, the class score obtained by \textbf{Average Sims} is scaled up to obtain the score for \textbf{Average Vecs} by more when the vectors are diffuse. In other words, averaging the vectors first implicitly upweights vectors corresponding to diverse subpopulations.

\begin{figure}
    \centering
    \includegraphics[width=0.65\linewidth]{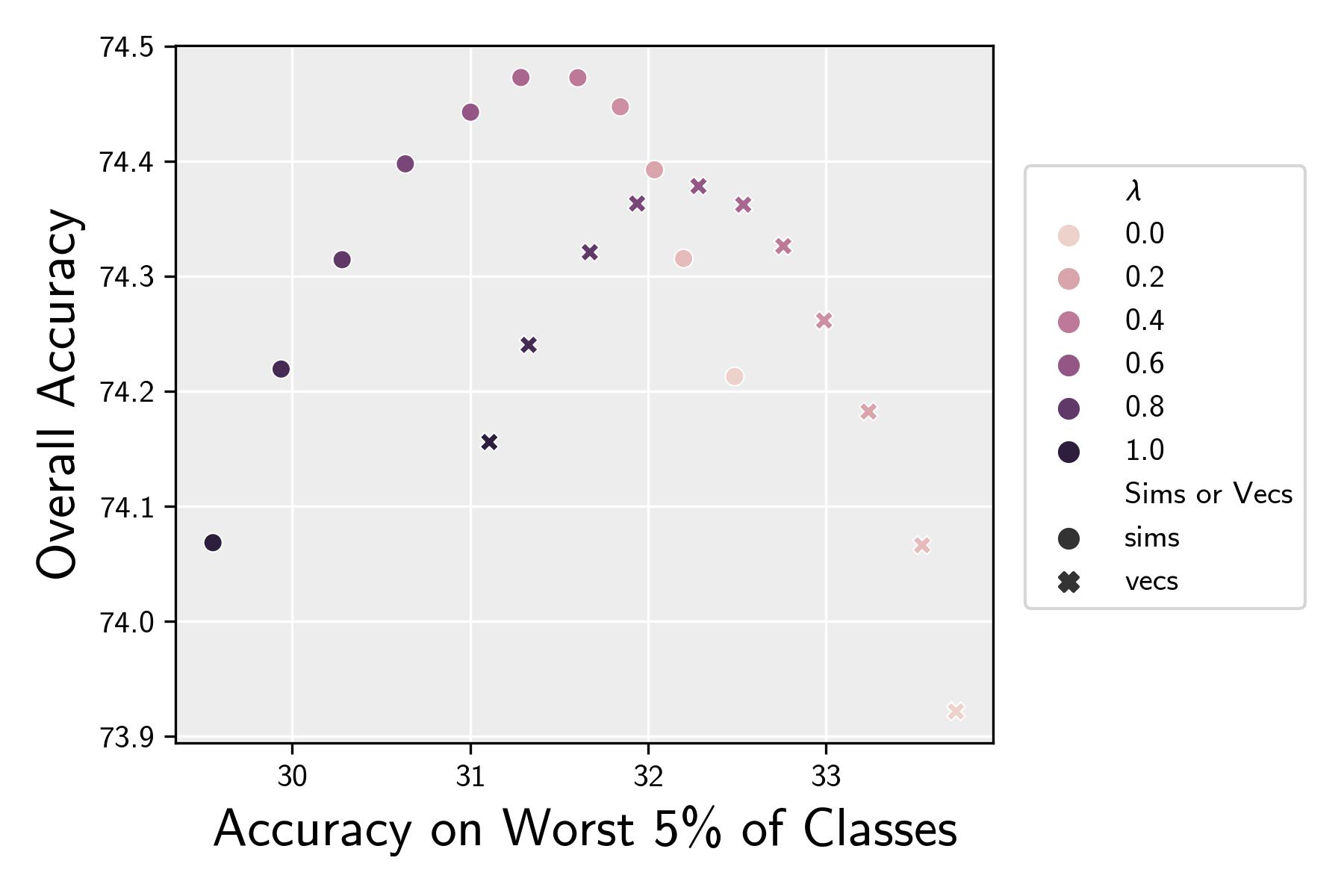}
    \caption{Averaging subpopulation vectors before computing similarity to an image embedding proves to be another way to trade overall accuracy for accuracy on the worst classes. That is, when we first compute similarity to each subpopulation and then average, we obtain higher overall accuracy but lower accuracy on the worst classes, compared to when we first average subpopulation vectors and then compute the similarity to the average vector.}
    \label{fig:sims_or_vecs}
\end{figure}

Based on this simple theory, we would expect the most classes with high diversity to have higher accuracy under \textbf{Average Vecs} compared to \textbf{Average Sims}, as their class scores are inflated more than the less diverse classes. The effect on overall accuracy, however, is not perfectly clear. To inspect this, we perform the same sweep over $k$ and $\lambda$ as in section \ref{sec:trade-off}, except now we additionally try replacing all similarity averaging with vector averaging. Figure \ref{fig:sims_or_vecs} shows the results. We average away $k$ for clarity.  Indeed, averaging over vectors improves accuracy on the worst classes. For high values of $\lambda=1$, we see averaging vectors also slightly improves overall accuracy. However, in the vast majority of values for $\lambda$, overall accuracy is hurt by averaging vectors instead of similarities. We hope this analysis provides insight as to the precise effect of averaging similarities or vectors, which may be relevant to others who wish to explore going beyond one vector per class. 

\end{document}